\documentclass[11pt]{article}

\usepackage{soul}
\usepackage{url}

\usepackage{tikz}
\usetikzlibrary{positioning, automata}
\usepackage{stmaryrd}  

\usepackage{amssymb}

\usepackage[utf8]{inputenc}
\usepackage{graphicx}
\usepackage{amsmath}
\usepackage{amsthm}
\usepackage{booktabs}
\usepackage{algorithm}
\usepackage{algorithmic}
\usepackage[switch]{lineno}
\usepackage{soul}
\usepackage{float}
\usepackage{mathtools}
\usepackage{booktabs}
\usepackage{arydshln}
\usepackage{soul}
\usepackage{simplebnf}
\usepackage{cancel}
\usepackage{multirow}

\usepackage{xcolor}
\usepackage{amsfonts}
\usepackage{xspace}
\usepackage[geometry]{ifsym}

\def\ltlweaknext{\raisebox{-0.55ex}{\mbox{\FilledCircle}}\xspace}
\def\ltluntil{\,\mathcal{U}\,\xspace}
\def\ltlreleases{\,\mathcal{R}\,\xspace}

\def\ltlnext{\bigcirc\xspace}
\def\ltlfinally{\Diamond\xspace}
\def\ltlglobally{\Box\xspace}

\newcommand{\LTL}{{\sc ltl}\xspace}
\newcommand{\LTLf}{{\sc ltl}$_\mathrm{f}$\xspace}

\newcommand{\mathimg}[1]{\vcenter{\hbox{\includegraphics[height=2ex]{imgs/#1}}}}

\begin{document}

\title{LTLZinc: a Benchmarking Framework for
Continual Learning and Neuro-Symbolic Temporal Reasoning}

\author{Luca Salvatore Lorello\thanks{Corresponding author. Part of the work was done while LSL was visiting KU Leuven.}\\
Department of Computer Science\\
University of Pisa\\
University of Modena and Reggio Emilia\\
Italy\\
\texttt{luca.lorello@phd.unipi.it}\\
\and
Nikolaos Manginas\\
KU Leuven\\
NCSR ``Demokritos''\\
Belgium, Greece\\
\texttt{nikolaos.manginas@kuleuven.be}\\
\and
Marco Lippi\\
Department of Information Engineering\\
University of Florence\\
Italy\\
\texttt{marco.lippi@unifi.it}\\
\and
Stefano Melacci\\
Department of Information Engineering\\
University of Siena\\
Italy\\
\texttt{stefano.melacci@unisi.it}
}




\maketitle

\begin{abstract}
{\bf Background:} 
    Neuro-symbolic artificial intelligence aims to combine neural architectures with symbolic approaches that can represent knowledge in a human-interpretable formalism. Continual learning concerns with agents that expand their knowledge over time, improving their skills while avoiding to forget previously learned concepts. Most of the existing approaches for neuro-symbolic artificial intelligence are applied to static scenarios only, and the challenging setting where reasoning along the temporal dimension is necessary has been seldom explored.
    
    {\bf Objectives:}
    In this work we introduce LTLZinc, a benchmarking framework that can be used to generate datasets covering a variety of different problems, against which neuro-symbolic and continual learning methods can be evaluated along the temporal and constraint-driven dimensions.
    
    {\bf Methods:}
    Our framework generates expressive temporal reasoning and continual learning tasks from a linear temporal logic specification over MiniZinc constraints, and arbitrary image classification datasets. Fine-grained annotations allow multiple neural and neuro-symbolic training settings on the same generated datasets.
    
    {\bf Results:}
    Experiments on six neuro-symbolic sequence classification and four class-continual learning tasks generated by LTLZinc, demonstrate the challenging nature of temporal learning and reasoning, and highlight limitations of current state-of-the-art methods.
    
    {\bf Conclusions:}
    We release the LTLZinc generator and ten ready-to-use tasks to the neuro-symbolic and continual learning communities,\footnote{\url{https://github.com/continual-nesy/LTLZinc}.} in the hope of fostering research towards unified temporal learning and reasoning frameworks.
\end{abstract}

\section{Introduction}

The integration of symbolic and sub-symbolic approaches has long been a challenge in artificial intelligence (AI). Throughout the decades, many attempts have been made to combine the expressive power of logic-based frameworks with the capability to handle uncertainty in data that is typical of models such as neural networks~\cite{deraedt2021statistical}. Not only can logic provide a natural formalism for incorporating background knowledge of a specific domain in a human-interpretable language, but it can also provide an invaluable tool to enhance the safety and trustworthiness of neural networks that, despite the stunning performance achieved in many tasks, very often lack reliability and interpretability~\cite{li2022interpretable}.
In this paper, we focus on a specific kind of knowledge whose integration with neural networks has been underexplored. We consider the challenging scenario of handling temporal knowledge, from a twofold perspective: reasoning over sequences and continual learning. The former consists of tasks where the available domain knowledge can guide reasoning over time, as in the case of a safety-critical system where events must follow a sequence of specific constraints~\cite{carnevali2024neuro}. The latter implements a learning setting where observability of classes, domains or tasks, can vary over time, as in the case of an agent that learns concepts through a curriculum of tasks having an increasing difficulty~\cite{lorello2024continual}.
%
Evaluating the performance of AI techniques for these tasks is particularly difficult, because learning unfolds over time and reasoning is performed on top of both relational and temporal components.
Few benchmarks are available, and those that exist seldom focus on all of the dimensions described above. Therefore, building on a preliminary investigation of this research avenue~\cite{lorello2025neuro}, we aim to fill this gap at the intersection of the research areas of neuro-symbolic and continual learning, by proposing a novel benchmarking framework, named LTLZinc, that allows to design a wide variety of problems concerning the use of temporal knowledge. LTLZinc allows to define temporal sequences of perceptual stimuli (e.g., images) possibly associated to different domains, by means of an easy-to-define formalism combining MiniZinc~\cite{nethercote2007minizinc} constraints and linear temporal logic formulas over finite traces (\LTLf)~\cite{de2013linear}.
We release to the scientific community both the generator and a collection of pre-defined tasks: six tasks for temporal reasoning on MNIST and Fashion MNIST (Section~\ref{sec:proposed-seq}), and four for class-continual learning tasks instantiated over MNIST and Cifar-100 (Section~\ref{sec:proposed-ccl}).
We believe that this benchmarking framework can become a valuable tool for researchers in the areas of neuro-symbolic and continual learning, where the need for novel, challenging and wide-ranging benchmarks is recognized as a crucial issue~\cite{manhaeve2024benchmarking}. 
The rest of this paper is structured as follows. Section~\ref{sec:background} provides background information on linear temporal logic over finite traces, neuro-symbolic AI and continual learning, Section~\ref{sec:related} overviews existing benchmarking frameworks for learning and reasoning over time. Section~\ref{sec:ltlzinc} describes in details the LTLZinc framework, formalizing problems, and discussing the generation procedure and its limitations. The next two sections describe the family of tasks LTLZinc can generate for neuro-symbolic temporal reasoning (Section~\ref{sec:seqtasks}) and continual learning (Section~\ref{sec:inctasks}).
Section~\ref{sec:seqexp} introduces the ready-to-use datasets for neuro-symbolic temporal reasoning and describes extensive experiments with existing modular neuro-symbolic architectures.
Section~\ref{sec:contexp} presents the ready-to-use datasets for continual learning and highlights the results of experiments extending classic continual learning strategies with temporal knowledge about data distribution.
Finally, Section~\ref{sec:conc} concludes the paper.

\section{Background}\label{sec:background}
This section briefly touches upon background information related to learning and reasoning over time.

\paragraph{Linear Temporal Logic Over Finite Traces.}
Linear temporal logic (\LTL)~\cite{pnueli1977temporal} is a modal logic capable of expressing temporal properties along discrete traces, in terms of reachability of states.
Linear temporal logic over finite traces (\LTLf) \cite{de2013linear} is the finite realization of \LTL.
Given a set of atomic propositions $AP = \left\{\texttt{p}, \texttt{q}, \dots\right\}$, a well-formed \LTLf formula $\phi$ possesses the following syntax:\footnote{\LTLf and \LTL share the same syntax, with the exception of the $\ltlweaknext$ operator (not present in \LTL).}
\begin{center}
\begin{bnf}
$\phi$ ::=
| $\top$ : true
| \texttt{p} : atomic proposition
| $(\neg \phi)$ : negation
| $(\phi) \wedge (\phi)$ : conjunction
| $(\phi) \vee (\phi)$ : disjunction
| $(\phi) \rightarrow (\phi)$ : implication 
| $(\phi) \leftrightarrow (\phi)$ : double implication
| $(\ltlnext \phi)$ : next
| $(\ltlweaknext \phi)$ : weak next
| $(\ltlglobally \phi)$ : globally
| $(\ltlfinally \phi)$ : finally
| $(\phi) \ltluntil (\phi)$ : until
| $(\phi) \ltlreleases (\phi)$ : releases.
\end{bnf}
\end{center}
We will omit parentheses when precedence is clear from the context. The set $\left\{\neg, \wedge, \ltlnext, \ltlweaknext, \ltluntil\right\}$ is functionally complete, and every other operator can be defined by means of equivalence relations.
Let $\pi_t \in 2^{AP}$ be an interpretation (i.e., a truth assignment for all the variables in $AP$ at timestep $t$) and $\pi = \pi_1\pi_2\dots\pi_n$ be a finite trace (i.e., a sequence of $n < \infty$ interpretations), then the semantics of \LTLf is defined by induction on the syntactic structure. A trace $\pi$ satisfies the formula $\phi$ at timestep $t \in \left\{1, 2, \dots, |\pi|\right\}$, denoted as $\pi, t \models \phi$, if and only if:
\begin{align*}
\pi, t &\models \texttt{p} &\text{iff } \texttt{p} \in \pi_t\\
\pi, t &\models \neg \phi &\text{iff } \pi, t \not\models \phi\\
\pi, t &\models \phi_1 \wedge \phi_2 &\text{iff } \pi, t \models \phi_1\text{ and } \pi, t \models \phi_2 \\
\pi, t &\models \ltlnext \phi &\text{iff } \pi, t + 1 \models \phi \text{ and } t < |\pi| \\
\pi, t &\models \ltlweaknext \phi &\text{iff } \pi, t + 1 \models \phi \ \text{or} \ t = |\pi| \\
\pi, t &\models \phi_1 \ltluntil \phi_2 &\text{iff for some } t' \leq |\pi|:\ \pi, t' \models \phi_2 \\ &&\ \text{and for all } t'' \leq t': \ \pi, t'' \models \phi_1.
\end{align*}
The satisfiability of the entire formula, with respect to the given trace, is $\pi \models \phi \text{ iff } \pi, 1 \models \phi$ (i.e., the first timestep of the trace satisfies the formula). Intuitively, the semantics of \LTLf can be understood as follows: $\ltlnext \phi$ is true if $\phi$ is true in the next interpretation of the trace (which must exist), similarly $\ltlweaknext \phi$ is true either if $\phi$ is true in the next interpretation, or the trace has ended, $\phi_1 \ltluntil \phi_2$ is true if $\phi_1$ holds true at least up to the first occurrence of $\phi_2$ (which must happen at some point in the trace).
Derived operators are: $\ltlfinally \phi \equiv \top \ltluntil \phi$ ($\phi$ will become true sometime in the future), $\ltlglobally\phi \equiv \neg \ltlfinally \neg \phi$ ($\phi$ is always true) and $\phi_1 \ltlreleases \phi_2 \equiv \neg (\neg \phi_1 \ltluntil \neg \phi_2)$ ($\phi_2$ must hold up to, and including, the first occurrence of $\phi_1$, if $\phi_1$ never becomes true, $\phi_2$ must be true forever).
There exist well-known patterns encoding common temporal properties in \LTL formulas~\cite{dwyer1998property}, however semantic differences between \LTL and \LTLf can lead to counter-intuitive and hard to identify behaviors, especially for formulas describing re-occurring events.
For example, the formulas $\ltlfinally\ltlglobally \phi$ and $\ltlglobally\ltlfinally \phi$ have different meaning in \LTL (``at some point in the future, $\phi$ will become true and never return to a false state``, and ``$\phi$ will become true infinitely often'', respectively), while in \LTLf they are equivalent (``$\phi$ is true in the final timestep'').
Traces of propositional assignments can be interpreted as strings over the alphabet $2^{AP}$ and \LTLf semantics can be re-defined in terms of language membership. The class of languages which can be defined by \LTLf formulas is a proper subset of regular languages~\cite{de2013linear}, and there exist $\mathsf{PSPACE}$-complete algorithms to convert a formula into a deterministic finite state automaton (DFA). Symbolic finite state automata (SFA)~\cite{veanes2010symbolic} extend DFAs to infinite alphabets, by replacing transition labels with expressions interpreted over Boolean algebras. Problems in LTLZinc are encoded by an extension of \LTLf to constraints over finite domains~\cite{nethercote2007minizinc}, meaning that they are still formally equivalent to DFAs (as every possible combination of assignments is still finite), however an SFA-based representation is exponentially more concise.

\paragraph{Neuro-Symbolic Integration of Logic Formulas.}
The main challenge of neuro-symbolic integration~\cite{besold2021neural} consists in providing an interface between two components: learning by means of neural networks, which requires representations in a continuous space, and reasoning, which often benefits from discrete representations of symbols.
A popular approach for the integration of logic knowledge, known as the model-theoretic approach~\cite{marra2024statistical}, is to relax truth assignments, in a way which extends Boolean algebra in a continuous and differentiable space.
Fuzzy logic~\cite{badreddine2022logic} extends interpretations in the continuous range $[0, 1]$ and replaces Boolean conjunction with a t-norm.\footnote{In propositional logic, every other operator is constructed by exploiting the definition $\neg p = 1 - p$ and equivalence axioms, for fuzzy first order logic systems, additional components are required.} The obtained expressions are equivalent to the original Boolean formula at boundary values, but allow differentiability by means of a progressive transition between truth values, which can constrain the learning procedure \cite{gnecco2015foundations,melacci2021domain}. Different choices of t-norms are possible, each characterized by different advantages and drawbacks. Although empirically useful for many applications, there are both theoretical and experimental evidences demonstrating a general unsuitability of fuzzy-based relaxations for learning in a differentiable setting~\cite{van2022analyzing}.
An alternative framework for model-theoretic neuro-symbolic integration, overcoming some issues with differentiable fuzzy logics \cite{van2022analyzing}, is based on probabilistic inference~\cite{xu2018semantic, manhaeve2018deepproblog, winters2022deepstochlog}. In this framework, with strong theoretical and computational foundations in statistical and relational artificial intelligence~\cite{raedt2016statistical}, Boolean propositions are interpreted as Bernoulli random variables and logic connectives as set operators.
 While in the case of fully independent random variables, the probabilistic interpretation and the product t-norm fuzzy interpretation both yield correct results, the latter incorrectly overestimates probabilities in the general case, due to the semantics of disjunction. The neutral sum problem is the phenomenon arising when disjunction semantics does not satisfy idempotence (i.e., $a \vee a = a$ in a Boolean setting, but $a \oplus a \not = a$ in its numeric extension), the disjoint sum problem arises when disjunction semantics does not correspond to set union (i.e., $a \vee b \leq a \oplus b$, being exactly equal only if $a$ and $b$ are independent). These problems also arise when computations are performed in a different base (e.g., in log-probability space, or when using different t-norms in a fuzzy setting). 
Weighted model counting (WMC) is a general framework for probabilistic inference, which takes into account proper disjunction semantics, to avoid the neutral and disjoint sum problems. 
Algebraic model counting (AMC)~\cite{kimmig2017algebraic} extends WMC by replacing Boolean operators with elements of an algebraic semiring, allowing to solve a plethora of probabilistic tasks within a single framework. WMC and AMC are in general $\textsc{\#p-complete}$~\cite{chavira2008probabilistic}, however they become tractable when logic formulas possess a specific structure. Knowledge compilation~\cite{darwiche2002knowledge} amortizes execution time by converting input formulas into equivalent target normal forms, on top of which clausal inference (and possibly other classes of tasks) can be executed in polynomial time. Different normal forms are characterized by trade-offs between size and efficiency, as well as their ability to address the neutral sum and disjoint sum problems. 
In this context, we mention sd-DNNF, the smooth and decomposable deterministic negated normal form, a popular knowledge compilation target language, which guarantees correct model counting in polynomial time, also in the presence of neutral and disjoint sums.

\paragraph{Distant Supervision.}

Both neuro-symbolic systems based on probability \cite{manhaeve2018deepproblog, winters2022deepstochlog} and those based on fuzzy logic \cite{van2022analyzing} are routinely applied to distant-supervision tasks, which is an important and well-explored application of neuro-symbolic artificial intelligence.
In a typical neuro-symbolic learning setting, a parametrized neural network $g_{\theta}: \mathcal{X} \mapsto \mathcal{Y}$ maps instances from the input space $\mathcal{X}$ to intermediate labels in a latent concept space $\mathcal{Y}$. A parameter-free symbolic function $\phi: \mathcal{Y}^n \mapsto \mathcal{Z}$ is then used to compute a final label from the set $\mathcal{Z}$. During the learning process, neural network parameters $\theta$ (which affect outputs at the level of $\mathcal{Y}$) are updated by stochastic gradient descent, guided by supervision at the level of $\mathcal{Z}$ (hence the adjective ``distant''). Given an indicator function $\llbracket \cdot \rrbracket$ and an annotated dataset $\mathcal{D} \subseteq \mathcal{X} \times \mathcal{Z}$, the optimization objective in distant supervision can be formally expressed as:\footnote{Here we assume a single neural network $g_\theta$ applied to $x_1, \dots, x_n$ perceptual inputs, i.e., we are assuming a parameter sharing setting.}
$$
\theta^{*} = \mathop{\mathrm{argmin}}_{\theta} \sum_{(x, z) \in \mathcal{D}} \llbracket \phi(g_{\theta}(x_{1}),\dots,g_{\theta}(x_{n})) = z \rrbracket.
$$
MNIST Addition~\cite{manhaeve2018deepproblog} and MNIST XOR~\cite{marconato2023not} are two popular benchmarks for distant supervision. In MNIST Addition, the input space $\mathcal{X} = \left\{\mathimg{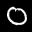}, \mathimg{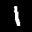}, \dots, \mathimg{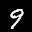}\right\}$\footnote{Throughout the paper we will use this semi-formal notation to represent image classification datasets in an easy to grasp fashion.} corresponds to images from the MNIST digits~\cite{lecun1998gradient} dataset, while the output label is an integer $\mathcal{Z} = [0, 19]$, corresponding to the sum of two intermediate labels $\mathcal{Y} = [0, 9]$. On the other hand, MNIST XOR is an apparently simpler task, in which images can only take values in $\mathcal{X} = \left\{\mathimg{mnist0}, \mathimg{mnist1}\right\}$ and output labels are a boolean $\mathcal{Z} = [0, 1]$, corresponding to the exclusive-or of two intermediate labels $\mathcal{Y} = [0, 1]$.
Distant supervision experiments on the two tasks, however, present radically different outcomes: while MNIST Addition is relatively easy to optimize, MNIST XOR is characterized by considerable optimization challenges.
These differences can be explained by observing the behavior of the symbolic function $\phi$. In neither case $\phi$ is an injection, therefore there are multiple possible inputs which can be mapped to the same output and disambiguation relies on observing data points in a meaningful way. A \textit{reasoning shortcut}~\cite{marconato2023not} is any spurious mapping of intermediate labels $\mathcal{Y}$ which, from the perspective of $\phi$, still produces valid output labels $\mathcal{Z}$. In MNIST Addition, reasoning shortcuts are present due to the fact that addition is a commutative operation, and the fact that multiple combinations yield the same sum (e.g., $\mathimg{mnist0} + \mathimg{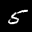} = \mathimg{mnist1} + \mathimg{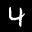} = \mathimg{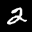} + \mathimg{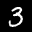} = 5$), however, when enough data is observed in an independent and identically distributed (i.i.d.) fashion, the chance of learning an incorrect mapping function $g_\theta$ is relatively low.
Conversely, in MNIST XOR, the symbolic function yields correct results under two possible intermediate mappings ($g: \left\{\mathimg{mnist0} \mapsto 0, \mathimg{mnist1} \mapsto 1\right\}$, and $g: \left\{\mathimg{mnist0} \mapsto 1, \mathimg{mnist1} \mapsto 0\right\}$), because the function is commutative and symmetric with respect to labels ($\phi(0,0) = \phi(1,1), \phi(0,1) = \phi(1,0)$). In this scenario, there is an equal chance for the optimizer to learn either the correct or the incorrect mapping, unless some prior hypothesis on the labeling of $\mathcal{Y}$ can be injected (e.g., by means of a shallow supervised pre-training).
Reasoning shortcuts are not entirely dependent on the nature of the symbolic mapping $\phi$, and data distribution plays an equally important role.
MNAdd-Shortcut~\cite{marconato2023neuro} is a  non-i.i.d. variation of MNIST Addition, where all the samples containing couples of even digits are observed before any sample containing couples of odd digits (and mixed odd-even couples are never observed). This simple modification in sampling increases dramatically the likelihood of learning reasoning shortcuts, instead of the intended mapping function.

\paragraph{Continual Learning.}
The main assumptions of learning over time are the facts that there exists an order relation between experiences observed by an agent, and that recalling past experiences is an expensive process mediated by a limited-size memory. In this setting, learning is characterized by a recency bias, with experiences closer in time having a stronger effect on the agent, as is widely known in the case of neural networks \cite{PARISI201954}.
If such bias is not adequately counteracted, the agent can be subject to a phenomenon known as \textit{catastrophic forgetting}: performance on past experiences drops dramatically, as soon as the agent attempts to learn on a new experience.
More generally, an agent with limited learning capabilities and finite-size memory is usually affected by the \textit{stability-plasticity trade-off} when it is required to learn different objectives at different times: being able to preserve old knowledge negatively affects the ability to assimilate new experiences, and vice versa. 
Scientific papers with studies that discuss learning over time can be found in the context of different fields, focusing on different aspects of the learning problem, which is known to introduce several challenges \cite{casoni2024pitfalls}.
Foundational aspects at the intersection with optimal control have been recently studied by \cite{melacci2024unified}, proposing a novel learning framework called Hamiltonian Learning. These studies point toward the direction of re-considering state-space models to make spatio-temporally local learning feasible \cite{tiezzi2025back}.
A large variety of literature falls within the Continual learning (CL) topic, a branch of deep learning concerned with developing machines with lifelong learning capabilities that progressively improve their skills without forgetting~\cite{de2021continual}. Approaches for CL are commonly categorized into multiple groups, as described in~\cite{wang2024comprehensive}, such as regularization-based, replay-based, architecture-based methods, among others, also intersecting the world of Reinforcement Learning \cite{abel2023definition}. The role of foundation models is multifold in learning over time. On one hand, they offer useful backbones that can be kept frozen \cite{zhou2024continual}. On the other hand, they can be slowly fine-tuned \cite{zhang2023slca} or patched with the introduction of specific adapters \cite{graziuso2024task}, making them a valid starting point to develop solutions that continuously adapt over time.
Traditional settings in continual learning research are based on the incremental learning framework~\cite{de2021continual}. Within this framework, a learning agent has access to a sequence of datasets $\mathcal{D}^t \subseteq \mathcal{X}^t \times \mathcal{Y}^t$, observed one at the time,\footnote{In the CL literature a single dataset $\mathcal{D}^t$ is called ``task''. To avoid confusion with the notion of task in neuro-symbolic AI, throughout this work we will use the terms ``episode'' or ``learning experience''.} for each timestep $0 \leq t < T$. Samples within each $\mathcal{D}^t$ can be shuffled and drawn in an i.i.d. fashion and trained until convergence (i.e., within each experience, a single dataset can be processed following traditional deep learning frameworks).
The goal of the agent within an incremental learning framework is to minimize the statistical risk:
$$\sum\limits_{t=0}^T \mathbb{E}_{\mathcal{D}^t}[\ell(f_{t;\theta}(\mathcal{X}^t), \mathcal{Y}^t)],$$
where $\ell$ is a loss function applied to neural network $f_{t;\theta}$, and access to previous datasets $\mathcal{D}^{t'}$, for $t' < t$, is limited or forbidden.
Restricting access to previous datasets makes the estimation of the statistical risk and its optimization challenging, as the empirical risk associated with a single dataset becomes an unreliable proxy of the entire distribution.
By imposing restrictions on the structure of the incremental learning framework, it is possible to define \textit{families of incremental settings}. Noting that the marginal distribution of inputs in the incremental framework satisfies $P(\mathcal{X}^t) \not = P(\mathcal{X}^{t+1})$ for each $0 \leq t < T-1$, three traditional settings can be defined by imposing a structure on the marginal distributions of outputs $P(\mathcal{Y}^t)$.
\textbf{Class-incremental learning} (CIL) is characterized by $P(\mathcal{Y}^t) \not = P(\mathcal{Y}^{t+1})$ and $\mathcal{Y}^t \subset \mathcal{Y}^{t+1}$, meaning that each experience increases the number of labels available for classification, presenting new samples, possibly associated with different distributions.
\textbf{Domain-incremental learning} (DIL), imposes instead $P(\mathcal{Y}^t) = P(\mathcal{Y}^{t+1})$ and $\mathcal{Y}^t = \mathcal{Y}^{t+1}$, meaning that the output distribution is stationary over time (while the input distribution changes).
Finally, \textbf{Task-incremental learning} (TIL) requires $\mathcal{Y}^t \not = \mathcal{Y}^{t+1}$, meaning that each neural network $f_{t;\theta}$ has to learn disjoint tasks (and also that a selector mechanism is required at inference time, to determine the proper task-id $t$ for each sample).
Outside the boundaries of incremental learning, additional frameworks have been proposed for continual learning~\cite{wang2024comprehensive}, for instance, task-free continual learning~\cite{aljundi2019task} does not rely on segregating each experience in discrete units, blurred boundary continual learning~\cite{bang2021rainbow} allows partial overlaps between experiences, online continual learning~\cite{aljundi2019gradient} forces a single unshuffled pass over each dataset (dropping the i.i.d. observability assumption within each experience), and continual pre-training~\cite{sun2020ernie} focuses on continually improving generalization on future tasks. A wider perspective on the problem of learning over time, framed on the current context in which AI is quickly evolving, is the one of Collectionless AI \cite{gori2023collectionless}, which focuses on lifelong learning with online learning dynamics in multi-agent distributed networks.

\section{Related Works}\label{sec:related}
Learning over time and reasoning about time are recognized as important components of artificial intelligence. However, evaluating progress in these areas poses significant challenges, as evaluation frameworks have reached limited maturity, compared to other fields of artificial intelligence. Assessment of learning over time is limited by the fact that the majority of frameworks focus on short horizons, with simplified (chain-like) temporal behavior, while temporal reasoning is often evaluated by ad-hoc experiments yielding hard-to-compare insights across different methods.

\paragraph{Traditional Learning Over Time.}
Traditional benchmarks for continual learning aim to evaluate the interaction between an agent and non-stationary streams of stimuli, in order to pinpoint forgetting issues and to characterize the agent's stability-plasticity trade-off. Benchmarks are often built around perceptually simple image classification datasets, in order to decouple the issues caused by non-stationary observations from those caused by a challenging learning setting, reducing the amount of experimental noise which can affect the assessment of a continual learning strategy.
SplitMNIST and Split-Cifar10/100/110~\cite{zenke2017continual,maltoni2019continuous} are popular class-incremental benchmarks, exposing the agent to class labels disjointly split into five consecutive learning episodes. In a similar fashion, SplitImageNet~\cite{rebuffi2017icarl}, SplitTinyImageNet~\cite{de2021continual}, SplitCUB200~\cite{lomonaco2021avalanche}, Split Omniglot~\cite{schwarz2018progress}, and others, attempt to increase the learning difficulty of class-incremental learning, by means of more complex perceptual features, longer temporal horizons or a larger number of classes.
PermutedMNIST~\cite{zenke2017continual} and PermutedOmniglot~\cite{schwarz2018progress} are synthetic task-incremental benchmarks where classification must be performed against images which are subject to a randomized pixel permutation,  different in each episode. Similarly, RotatedMNIST~\cite{lomonaco2021avalanche} and RotatedOmniglot~\cite{lomonaco2021avalanche} present images with a random rotation chosen for each episode.
Core50~\cite{lomonaco2017core50} is a hierarchical classification benchmark subject to discrete domain shifts (achieved by capturing images of the same objects under 11 different lighting and background contidions), which can be used for class-incremental and domain-incremental experiments.
ConCon~\cite{busch2024truth} is a confounded visual dataset characterized by a single classification rule, associated with different spurious correlations for each episode. Experiments on ConCon highlight the fact that, in spite of low forgetting, traditional continual learning approaches are significantly more susceptible to spurious correlation than baselines observing images in an atemporal fashion, even when equipped with infinite-size memory buffers.
Traditional benchmarks attempt to model non-stationary observations in an unnatural fashion, by splitting data according to episodes, which, taken in isolation, are still characterized by i.i.d. observations. Stream51~\cite{roady2020stream} is a class-incremental benchmark characterized by images extracted from consecutive video frames, exposing the agent to a domain shift which is similar to how humans perceive the world.
CLEAR~\cite{lin2021clear} addresses both domain shift and concept drift, by exposing the agent to images collected over a long timeframe (10 visual concepts evolving over 10 years).
Another of the issues of traditional continual learning benchmarks is their incremental nature, which does not allow to assess the agent response to cyclic observations. CIR~\cite{hemati2023class} is a benchmarking framework capable of generating streams of repeating observations for class-based and domain-based continual learning experiments with non-incremental patterns. The CIR generator allows to create streams according to either of two modalities: slot-based generation, which constrains the number of samples observed for each class inside each experience, and sampling-based generation, which is based on defining different probability distributions for each class. Albeit a simple modification of traditional incremental learning settings, experiments on CIR expose weaknesses of traditional continual learning strategies, when classes are re-observed multiple times during the learning experiences.
A similar framework with re-occourring classes is SCoLe~\cite{lesort2023challenging}. In SCoLe, streams are protracted for extremely long temporal horizons (potentially infinite), and the agent is able to observe only a subset of classes at the time, with the goal of decoupling the evaluation of short-term catastrophic forgetting and long-term knowledge acquisition. Classes observed inside each episode can be restricted to follow different probability distributions, or a set of hard-coded constraints (such as appearing only for a certain number of episodes).
Infinite dSprites~\cite{dziadzio2024infinite} is a procedural generator for potentially infinite streams of potentially infinite classses (governed by finite generative factors), with the aim of assessing the goodness of disentangled representations, open-set classification and zero-shot generalization in a continual learning setting.
CLEAR, CIR, SCoLe and Infinite dSprites are characterized by temporal priors which position them in a setting which is more challenging than traditional incremental learning. However, since these priors have limited degrees of freedom, ad-hoc solutions capable of hard-coding such priors can significantly outperform more general continual learning strategies. In contrast, our evaluation framework relies on \LTLf to enforce arbitrary temporal behavior over finite sequences of episodes. 

\paragraph{Learning Over Time with Background Knowledge.} There are limited works lying at the intersection between continual learning and neuro-symbolic artificial intelligence. As a result, few tools are available to benchmark reasoning capabilities over time.
MNAdd-Seq, MNAdd-Shortcut, and CLE4EVR~\cite{marconato2023neuro} characterize the temporal evolution of reasoning shortcuts~\cite{marconato2023not} over simple arithmetic and concept-based reasoning domains framed within traditional incremental learning framework, with disjoint class partitions observed one at the time across short temporal horizons.
KANDY~\cite{lorello2024kandy} is a benchmarking framework for task-incremental abstract visual reasoning. Instances of KANDY are binary classification datasets generated sequentially, according to a different Prolog rule for each episode. Positive and negative images are Kandinsky patterns, consisting of simple geometric shapes combined recursively according to spatial primitives. User-specified datasets, allow learning agents to be exposed to arbitrarily complex first-order logic classification rules, enabling curriculum-learning and multi-task generalization experiments over a simple perceptual domain.
KANDY-Concepts~\cite{lorello2024continual} is an instance of KANDY for concept-based continual learning, where classification rules are built around object properties, and presented to the agent with increasing complexity. 
Although KANDY and LTLZinc share similar expressivity in terms of first-order reasoning, KANDY is still an incremental framework, where the agent is exposed to each task only once. Moreover, task ordering in KANDY entirely relies on human annotations, while LTLZinc allows to automatically generate multiple curricula, from a single \LTLf formula.

\paragraph{Reasoning About Time.}
In spite of the increasing popularity of temporal reasoning for visual question answering on videos~\cite{sun2021video} and chain-of-thought temporal reasoning capabilities of large language models~\cite{xiong2024large,ji2025chain,chu2023timebench}, only a handful of works address the issue of benchmarking temporal reasoning from a neuro-symbolic perspective.
Due to the limited availability of evaluation frameworks in the domain of temporal reasoning, novel methods are often validated against synthetic datasets generated ad-hoc. In the context of \LTLf-driven reasoning, we cite the synthetic datasets proposed by DeepDFA~\cite{umili2023grounding,umili2024deepdfa} and NesyA~\cite{manginas2024nesya}, and the one proposed by NS-TPP~\cite{yang2024neuro} for temporal point processes.
LTLUnsolved254 and LTLPattern126~\cite{hahn2020teaching} are datasets of symbolic traces (of length 254 and 126, respectively) satisfying a collection of human-curated \LTL formulas, with the aim of evaluating trace generation capabilities of transformer-based architectures.
Temporal Logic Video~\cite{choi2024towards} is a synthetic collection of datasets for long-horizon activity and object detection in videos, generated as sequences of images, sampled according to probabilistic automata corresponding to five simple \LTL formulas. Although stemming from similar considerations, Temporal Logic Video and LTLZinc are characterized by significant differences, in terms of expressivity (propositional vs. first order), scope (action understanding vs. generalized temporal learning and reasoning), customizability (static datasets vs. fully-customizable generation framework), and temporal behavior (simple behavior over long sequences vs. complex behavior over possibly short horizons).
LTLBench~\cite{tang2024ltlbench} is a benchmarking framework and a dataset of 2000 temporal reasoning challenges for the evaluation of large language models. In LTLBench, instances are natural language prompts containing a context and an hypothesis. The goal of the agent in LTLBench is to determine whether the hypothesis holds in the given context. The context is built from a random directed graph encoding event dependencies, while the hypothesis is an \LTL formula generated from the context, by randomly sampling nodes from the graph and \LTL operators until a target formula length is achieved. Both the context and the hypothesis are converted into natural language sentences, according to fixed templates. Finally, the symbolic label is generated by converting the context graph into a labeled transition system and checked against the \LTL formula by means of the NuSMV~\cite{cimatti2002nusmv} model checker. LTLBench and LTLZinc differ considerably, with respect to multiple aspects. Perceptual stimuli in LTLBench are textual prompts with limited variability in terms of structure, while LTLZinc stimuli are image classification datasets of arbitrary complexity. LTLBench formulas are generated randomly without user interaction, on the other hand LTLZinc requires user-defined formulas, which can however encode semantically meaningful behaviors. More significant differences lie on the expressivity (LTLBench relies on atomic propositions, while LTLZinc supports relational constraints over integers and enumerations) and scope (large language models reasoning capabilities vs. generalized temporal learning and reasoning) of the two benchmarks.

\section{The LTLZinc Framework}\label{sec:ltlzinc}
We hereby introduce our benchmarking framework for relational and temporal learning and reasoning, covering notation, problem definition, the generation algorithm and its limitations in terms of computational complexity.

\paragraph{Notation.} Let $\mathcal{X}_j, 0 \leq j < N$ be a set of $N$ \textit{perceptual domains}, each associated to a \textit{symbolic domain} $\mathcal{Y}_j$. As an example, $\mathcal{X}_0$ and $\mathcal{X}_1$ could be the domains of MNIST digits and Fashion-MNIST articles~\cite{xiao2017fmnist}, respectively, whereas $\mathcal{Y}_0$ and $\mathcal{Y}_1$ the corresponding sets of (symbolic) classes. We will indicate with $x_j^{t} \in \mathcal{X}_j$ and $y_j^{t} \in \mathcal{Y}_j$ a \textit{perceptual stimulus} and its \textit{symbolic label} observed at discrete time $0 \leq t < T$.\footnote{To avoid confusing the stimulus/domain index with the time index, we will sometimes use letter subscripts in place of numbers (e.g., $A$, $B$ in place of $0$, $1$) to refer to different stimuli, symbols, and domains (e.g., $x_{A}^t$, $x_{B}^t$, $y_{A}^t$, $y_{B}^t$, $\mathcal{X}_{A}$, $\mathcal{X}_{B}$, $\mathcal{Y}_A$, $\mathcal{Y}_B$, etc.).} 
Let $\mathcal{C}$ be a \textit{mapping of relational constraints} from a string identifier to a predicate $\texttt{p/k}$ of arity $1 \leq k < N$ between symbolic labels; at each timestep $t$, the tuple of labels $\langle y^t_0, y^t_1, \dots, y^t_{N-1} \rangle$ corresponding to a certain stimulus $\langle x^t_0, x^t_1, \dots, x^t_{N-1} \rangle$ may or may not satisfy some predicates in $\mathcal{C}$. Throughout this work, we will consider constraints in $\mathcal{C}$ to have MiniZinc~\cite{nethercote2007minizinc} semantics.
For instance, let us consider the following example:
\begin{align*}
\mathcal{X}\colon\quad&\mathcal{X}_A, \mathcal{X}_B, \mathcal{X}_C = [\mathimg{mnist0},\mathimg{mnist9}]\\
 \mathcal{Y}\colon\quad&\mathcal{Y}_A, \mathcal{Y}_B, \mathcal{Y}_C = [0,9]\\
 \mathcal{C}\colon\quad&\texttt{sum}(A, B, C): A+B=C\\
 &\texttt{same}(A, B, C): \texttt{all\_equal}([A, B, C]).
\end{align*}
We may observe the following perceptual stimuli $\mathcal{X}^t = [img_A, img_B, img_C]$ (mapped to their corresponding symbolic labels $\mathcal{Y}^t = [A, B, C]$) at four different timesteps $\mathcal{X}^0 = [\mathimg{mnist3}, \mathimg{mnist5},\mathimg{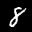}], \mathcal{X}^1 = [\mathimg{mnist2}, \mathimg{mnist5},\mathimg{mnist3}], \mathcal{X}^2 = [\mathimg{mnist4}, \mathimg{mnist4},\mathimg{mnist4}], \mathcal{X}^3 = [\mathimg{mnist0}, \mathimg{mnist0},\mathimg{mnist0}]$. For each of them, we can check which relations in $\mathcal{C}$ hold: $[\{\texttt{sum}: \top, \texttt{same}: \bot\}, \{\texttt{sum}: \bot, \texttt{same}: \bot\}, \{\texttt{sum}: \bot, \texttt{same}: \top\}, \{\texttt{sum}: \top, \texttt{same}: \top\}]$. 
Finally, let $\mathcal{F}$ be a \textit{temporal specification}, defined as an \LTLf formula over the set of relations in $\mathcal{C}$,  grounded with values in $\mathcal{Y}$. For example, $\mathcal{F}: \ltlglobally(\texttt{sum}(A,B,C) \leftrightarrow \ltlnext \texttt{same}(A,B,C))$.\footnote{We will omit terms from predicates appearing in $\mathcal{F}$ (e.g., $\ltlglobally(\texttt{sum} \leftrightarrow \ltlnext \texttt{same})$) when the grounded variables are clear from the context.} This formula models the fact that it will always be true that $\texttt{sum/3}$ is satisfied in a given timestep if and only if the next one satisfies $\texttt{same/3}$. 
This framework allows the definition of arbitrary finite-length linear-time first-order reasoning settings.

\paragraph{LTLZinc Problems.}
An LTLZinc problem is the tuple $\langle \mathcal{X}, \mathcal{Y}, \mathcal{C}, \mathcal{F}\rangle$, instantiated over a finite time horizon $T$ as a collection of observations in either of two modalities.
In \textbf{sequential mode}, an LTLZinc problem produces a \textit{dataset of sequences}, composed of perceptual stimuli $\mathcal{S}_\mathcal{X} = ([x_0^t, \dots, x_{N-1}^t])_{t=0}^{T-1}$, and annotations: $\mathcal{S}_\mathcal{Y} = ([y_0^t, \dots, y_{N-1}^t])_{t=0}^{T-1}$ (symbolic label annotations), $\mathcal{S}_\mathcal{C} = ([c_0^t, \dots, c_{|\mathcal{C}|-1}^t])_{t=0}^{T-1}$ (constraint annotations, $c_i^t \in [\top, \bot]$), and $\mathcal{L} = [\top, \bot]$ (sequence label annotation, true if and only if the sequence satisfies $\mathcal{F}$). Note that the first three are sequential annotations, associated with each timestep of the input stimulus, while the last one is a single binary value associated with the entire sequence. 
Conversely, in \textbf{incremental mode}, an LTLZinc problem is processed into a \textit{sequence of datasets} $\mathcal{D} = [\langle \mathcal{X}^t, \mathcal{Y}^t \rangle s.t. \mathcal{Y}^t \models \mathcal{C}^t]_{t=0}^{T-1}$, given $[\mathcal{C}^t]_{t=0}^{T-1} \models \mathcal{F}$.
In practice, this means that we first generate a sequence of constraint interpretations satisfying the temporal specification, and then, for each timestep, we sample an image classification dataset $\langle \mathcal{X}, \mathcal{Y}\rangle$ coherent with the assignment.
In both cases, the problem definition $\langle \mathcal{X}, \mathcal{Y}, \mathcal{C}, \mathcal{F}\rangle$ can be exploited as background knowledge.

\paragraph{The LTLZinc Dataset Generator.}
LTLZinc specifications augment LTLZinc problems with auxiliary information.
Arbitrary \textit{dataset splits} can be defined with different characteristics (e.g., a test split can present out-of-distribution images, or additional class labels), by means of \textit{domain mappings}, which associate a split with different datasets $\langle \mathcal{X}_{split}, \mathcal{Y}_{split}\rangle$.
\textit{Streams} are mappings $variable\_name \mapsto \langle \mathcal{X}, \mathcal{Y} \rangle$, which allow dynamic associations between perceptual inputs $\langle \mathcal{X}_i, \mathcal{Y}_i \rangle$ and their grounding in the temporal specification $\mathcal{F}$, allowing one-to-one, one-to-many or many-to-many associations between variables and perceptual stimuli observed by the agent. Streams allow to model \textit{partial observability over time}, which can be beneficial both to describe incomplete knowledge or to model domain drifts. Given a constraint, e.g., $\texttt{p}(A,B): A = B$, and a formula with stream specifications, e.g., $\texttt{p}(A \mapsto \langle\mathcal{X}_1, \mathcal{Y}_1\rangle, B \mapsto \langle\mathcal{X}_2, \mathcal{Y}_2\rangle) \wedge \ltlfinally \texttt{p}(A \mapsto \langle\mathcal{X}_3, \mathcal{Y}_3\rangle, B \mapsto \langle\mathcal{X}_4, \mathcal{Y}_4\rangle)$,\footnote{For readability purposes, we will omit stream specifications when the mapping is one-to-one, i.e., $\texttt{p}(A)$ implicitly means $\texttt{p}(A\mapsto\langle\mathcal{X}_A, \mathcal{Y}_A\rangle)$. In our software, streams are decoupled from temporal formulas, for more readable LTLZinc specification files.} the LTLZinc generator will replace the tuple $(A,B)$ with the domains corresponding to each timestep.\footnote{Internally, different streams applied to the same constraint are converted to different constraints by means of syntactic substitution. The example above is equivalent to $\texttt{p}_0(W,X) \wedge \ltlfinally \texttt{p}_1(Y,Z)$, with $W,X,Y,Z$ belonging to different domains.}
Prior to generation, LTLZinc converts the temporal specification $\mathcal{F}$ into an SFA, by means of an off-the-shelf compiler,\footnote{\url{https://github.com/whitemech/flloat}} the generated automaton is included to the LTLZinc specification as an additional form of background knowledge and it is exploited in the following generation phase.

For both sequential and incremental modes, a sequence is generated by means of random walks of length $T$ along the automaton, starting from the initial state. The sequence satisfies the temporal specification (i.e., it is associated with a positive label $\mathcal{L} = \top$) if and only if the the generated trace ends in an accepting state of the automaton.
Such trace is generated by a $T$-limited randomized depth-first search over the automaton, taking into account \textit{biasing options} which can guide the process to avoid, for example, self-loops at the beginning of a sequence, or, avoid starvation for ``orphan'' constraints, appearing in $\mathcal{C}$, but not in $\mathcal{F}$ (meaning that they are known predicates which however do not affect temporal behavior, and thus may never be observed if search is not biased).
For instance, let us consider a dataset consisting of animals, plants and inanimate objects, each associated with a constraint, and a temporal specification indicating that plants will be observed at least once along the sequence. The predicates $\texttt{is\_animal}$ and $\texttt{is\_inanimate}$ are orphans which do not affect the temporal behavior, and act as ``don't care'' values during sampling.\footnote{This means that, when the dataset is used for sequence classification, the notions of ``animal'' and ``inanimate object'' can be safely discarded, however, when the same dataset is used for image classification, they correspond to a subset of the learning targets for which no prior hypothesis is available on their temporal distribution.} At the same time, the automaton associated with the temporal specification is characterized by an accepting sink state which is directly connected to the initial state. Such structure can bias sampling towards non-informative sequences: there is a $0.33\%$ probability of sampling at the first timestep the image of a plant, and every subsequent timestep becomes irrelevant for the sequence label, while we have no guarantee that animals or inanimate objects will be observed at least once by the end of the sequence (i.e., we may have generated a sequence consisting of images of plants only). By counter-biasing the search process, however, we can address both problems, by, for example, avoiding sink states with a probability decreased exponentially over time, or by guiding the generator to observe positive instances of all the orphan predicates at least once with a best effort strategy (i.e., if it is impossible to observe all of them along the sequence, optimize for the maximum number of observations).\footnote{Additional biasing options are available in our software.}

Each transition encountered during the search procedure corresponds to an assignment of constraint labels. A MiniZinc instance is built by means of reification of labels in $\mathcal{C}$ into additional variables, which are constrained to assume values coherent with the transition guard, then the constraint satisfaction problem is solved for the remaining variables (which are assignments for $\mathcal{Y}$ and, possibly, some values of $\mathcal{C}$ which are irrelevant for the given transition). In order to avoid computing a different solution for every timestep of every sequence, LTLZinc computes all the possible solutions for each transition in the automaton once,\footnote{In order to avoid keeping in memory unnecessary solutions for transitions which are never sampled, LTLZinc performs lazy insertions the first time a new assignment is encountered.} and it stores them in a cache, from which a random solution can be sampled in constant time as often as required.

The main sequence sampling procedure, highlighted so far, is employed differently for sequential and incremental generation.
In \textbf{sequential-mode}, the generator samples a target number of sequences $N$ for each dataset split. The generated sequences can be either \textit{all positives} (i.e., satisfying the temporal specification), or \textit{balanced positive and negative} sequences.\footnote{Additionally, low level functions in our code can generate a single positive or negative sequence, for experiments on unbalanced datasets.}
A sequential dataset can have sequences of random length within a $[min, max]$ (inclusive) range, or fixed length $T$ (corresponding to the range $[T, T]$).
The limited depth first procedure is called $N$ times, each time fixing a random target length $min \leq T' \leq max$ and selecting a target label (positive or negative).
In case the sampled sequence violates the desired target, the procedure backtracks, attempting a different randomized subpath along the automaton. In case no path of length $T'$ satisfies the target label, successive attempts are made by lowering the sequence length to $T' - 1$, until the minimum value is reached (at which point the task specification cannot be satisfied).
After each sequence in the dataset is generated, a solution for each transition is randomly sampled from the constraint cache, and variables irrelevant for each timestep are assigned arbitrarily. Each sequence in the generated dataset is labeled with: automaton trace (which transitions $s_i\rightarrow s_j$ were sampled), variable assignments (distinguishing irrelevant ones from those required for the transition), constraint values (distinguishing irrelevant and orphan ones from those affecting the transition), and the sequence label.
In \textbf{incremental-mode}, only a single sequence is sampled. Such sequence is always positive and with a fixed sequence length $T$. Instead of having $N$ different sequences, an incremental dataset is characterized by $T$ different episodes, each containing $N$ samples, which are built by extracting $N$ random assignments from the constraint cache. In this case, the sequence of traversed automaton states are a property of the entire dataset, and can be considered a form of background knowledge (i.e., a curriculum of experiences), instead of a sample-level annotation. Seeding the random generator with different values in incremental-mode will produce different curricula, which however share the property of satisfying the same temporal specification. This can be beneficial when exploring the effect of task ordering in continual learning~\cite{mannelli2024tilting}.

\paragraph{Computational Complexity of Generation.}
LTLZinc relies on multiple non-polynomial steps to generate a dataset coherent with the user-defined specifications.
As time and space complexities for each stage depend on different quantities, we hereby limit our analysis of the generation procedure to an informal discussion.
The conversion between \LTLf and SFA requires bi-exponential time with respect to the formula length and the output automaton has, in the worst case, a number of states exponential with respect to the number of atomic propositions appearing in the formula, however this step is performed only once during generation and meaningful human-defined formulas tend to be short.
The limited depth-first search procedure at the core of sequence sampling has a worst-case running time which is polynomial with respect to the number of potential next states and exponential with respect to the target sequence length, however such case will never be triggered in practice, as it would require an automaton which is fully connected (each state is a successor of every other state) and it is unlikely for an \LTLf formula to be converted to an automaton with such structure. Moreover, the randomized successor sampling would need to consistently select a next state yielding a failed search at each step.
Search is repeated a number of times corresponding to the number of target sequences (corresponding to the dataset size for \textbf{sequential-mode}, and exactly one for \textbf{incremental-mode}).
Each transition of the SFA requires to solve a different constraint satisfaction problem (which is a procedure at least NP-Hard), and, in theory, this process would need to be repeated for each timestep of every sequence.
However, the constraint cache trades-off time for memory requirements. In this way, the number of constraint satisfaction problems to solve, is in the worst case equal to the number of transitions of the automaton (which in practice is much smaller than the theoretical worst case).\footnote{Given the regular structure of transition guards, a straightforward improvement upon the current caching mechanism can be implemented by memorizing partial solutions, and solving smaller CSP's as intersections of previously known solutions. We reserve this improvement to future work.}

\section{Sequential Mode Tasks}\label{sec:seqtasks}
A learning and reasoning agent can exploit a sequential LTLZinc dataset $\mathcal{S} = \mathcal{S}^+ \cup \mathcal{S}^-$, generated from a specification $\langle \mathcal{X}, \mathcal{Y}, \mathcal{C}, \mathcal{F} \rangle$, in multiple ways. We envision three basic task families, covering a large portion of the temporal and relational reasoning spectrum.
For all three, we will consider the following LTLZinc problem as an example:
\begin{align*}
 \mathcal{X}\colon\quad&\mathcal{X}_A = \left\{\mathimg{mnist0}, \mathimg{mnist1}, \mathimg{mnist2}, \mathimg{mnist3}, \mathimg{mnist4}, \mathimg{mnist5}, \mathimg{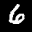}, \mathimg{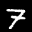}, \mathimg{mnist8}, \mathimg{mnist9}\right\};\\
 &\mathcal{X}_B = \left\{\mathimg{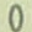}, \mathimg{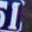}, \mathimg{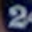}, \mathimg{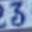}, \mathimg{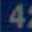}, \mathimg{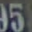}, \mathimg{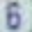}, \mathimg{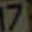}, \mathimg{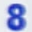}, \mathimg{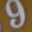}\right\};\\
 &\mathcal{X}_C = \left\{ \mathimg{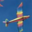}, \mathimg{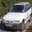}, \mathimg{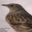}, \mathimg{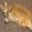}, \mathimg{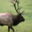}, \mathimg{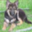}, \mathimg{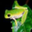}, \mathimg{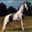}, \mathimg{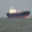}, \mathimg{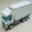}\right\}\\
 \mathcal{Y}\colon\quad&\mathcal{Y}_A, \mathcal{Y}_B = [0,9]; \\
 &\mathcal{Y}_C = \left\{\text{airplane}, \text{automobile}, \text{bird}, \text{cat}, \text{deer}, \text{dog}, \text{frog}, \text{horse}, \text{ship}, \text{truck}\right\}\\
 \mathcal{C}\colon\quad&\texttt{p}(A, B, C): A = 2 \cdot B \vee B = 2 \cdot C;\\
 &\texttt{q}(A, B): \texttt{all\_different}([A,B]);\\
 &\texttt{r}(C): C \in \left\{\text{bird}, \text{cat}, \text{deer}, \text{dog}, \text{frog}, \text{horse}\right\}\\
 \mathcal{F}\colon\quad&\ltlfinally r(C) \wedge ((p(A,B,C) \leftrightarrow \ltlnext q(A,B)) \ltluntil r(C)).
\end{align*}
The example above highlights the use of different datasets (MNIST digits, SVHN~\cite{netzer2011reading}, Cifar-10~\cite{krizhevsky2009learning}, two of which are mapped to the same symbolic labels), and three different kinds of constraints.
Relational constraints $\texttt{p}$ and $\texttt{q}$ link together different symbolic labels, while $\texttt{r}$ is a propositional constraint which limits (for the specific timesteps in which it holds true) the possible values of a single variable.\footnote{A propositional constraint can also describe some semantic feature of single symbolic values (e.g., $\texttt{is\_bird}(X): \texttt{has\_wings}(X) \wedge \texttt{has\_beak}(X)$), enabling concept-based temporal reasoning. Internally LTLZinc collates every non-integer symbolic label to a universe enumeration, allowing lexicographic comparisons also across different datasets.} At the same time, $\texttt{p}$ is an arbitrary logic-arithmetic expression,\footnote{In MiniZinc, enumerations are cast to integers in lexicographic order, therefore the implicit equivalences $\left\{airplane = 0, automobile = 1, \ldots, truck = 9\right\}$ hold. Predicate $\texttt{p}$ can be true, for example, with $Y_A = \text{don't care}, Y_B = 6, Y_C = \text{cat (3)}$.} while $\texttt{q}$ is a \textit{global constraint} from the MiniZinc library, associated with well-known semantics ($A \not = B$).
The automaton corresponding to the temporal specification is shown in Figure~\ref{fig:example-dfa}. From it we can sample positive sequences, such as $s^+ = (\left[\mathimg{mnist3},\mathimg{svhn2}, \mathimg{cifartru}\right], \left[\mathimg{mnist8},\mathimg{svhn4}, \mathimg{cifaraut}\right], \left[\mathimg{mnist2},\mathimg{svhn1}, \mathimg{cifarair}\right], \left[\mathimg{mnist3},\mathimg{svhn2}, \mathimg{cifartru}\right], \left[\mathimg{mnist5},\mathimg{svhn5}, \mathimg{cifarcat}\right])$, corresponding to the constraint trace $(\left[\neg p, \_, \neg r\right], \left[p, \neg q, \neg r\right], \left[p, \neg q, \neg r\right], \left[\neg p, q, \neg r\right], \left[\_, \neg q, r\right])$, or negative sequences, e.g., $s^- = (\left[\mathimg{mnist3},\mathimg{svhn2}, \mathimg{cifartru}\right], \left[\mathimg{mnist8},\mathimg{svhn4}, \mathimg{cifaraut}\right], \left[\mathimg{mnist2},\mathimg{svhn1}, \mathimg{cifarair}\right], \left[\mathimg{mnist0},\mathimg{svhn0}, \mathimg{cifarair}\right], \left[\mathimg{mnist5},\mathimg{svhn5}, \mathimg{cifarcat}\right])$, associated with constraints $(\left[\neg p, \_, \neg r\right], \left[p, \neg q, \neg r\right], \left[p, \neg q, \neg r\right], \left[\_, \neg q, \_\right], \left[\_, \_, \_\right])$.\footnote{Underscores correspond to ``don't care'' truth assignments.}

\begin{figure}
    \centering
    \begin{tikzpicture}[mystate/.style={
          circle, draw, inner sep=6pt 
    },accepting/.style={double distance=3pt, outer sep=1.5pt+\pgflinewidth}]
        \node[mystate, initial] (q1) {$1$};
        \node[mystate, below right=3cm of q1] (q2) {$2$};
        \node[mystate, right=3cm of q2] (q3) {$3$};
        \node[mystate, accepting, below=6cm of q1] (q4) {$4$};
        \node[mystate, below=3cm of q3] (q5) {$5$};

        \path[->]
            (q1) edge[bend left] node[left] {$\neg \texttt{p} \wedge \neg \texttt{q}$} (q2)
            (q1) edge[bend left] node[above] {$\texttt{p} \wedge \neg \texttt{r}$} (q3)
            (q1) edge[bend right] node[left] {$\texttt{r}$} (q4)
            (q2) edge[loop left] node[left] {$\neg \texttt{p} \wedge \neg \texttt{q} \wedge \neg \texttt{r}$} (q2)
            (q2) edge[bend right] node[below] {$\texttt{p} \wedge \neg \texttt{q} \wedge \neg \texttt{r}$} (q3)
            (q2) edge[bend right] node[right] {$\neg \texttt{q} \wedge \texttt{r}$} (q4)
            (q2) edge[bend right] node[above] {$\texttt{q}$} (q5)
            (q3) edge[bend right] node[above] {$\neg \texttt{p} \wedge \texttt{q} \wedge \neg \texttt{r}$} (q2)
            (q3) edge[loop right] node[right] {$\texttt{p} \wedge \texttt{q} \wedge \neg \texttt{r}$} (q3)
            (q3) edge[bend left] node[below] {$\texttt{q} \wedge \texttt{r}$} (q4)
            (q3) edge[bend left] node[right] {$\neg \texttt{q}$} (q5)
            (q4) edge[loop below] node[below] {$\top$} (q4)
            (q5) edge[loop below] node[below] {$\top$} (q5);

    \end{tikzpicture}
    \caption{A symbolic automaton for the \LTLf formula $\ltlfinally \texttt{r} \wedge ((\texttt{p} \leftrightarrow \ltlnext \texttt{q}) \ltluntil \texttt{r})$. As $\left\{\texttt{p}, \texttt{q}, \texttt{r}\right\}$} can only assume values in $\left\{\top, \bot\right\}$, this is in practice a condensed representation of a deterministic finite state automaton with transitions defined over an alphabet of $2^3$ symbols.
    \label{fig:example-dfa}
\end{figure}

\paragraph{Sequence Classification with Relational and Temporal Knowledge.}
Given a dataset of positive and negative sequences $\mathcal{S}$, sequence classification is the problem of learning a binary classifier $f: \mathcal{S} \mapsto \left\{0, 1\right\}$ from annotated examples, such that the input sequence $s$ is mapped to the positive class if and only if $s \models \langle \mathcal{X}, \mathcal{Y}, \mathcal{C}, \mathcal{F} \rangle$. Various degrees of knowledge injection are possible, based on the amount of information from the tuple $\langle \mathcal{X}, \mathcal{Y}, \mathcal{C}, \mathcal{F} \rangle$ and annotations $\langle \mathcal{S}_\mathcal{Y}, \mathcal{S}_\mathcal{C}, \mathcal{L}\rangle$ available to the learning agent, with end-to-end sequence classification being the weakest (only sequence labels $\mathcal{L}$ available) and fine-grained supervision at the relational and temporal levels being the strongest (i.e. the entirety of the LTLZinc problem and annotation tuples available). Additionally, the automaton trace annotated at generation time, can be exploited as an oracular source of knowledge, for agents exploiting state-based representations, such as recurrent neural networks, memory-augmented neural networks or neuro-symbolic automata~\cite{manginas2024nesya}.
Instantiated in the previous example, a sequence classifier would attempt to learn $\mathcal{F}$ as ``for a positive sequence, $\texttt{r}$ is guaranteed to be verified eventually, and $\texttt{p}$ and $\texttt{q}$ will alternate in consecutive timesteps, until $\texttt{r}$ is verified''. In order to achieve this objective, knowledge about $\mathcal{C}$ or $\mathcal{Y}$ might be given, or learned jointly with the sequence label.

\paragraph{Temporally-Distant Supervision.}
The goal of distant supervision is to learn a certain number of classifiers $g_i: \mathcal{X}_i \mapsto \mathcal{Y}_i$, without explicitly providing annotations at the symbolic level $\mathcal{Y}_i$. When instantiated over time, distant supervision translates into learning all the $g_i$, such that their mapping satisfies the LTLZinc problem $\langle\mathcal{X}, \mathcal{Y}, \mathcal{C}, \mathcal{F}\rangle$.
Temporally-distant supervision in its ``purest'' form is an extremely weak learning signal, as multiple variable assignments ($g_i$) must be learned from sequence-level annotations $\mathcal{L}$: for this reason it is possible to also exploit intermediate-level annotations (such as automaton traces, or constraint values).
In the given example, temporally-distant supervision corresponds to learning to classify decimal digits (either with two disjoint mappings for MNIST and SVHN, or employing a joint classifier robust against different perceptual domains) and the ten Cifar-10 classes, by having access to $\mathcal{F}$ and $\mathcal{C}$.
The challenges of distant supervision are greatly amplified by the relational and temporal nature of LTLZinc.
Even over finite domains, a single constraint can be satisfied by a number of assignments exponential with the size of domains, and its nature can be more or less subject to the risk of reasoning shortcuts (e.g., an \texttt{all\_different} constraint allows any permutation of label mappings, as long as images from the same ground truth label can be mapped to the same shortcut label, while a comparison on the lexicographic ordering of labels has a much smaller, but still non-zero, chance of reasoning shortcuts).
Introducing a temporal component, further increases the size of reasoning shortcuts, as every timestep compounds additional shortcuts (in a multiplicative fashion if constraints holding in consecutive timesteps do not overlap).

\paragraph{Constraint Induction Over Time.}
The set of relational constraints $\mathcal{C}$ can be seen as a collection of inducible predicates in an inductive logic programming (ILP) setting. However, while in traditional ILP problems predicate semantics is learned directly from observations about their truth values $\mathcal{S}_\mathcal{C}$, the goal of constraint induction over time is to learn the semantics of predicates in $\mathcal{C}$, given indirect observations on sequence labels $\mathcal{L}$ and image annotations $\mathcal{S}_\mathcal{Y}$. In this setting, traditional ILP examples are provided indirectly by means of temporal knowledge $\mathcal{F}$: the agent has to perform learning over multiple inducibles at different timesteps, aided by an auxiliary (possibly learned as well) selector function, which dynamically associates each example with the correct target.
Grounding a constraint induction over time task in the provided example, the agent has to learn three different binary functions, one for each of $\{\texttt{p}, \texttt{q}, \texttt{r}\}$. Please note that constraints such as $\texttt{p}$ can be significantly harder to learn than $\texttt{q}$ (as it belongs to a library of well-known relations, allowing, for example, brute-force approaches) and $\texttt{r}$ (as it is propositional in nature and the search space is significantly smaller). Difficulty is additionally increased in the case of ambiguities in the temporal behavior (e.g., we are not aware of whether the initial timestep satisfies $r$ or not), as multiple hypotheses need to be considered.
Moreover, as inducible constraints can be observed in a very unbalanced fashion (e.g., because of rare events), the temporal specification can also aid in selecting relevant samples. Observing the automaton corresponding to the provided example (Figure~\ref{fig:example-dfa}), it is clear that constraint $\texttt{r}$ is heavily unbalanced towards negative observations, as seven transitions require it to be negative, three to be positive, and four do not depend on its truth value, and that positive values are always observed when entering state ``4'' (which is a sink, therefore after the first observation of a positive $\texttt{r}$ there is no longer any useful temporal information to exploit).

\section{Incremental Mode Tasks}\label{sec:inctasks}
The most popular experimental settings in continual learning are \textit{incremental} in nature: agents constantly face new concepts, and are expected to take advantage of the entire learning experience, without being exposed to past experiences a second time. This assumption makes it is possible to benchmark various aspects of an agent learning over time, however it also oversimplifies the temporal behavior a learning agent could be exposed to in the real world.
Assuming we can represent a single learning experience at timestep $t$ as a proposition $\varphi^t = \text{``the agent is exposed to experience t''}$, \LTLf (or \LTL in infinite-horizon settings, i.e., life-long learning) can model a traditional continual learning setting, solely by exploiting conjunctions and the next operator to define a chain of experiences: 
$$\mathcal{F} = \bigwedge\limits_{t = 0}^{T} \underbracket{\ltlnext \dots \ltlnext}_{t\text{ times}}\varphi^t.$$
In fact, \LTLf expressivity allows to define behaviors more complex than simple chains of events: experiences can be defined in terms of invariances, eventual guarantees, conditional triggers, and so on. For example, a continual learning agent instantiated in an autonomous driving setting, can expect to observe a day-night domain shift, not just once (as popular setups may assume), but repeated periodically, and at the same time, it can assume that certain warning signs correspond to specific dangers in some future timesteps.
LTLZinc allows to define the semantics of each experience $\varphi^t$ in a way which can be exploited to push continual learning research towards a more realistic setting, without sacrificing control over experimental variables.
Moreover, as assumptions, like the ones exemplified above, can be formalized as LTLZinc problems, they can be injected into the learning agent as exploitable knowledge, which can counteract catastrophic forgetting in a targeted fashion.

\paragraph{Class-Incremental Learning.}
Class-incremental learning, is an instance of multi-class learning, in which the agent is expected to learn disjoint groups of labels at each episode, being able, by the end of training, to recall all the classes.
LTLZinc can formalize CIL by presenting \textbf{disjoint partitions} at each timestep, i.e., by varying the tuple $\langle\mathcal{X}^t, \mathcal{Y}^t\rangle$ dynamically. This can be achieved in two ways, by defining disjoint domains, $\langle\mathcal{X}^0, \mathcal{Y}^0\rangle, \ldots, \langle\mathcal{X}^{T-1}, \mathcal{Y}^{T-1}\rangle$, filtered by a single stream over a unary constraint, or by defining a universe domain $\langle\mathcal{X}, \mathcal{Y}\rangle$, controlled by multiple propositional constraints in the form $\texttt{p}_t = Y \in \mathcal{Y}^t$, such that $\mathcal{Y}^i \cap \mathcal{Y}^j = \varnothing, \forall i \not = j \in \left[0, T\right[$ and $\bigcup\limits_{i = 0}^{T - 1} \mathcal{Y}^i = \mathcal{Y}$.
For instance, the following is an example of a CIL experience on MNIST, where at each timestep images can belong only to either of two digit labels:
\begin{align*}
 \mathcal{X}\colon\quad&\mathcal{X} = \left\{\mathimg{mnist0}, \mathimg{mnist1}, \mathimg{mnist2}, \mathimg{mnist3}, \mathimg{mnist4}, \mathimg{mnist5}, \mathimg{mnist6}, \mathimg{mnist7}, \mathimg{mnist8}, \mathimg{mnist9}\right\}\\
 \mathcal{Y}\colon\quad&\mathcal{Y} = [0,9]\\
 \mathcal{C}\colon\quad&\texttt{p}_0(Y): Y \in \left\{0, 1\right\};\\
 &\texttt{p}_1(Y): Y \in \left\{2, 3\right\};\\
 &\texttt{p}_2(Y): Y \in \left\{4, 5\right\};\\
 &\texttt{p}_3(Y): Y \in \left\{6, 7\right\};\\
 &\texttt{p}_4(Y): Y \in \left\{8, 9\right\}\\
 \mathcal{F}\colon\quad&\texttt{p}_0 \wedge \ltlnext \texttt{p}_1 \wedge \ltlnext\ltlnext \texttt{p}_2 \wedge \ltlnext\ltlnext\ltlnext \texttt{p}_3 \wedge\ltlnext\ltlnext\ltlnext\ltlnext \texttt{p}_4.
\end{align*}
Class-incremental learning can be generalized to a setting we call \textbf{class-continual learning}, by allowing $\mathcal{F}$ to be an arbitrary \LTLf formula and by dropping the requirement $\mathcal{Y}^i \cap \mathcal{Y}^j = \varnothing, \forall i \not = j \in \left[0, T\right[$.

\paragraph{Domain-Incremental Learning and Concept Drift.}
In Domain-incremental learning, the agent learns a single classifier, being presented with perceptual stimuli subject to distribution shift over time. By the end of training, the agent is expected to have good performance regardless of input distribution. LTLZinc can formalize DIL tasks by means of a single dummy constraint $\texttt{p}(Y) = \top$, with different streams linking it to \textbf{disjoint perceptual domains} $\mathcal{X}_i \not = \mathcal{X}_j \forall i\not = j$ over time. \textbf{Domain adaptation} experiments can be encoded by allowing perceptual domains to share some images.
The following example models a digit classifier exposed to three different input distributions (MNIST, SVHN and Cifar-10 interpreted with integer labels).
\begin{align*}
 \mathcal{X}\colon\quad&\mathcal{X}_{MNIST} = \left\{\mathimg{mnist0}, \mathimg{mnist1}, \mathimg{mnist2}, \mathimg{mnist3}, \mathimg{mnist4}, \mathimg{mnist5}, \mathimg{mnist6}, \mathimg{mnist7}, \mathimg{mnist8}, \mathimg{mnist9}\right\};\\
 &\mathcal{X}_{SVHN} = \left\{\mathimg{svhn0}, \mathimg{svhn1}, \mathimg{svhn2}, \mathimg{svhn3}, \mathimg{svhn4}, \mathimg{svhn5}, \mathimg{svhn6}, \mathimg{svhn7}, \mathimg{svhn8}, \mathimg{svhn9}\right\};\\
 &\mathcal{X}_{CIFAR-10} = \left\{ \mathimg{cifarair}, \mathimg{cifaraut}, \mathimg{cifarbir}, \mathimg{cifarcat}, \mathimg{cifardee}, \mathimg{cifardog}, \mathimg{cifarfro}, \mathimg{cifarhor}, \mathimg{cifarshi}, \mathimg{cifartru}\right\}\\
 \mathcal{Y}\colon\quad&\mathcal{Y} = [0, 9]\\
 \mathcal{C}\colon\quad&\texttt{p}(Y): \top\\
 \mathcal{F}\colon\quad&\texttt{p}(Y \mapsto \langle\mathcal{X}_{MNIST}, \mathcal{Y}\rangle) \wedge \ltlnext \texttt{p}(Y\mapsto \langle \mathcal{X}_{SVHN}, \mathcal{Y}\rangle) \wedge \ltlnext\ltlnext \texttt{p}(Y \mapsto \langle \mathcal{X}_{CIFAR-10}, \mathcal{Y} \rangle).
\end{align*}
\textbf{Concept drift} is the phenomenon in which the same perceptual stimulus changes semantics over time. It is straightforward to encode concept drift in LTLZinc, by exploiting streams to map the same $\mathcal{X}$ to different $\mathcal{Y}_i$, for example:
\begin{align*}
 \mathcal{X}\colon\quad&\mathcal{X} = \left\{ \mathimg{cifarair}, \mathimg{cifaraut}, \mathimg{cifarbir}, \mathimg{cifarcat}, \mathimg{cifardee}, \mathimg{cifardog}, \mathimg{cifarfro}, \mathimg{cifarhor}, \mathimg{cifarshi}, \mathimg{cifartru}\right\}\\
 \mathcal{Y}\colon\quad&\mathcal{Y}_0 = [0, 9];\\
 &\mathcal{Y}_1 = \left\{airplane, automobile, bird, cat, deer, dog, frog, horse, ship, truck\right\};\\
 &\mathcal{Y}_2 = \left\{animal, vehicle\right\}\\
 \mathcal{C}\colon\quad&\texttt{p}(Y): \top\\
 \mathcal{F}\colon\quad&\texttt{p}(Y \mapsto \langle\mathcal{X}, \mathcal{Y}_0\rangle) \wedge \ltlnext \texttt{p}(Y\mapsto \langle \mathcal{X}, \mathcal{Y}_1\rangle) \wedge \ltlnext \ltlnext \texttt{p}(Y\mapsto \langle \mathcal{X}, \mathcal{Y}_2\rangle).
\end{align*}

\paragraph{Task-Incremental Learning.}
Task-incremental learning is an instance of multi-task learning, where the agent is exposed to one task at the time and, by the end of the training experience, it is expected to perform well in all of them. Assuming we can describe a single task by means of an invariance constraint (e.g., a relation linking multiple inputs, which is always true when experiencing the given task), a task-incremental experience can be described by guaranteeing that each constraint is true exactly once. For example, the four modulo-10 arithmetic operations on MNIST can be described as the following LTLZinc problem:\footnote{Formally, for a TIL setting we would also need to assume constraints to be disjoint in time, and $\mathcal{F}$ such that $p_{add}$ is also false in each timestep except the first one, $p_{sub}$ is false everywhere except the second timestep, and so on.}
\begin{align*}
 \mathcal{X}\colon\quad&\mathcal{X}_A, \mathcal{X}_B, \mathcal{X}_C = \left\{\mathimg{mnist0}, \mathimg{mnist1}, \mathimg{mnist2}, \mathimg{mnist3}, \mathimg{mnist4}, \mathimg{mnist5}, \mathimg{mnist6}, \mathimg{mnist7}, \mathimg{mnist8}, \mathimg{mnist9}\right\}\\
 \mathcal{Y}\colon\quad&\mathcal{Y}_A, \mathcal{Y}_B, \mathcal{Y}_C = [0,9]\\
 \mathcal{C}\colon\quad&\texttt{add}(A, B, C): A + B \equiv C \mod{10};\\
 &\texttt{sub}(A, B, C): A - B = C \wedge A \geq B;\\
 &\texttt{mul}(A, B, C): A \cdot B \equiv C \mod{10};\\
 &\texttt{div}(A, B, C): A \div B = C \wedge B \not = 0\\
 \mathcal{F}\colon\quad&\texttt{add} \wedge \ltlnext \texttt{sub} \wedge \ltlnext\ltlnext \texttt{mul} \wedge \ltlnext\ltlnext\ltlnext \texttt{div}.
\end{align*}
Like class-incremental problems, we can generalize TIL to \textbf{task-continual learning}, by allowing arbitrary formulas for $\mathcal{F}$.
It is important to note that, in this setting, constraints in $\mathcal{C}$ are \textit{invariance relations}, characterizing the expected input-output behavior, therefore, their labels act as a binary encoding\footnote{If the LTLZinc problem corresponds to a traditional task-incremental setting, like the example above, this encoding will be one-hot, in general multiple tasks may be active at a given time.} of what is known as \textit{task-id} in the continual learning literature, i.e., an oracular descriptor of which task the agent is supposed to solve in the current timestep.
In the LTLZinc generator, streams can optionally be annotated with a direction (input/output). Directions do not affect the generation procedure, however they can provide the agent information about which variables are dependent from the others. This is especially useful in task-incremental settings like the proposed example, where uncertainty can be significantly reduced by noting that values for $\mathcal{Y}_C$ depend on the values of $\mathcal{Y}_A, \mathcal{Y}_B$.

\section{Sequence Classification with LTLZinc}\label{sec:seqexp}
This section presents LTLZinc experiments conducted on six tasks for neuro-symbolic sequence classification over simple perceptual domains and complex relational and temporal knowledge.

\subsection{Proposed Tasks}\label{sec:proposed-seq}
We experiment on six sequence classification tasks where positive sequences satisfy the temporal specification, while negative sequences violate it. Note that this means that negative sequences may still satisfy the provided constraints, albeit at the ``wrong'' time. Every task is proposed in two variants, one characterized by short sequences (random length in $[10, 20]$) and the other using long sequences (random length in $[50, 100]$). Both versions have datasets of 400 sequences, split into 320 training, 40 validation and 40 test sequences.
Images are sampled from the MNIST digits and the Fashion MNIST datasets.
Images for each split are sampled according to the original image classification dataset splits (i.e., for MNIST and FMNIST datasets, we sample training and validation set from the original 60,000 training images and the test set from the 10,000 test images).
Each sequence is associated with the following annotations: image labels for each time step, constraint validity for each time step, automaton state trace, sequence label. The automaton used to generate each task is also available as background knowledge.

\paragraph{Task 1.} ``It will always be the case that $Y$ is lexicographically smaller than $Z$ if and only if two steps later $V$, $W$ and $Z$ belong to the same class''. This task focuses on comparisons and short term memory capabilities of the agent.
\begin{align*}
 \mathcal{X}\colon\quad&\mathcal{X}_Y, \mathcal{X}_Z = \{\mathimg{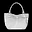}, \mathimg{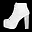}, \mathimg{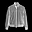}, \mathimg{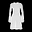}, \mathimg{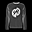}, \mathimg{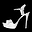}, \mathimg{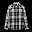}, \mathimg{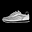}, \mathimg{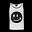}, \mathimg{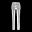}\};\\
 &\mathcal{X}_V, \mathcal{X}_W, \mathcal{X}_X = \{\mathimg{fmnist5}, \mathimg{fmnist6}, \mathimg{fmnist7}, \mathimg{fmnist8}, \mathimg{fmnist9}\}\\
 \mathcal{Y}\colon\quad&\mathcal{Y}_Y, \mathcal{Y}_Z = \left\{bag, boot, coat, dress, pullover, sandal, shirt, sneaker, top, trouser\right\};\\
 &\mathcal{Y}_V, \mathcal{Y}_W, \mathcal{Y}_X = \left\{sandal, shirt, sneaker, top, trouser\right\}\\
 \mathcal{C}\colon\quad&\texttt{p}(Y, Z): Y < Z;\\
 &\texttt{q}(V, W, X): \texttt{all\_equal}([V,W,X])\\
 \mathcal{F}\colon\quad&\ltlglobally (\texttt{p}(Y,Z) \leftrightarrow \ltlnext\ltlnext \texttt{q}(V,W,X)).
 \end{align*}
The corresponding automaton has 8 states. Self-loops are avoided with a linear decay with rate $0.1$.\footnote{This means that the first time depth-first search encounters a loop state as potential successor, it will discard it from the successors list with probability $1-(0\cdot0.1)$, the second time with probability $1-(1\cdot 0.1)$, and so on. After this pruning step, a successor is sampled uniformly from the list.}

\paragraph{Task 2.} ``It will always be true that, if $Y$ is lexicographically smaller than $Z$ for three consecutive timesteps, then $V$, $W$ and $X$ will share the same labels in the successive step''. This task employes the same domains and constraints of Task 1, but applies them to a different temporal specification, requiring a longer memory window.
\begin{align*}
 \mathcal{X}\colon\quad&\mathcal{X}_Y, \mathcal{X}_Z = \{\mathimg{fmnist0}, \mathimg{fmnist1}, \mathimg{fmnist2}, \mathimg{fmnist3}, \mathimg{fmnist4}, \mathimg{fmnist5}, \mathimg{fmnist6}, \mathimg{fmnist7}, \mathimg{fmnist8}, \mathimg{fmnist9}\};\\
 &\mathcal{X}_V, \mathcal{X}_W, \mathcal{X}_X = \{\mathimg{fmnist5}, \mathimg{fmnist6}, \mathimg{fmnist7}, \mathimg{fmnist8}, \mathimg{fmnist9}\}\\
 \mathcal{Y}\colon\quad&\mathcal{Y}_Y, \mathcal{Y}_Z = \left\{bag, boot, coat, dress, pullover, sandal, shirt, sneaker, top, trouser\right\};\\
 &\mathcal{Y}_V, \mathcal{Y}_W, \mathcal{Y}_X = \left\{sandal, shirt, sneaker, top, trouser\right\}\\
 \mathcal{C}\colon\quad&\texttt{p}(Y, Z): Y < Z;\\
 &\texttt{q}(V, W, X): \texttt{all\_equal}([V,W,X])\\
 \mathcal{F}\colon\quad&\ltlglobally ((\texttt{p}(Y, Z) \wedge \ltlnext \texttt{p}(Y, Z) \wedge \ltlnext \ltlnext \texttt{p}(Y, Z) \rightarrow\ltlnext\ltlnext\ltlnext \texttt{q}(V, W, X)).
 \end{align*}
The corresponding automaton has 5 states. Self-loops are avoided with a linear decay with rate $0.1$, while accepting and non-accepting sink states are avoided with a linear decay with rate $0.01$.

\paragraph{Task 3.} ``$X$ will be less than the sum $Y + Z$, up to the point in which the next state has $X$, $Y$ and $Z$ belonging to different classes''. This task requires arithmetic comparisons and is characterized by a challenging temporal behavior (there may be positive sequences satisfying trivially the second constraint twice and never the first).
\begin{align*}
 \mathcal{X}\colon\quad&\mathcal{X}_X, \mathcal{X}_Y, \mathcal{X}_Z = \left\{\mathimg{mnist0}, \mathimg{mnist1}, \mathimg{mnist2}, \mathimg{mnist3}, \mathimg{mnist4}, \mathimg{mnist5}, \mathimg{mnist6}, \mathimg{mnist7}, \mathimg{mnist8}, \mathimg{mnist9}\right\}\\
 \mathcal{Y}\colon\quad&\mathcal{Y}_X, \mathcal{Y}_Y, \mathcal{Y}_Z = [0, 9]\\
 \mathcal{C}\colon\quad&\texttt{p}(X, Y, Z): \texttt{all\_different}([X,Y,Z]);\\
 &\texttt{q}(X, Y, Z): X < Y + Z\\
 \mathcal{F}\colon\quad&\ltlfinally \texttt{p}(X, Y, Z) \wedge (\texttt{q}(X, Y, Z) \ltluntil \ltlnext \texttt{p}(X, Y, Z)).
 \end{align*}
The corresponding automaton has 5 states. Self-loops are avoided with a linear decay with rate $0.1$, while accepting and non-accepting sink states are avoided with a linear decay with rate $0.01$.

\paragraph{Task 4.} ``$X$ will be less than the sum $Y + Z$, up to the point in which the next state has $X$, $Y$ and $Z$ belonging to different classes''. This task is the same as Task 3, with different perceptual domains (MNIST and FMNIST interpreted as integer classes).
\begin{align*}
 \mathcal{X}\colon\quad&\mathcal{X}_X = \left\{\mathimg{mnist0}, \mathimg{mnist1}, \mathimg{mnist2}, \mathimg{mnist3}, \mathimg{mnist4}, \mathimg{mnist5}, \mathimg{mnist6}, \mathimg{mnist7}, \mathimg{mnist8}, \mathimg{mnist9}\right\};\\
 &\mathcal{X}_Y, \mathcal{X}_Z = \left\{\mathimg{fmnist0}, \mathimg{fmnist1}, \mathimg{fmnist2}, \mathimg{fmnist3}, \mathimg{fmnist4}, \mathimg{fmnist5}, \mathimg{fmnist6}, \mathimg{fmnist7}, \mathimg{fmnist8}, \mathimg{fmnist9}\right\}\\
 \mathcal{Y}\colon\quad&\mathcal{Y}_X, \mathcal{Y}_Y, \mathcal{Y}_Z = [0, 9]\\
 \mathcal{C}\colon\quad&\texttt{p}(X, Y, Z): \texttt{all\_different}([X,Y,Z]);\\
 &\texttt{q}(X, Y, Z): X < Y + Z\\
 \mathcal{F}\colon\quad&\ltlfinally \texttt{p}(X, Y, Z) \wedge (\texttt{q}(X, Y, Z) \ltluntil \ltlnext \texttt{p}(X, Y, Z)).
 \end{align*}
The corresponding automaton has 5 states. Self-loops are avoided with a linear decay with rate $0.1$, while accepting and non-accepting sink states are avoided with a linear decay with rate $0.01$.

\paragraph{Task 5.} ``$W + X = Y + Z$ is satisfied every other timestep''. This task is characterized by arithmetic reasoning over a simple temporal specification.
\begin{align*}
 \mathcal{X}\colon\quad&\mathcal{X}_W, \mathcal{X}_X, \mathcal{X}_Y, \mathcal{X}_Z = \left\{\mathimg{mnist0}, \mathimg{mnist1}, \mathimg{mnist2}, \mathimg{mnist3}, \mathimg{mnist4}, \mathimg{mnist5}, \mathimg{mnist6}, \mathimg{mnist7}, \mathimg{mnist8}, \mathimg{mnist9}\right\}\\
 \mathcal{Y}\colon\quad&\mathcal{Y}_W, \mathcal{Y}_X, \mathcal{Y}_Y, \mathcal{Y}_Z = [0, 9]\\
 \mathcal{C}\colon\quad&\texttt{p}(W, X, Y, Z): W + X = Y + Z\\
 \mathcal{F}\colon\quad&\ltlglobally (\texttt{p}(W,X,Y,Z) \leftrightarrow \ltlweaknext \neg \texttt{p}(W,X,Y,Z)).
 \end{align*}
The corresponding automaton has 4 states. Self-loops are avoided with a linear decay with rate $0.1$, while accepting and non-accepting sink states are avoided with a linear decay with rate $0.01$.

\paragraph{Task 6.} ``$X + Y = Z$ and $X + Y = 2\cdot Z$ alternate each other''. This task is characterized by arithmetic reasoning over a simple temporal specification.
\begin{align*}
 \mathcal{X}\colon\quad&\mathcal{X}_X, \mathcal{X}_Y, \mathcal{X}_Z = \left\{\mathimg{mnist0}, \mathimg{mnist1}, \mathimg{mnist2}, \mathimg{mnist3}, \mathimg{mnist4}, \mathimg{mnist5}, \mathimg{mnist6}, \mathimg{mnist7}, \mathimg{mnist8}, \mathimg{mnist9}\right\}\\
 \mathcal{Y}\colon\quad&\mathcal{Y}_X, \mathcal{Y}_Y, \mathcal{Y}_Z = [0, 9]\\
 \mathcal{C}\colon\quad&\texttt{p}(X, Y, Z): X + Y = Z;\\
 &\texttt{q}(X, Y, Z): X + Y = 2 \cdot Z\\
 \mathcal{F}\colon\quad&\ltlglobally (\texttt{p}(X,Y,Z) \leftrightarrow \ltlweaknext \neg \texttt{q}(X,Y,Z)).
 \end{align*}
The corresponding automaton has 4 states. Self-loops are avoided with a linear decay with rate $0.1$, while accepting and non-accepting sink states are avoided with a linear decay with rate $0.01$.

\subsection{Methodology}
\label{sec:setup}

\begin{figure}
    \centering
    \includegraphics[width=0.33\linewidth]{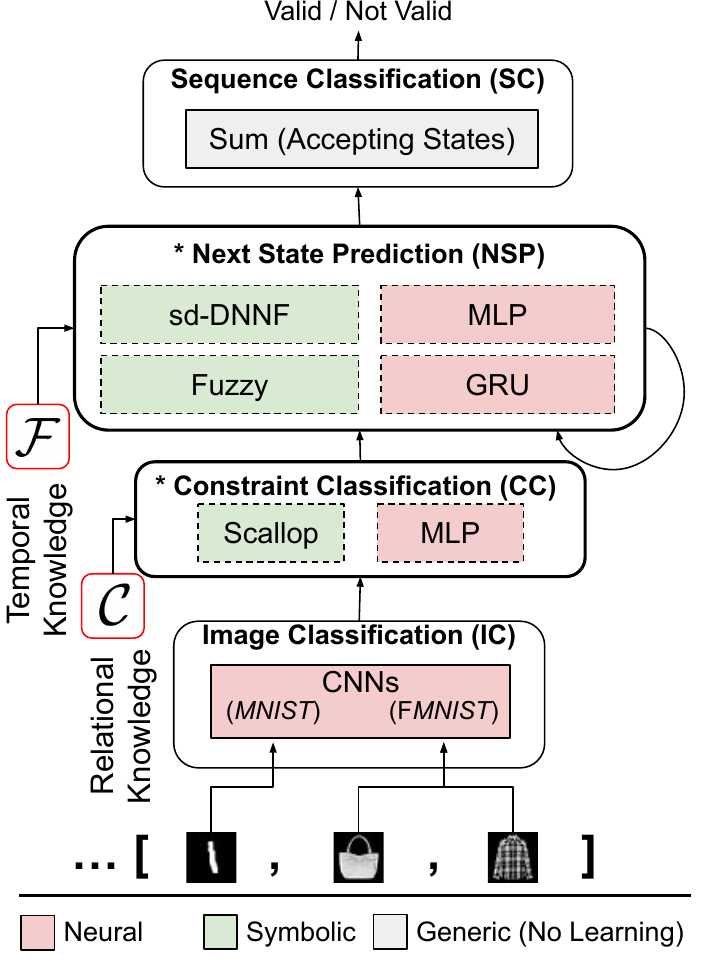}
    \caption{Stages of our architecture for sequence classification experiments. We implemented {\sc cc} and {\sc nsp} in multiple ways (green: symbolic, reddish: neural).}
    \label{fig:pipeline}
\end{figure}

We model the whole sequence classification task by a multi-stage pipeline composed of the following sub-tasks, also sketched in Figure~\ref{fig:pipeline}: ({\sc\small ic}) image classification, mapping data from each $\mathcal{X}_i$ to the corresponding $\mathcal{Y}_i$; ({\sc\small cc}) constraint classification, leveraging ``relational'' knowledge; ({\sc\small nsp}) next state prediction, leveraging ``temporal'' knowledge; ({\sc\small sc}) sequence classification, i.e., the binary classification problem.

\paragraph{{\sc (ic)} Image Classification.} The first stage corresponds to a traditional neural-network-based image classification task, estimating the probability of the symbol/class assignment $[y_{0}^{t}, \ldots, y_{N-1}^{t}]$, i.e., $P_{\text{\sc\tiny ic}}([y_{0}^{t}, \ldots, y_{N-1}^{t}] \mid [x_{0}^{t}, \ldots, x_{N-1}^{t}])$. 
Since the proposed LTLZinc tasks consist of perceptually simple images, our implementation is based on a small convolutional architecture (details in Appendix~\ref{app:backbone}), for each perceptual domain (e.g., for Task 4 of Section~\ref{sec:seqtasks}, we have two instances of the same architecture: one to predict symbol $X$ from the MNIST domain and another one for symbols $Y, Z$ belonging to the FMNIST domain, see Figure~\ref{fig:pipeline}).

\paragraph{{\sc (cc)} Constraint Classification.} The set of all the image classes predicted by the \textsc{ic} module, is mapped to $|\mathcal{C}|$ validity values, each of them indicated with $\beta_i$, one for each of the constraints defined in the relational knowledge $\mathcal{C}$. This stage estimates $P_{\text{\sc\tiny cc}}([\beta_{0}^{t}, \ldots, \beta_{|\mathcal{C}| - 1}^{t}] \mid [y_{0}^{t}, \ldots, y_{N-1}^{t}])$. As relational knowledge is explicitly provided in our setting, this map can be implemented with any knowledge injection technique. 
Our experiments focus on tasks characterized by constraints which can be expressed in Datalog.
Hence, we chose Scallop~\cite{li2023scallop}, a neuro-symbolic engine capable of probabilistic reasoning over Datalog programs, supporting inference over multiple provenance semirings~\cite{green2007provenance},\footnote{In initial explorations, the default top-$k$ proofs provenance ($k=1$) resulted in the best trade-off between inference time and accuracy.} We also explored the effect of exact probabilistic inference by means of ProbLog~\cite{de2007problog} programs.
Scallop and ProbLog programs are differentiable end-to-end, however they do not possess trainable parameters. To increase module flexibility, we augment our architecture with a set of additional learnable calibration parameters, which independently apply temperature rescaling to both input and output probabilities, each with their own independent weight. 
Learning takes place jointly with the \textsc{ic} module, and it allows us to control both the entropy of the distributions and the confidence of prediction, while preserving the argmax.
We compare symbolic methods with a multi-layer perceptron equipped with 64 hidden neurons.

\paragraph{{\sc (nsp)} Next State Prediction.} The temporal reasoning component is rooted on the definition of a discrete space of $M$ elements, each consisting of a state of the SFA equivalent to the temporal specification $\mathcal{F}$, based on the observed validity values of relational constraints, $\beta_i^t$'s. 
In this way, the temporal reasoning task boils down to a next-state prediction problem, i.e., a recurrent classification problem where $\alpha^t \in [0, M-1]$ is the predicted class/next-state, with probability $P_{\text{\sc\tiny nsp}}(\alpha^{t} \mid [\beta_{0}^{t}, \ldots, \beta_{\mathcal{V}-1}^{t}], \alpha^{t-1})$.\footnote{Here and in the following, $t = -1$ is the initial instant, where the automaton of the \textsc{nsp} module is initialized to state ``$0$''.} 
We compare two learning-based approaches (a multi-layer perceptron, MLP, and a gated recurrent unit, GRU-L, both with 64 hidden/state neurons, respectively), and two symbolic approaches. 
GRUs are augmented with a simple encoder-decoder, to convert the continuous hidden state into $M$ discrete classes.
For the symbolic approaches, we encode the ground truth finite state automaton, following the works of~\cite{umili2023grounding} and~\cite{manginas2024nesya}. 
Such automaton is a set of propositional logic formulas as $\textsc{\small prev\_state} \wedge \textsc{\small trans\_label} \rightarrow \textsc{\small next\_state}$, encoded by logic tensor networks~\cite{badreddine2022logic} (here named Fuzzy) and sd-DNNF~\cite{darwiche2002knowledge}, respectively.
As in the \textsc{cc} module, automata-based \textsc{nsp}'s predictions are temperature-calibrated by learnable parameters.

\paragraph{{\sc (sc)} Sequence Classification.} The final output $\alpha^{T-1}$ of {\sc nsp} directly encodes the final state of the automaton. Hence, we can perform sequence classification in closed form: $P_{\text{\sc\tiny sc}}(f(\mathcal{S}) = 1 \mid \alpha^{T-1}) = \sum\limits_{s \in \text{\sc\tiny accepting}} P(\alpha^{T-1} = s)$. 

\paragraph{Training the Pipeline.}
The four stages are combined into: 
\begin{equation}
\nonumber
    P(f(\mathcal{S}) = 1) = \left( \prod_{t=0}^{T-1} P^t_{\text{\sc\tiny ic}} P^t_{\text{\sc\tiny cc}} P^t_{\text{\sc\tiny nsp}} \right) P_{\text{sc}}(f(\mathcal{S}) = 1 \mid \alpha^{T-1}),
\end{equation}
where $P^t_{\cdot}$ is the shorthand notation for the already introduced probabilities with arguments at time $t$.
We train our multi-stage architectures with four loss functions, each weighted by a $\lambda_{\bullet}$ hyper-parameter. \textsc{ic} and \textsc{nsp} (the latter conditioned on the previous state) exploit a categorical cross-entropy loss, while \textsc{cc} and \textsc{sc} a binary cross-entropy loss.
Preliminary experiments demonstrated that training diverges due to extremely low initial confidence in image classification,\footnote{This is especially true when the \textsc{cc} module is a Scallop/ProbLog program, as at the beginning of training, misclassified images lead to confidently predicting violated constraints most of the time.} making the optimizer unable to converge, except in very simple tasks. This behavior is well known in the literature~\cite{manhaeve2021approximate,van2024independence,maene2024hardness}. 
To overcome this issue, we bootstrap the \textsc{ic} module with a $1$-epoch pre-training phase, using the \textsc{ic}-loss only, to ensure a good starting image classification.

\subsection{Analysis of Neuro-Symbolic Stages}
The purpose of the modular architecture shown in Figure~\ref{fig:pipeline} is to investigate the effect of interactions betweeen different reasoning methods. We will now briefly discuss the expected behavior of neuro-symbolic combinations (i.e., neural \textsc{ic}, symbolic \textsc{cc} and symbolic \textsc{nsp}), in light of their semantics.
Our architecture is a feed-forward pipeline of stages with a unified input-output interface: each stage takes continuous inputs in the range $[0, 1]$ (with the exception of \textsc{ic}), and returns outputs in the same range.
Modules with different semantics can be mixed and matched (e.g., neural \textsc{ic}, Scallop \textsc{cc} and Fuzzy \textsc{nsp}) by means of this unified interface.
It is however important to note that the correctness of each module is based on specific assumptions, in particular, many probabilistic systems rely on input variables which are independent~\cite{van2024independence} to compute correct output probabilities. Specifically the \textsc{nsp} probabilistic module considered here (sd-DNNF) does make this assumption.

We shall now consider a numerical example from the task of Section~\ref{sec:seqtasks} to demonstrate our NeSy infrastructure. Consider the following probability distributions computed by the neural component (\textsc{ic}) of the
system for the input $(\mathimg{mnist2}, \mathimg{svhn1}, \mathimg{cifarhor})$:
$$
\begin{aligned}
P_{A}(\mathcal{Y}_A \mid \mathimg{mnist2}) &= \mathrm{Categorical}(
[0.05, 
0.00, 
0.50, 
0.00, 
0.30, 
0.00, 
0.10, 
0.05, 
0.00, 
0.00 
])
 \\
P_{B}(\mathcal{Y}_B \mid \mathimg{svhn1}) &= \mathrm{Categorical}(
[0.00, 
0.80, 
0.00, 
0.10, 
0.10, 
0.00, 
0.00, 
0.00, 
0.00, 
0.00 
]) \\
P_{C}(\mathcal{Y}_C \mid \mathimg{cifarhor}) &= \mathrm{Categorical}(
\underbrace{
[0.15,
0.00, 
0.00, 
0.00, 
0.05, 
0.00, 
0.00, 
0.80, 
0.00, 
0.00]
}_{\text{airplane, automobile, bird, cat, deer, dog, frog, horse, ship, truck}}
).
\end{aligned}
$$
The domains of the variables $(\mathcal{Y}_A, \mathcal{Y}_B, \mathcal{Y}_C)$ are given in Section \ref{sec:seqtasks}: the first two variables simply take values from $[0, 9]$, while the domain of the third is an enumeration (given below the vector of probabilities), which can also be mapped to $[0, 9]$.
Consider now the constraint $\texttt{p}(A, B, C): A = 2 \cdot B \vee \ B = 2 \cdot C$ from the example in Section \ref{sec:seqtasks}. 
Its probability can be computed via:
$$
\begin{aligned}
\mathrm{Pr}(\texttt{p}(A, B, C)) &= \sum_{A, B, C} 
    P_{a}(\mathcal{Y}_A = A \mid \mathimg{mnist2}) 
    P_{b}(\mathcal{Y}_B = B \mid \mathimg{svhn1}) 
    P_{c}(\mathcal{Y}_C = C \mid \mathimg{cifarhor}) 
    \llbracket \texttt{p}(A, B, C) = \top \rrbracket \\
        &= \boxed{
            P_A(\mathcal{Y}_A = 2 \mid \mathimg{mnist2}) 
            \cdot  P_B(\mathcal{Y}_B = 1 \mid \mathimg{svhn2}) 
            \cdot P_C(\mathcal{Y}_c = \mathrm{horse} \mid  \mathimg{cifarhor})} \\
        &+  P_A(\mathcal{Y}_A = 6 \mid \mathimg{mnist2}) \cdot  P_B(\mathcal{Y}_B = 3 \mid \mathimg{svhn2}) \cdot P_C(\mathcal{Y}_C = \mathrm{horse} \mid \mathimg{cifarhor})
        + \dots \\
        &= \boxed{0.5 \cdot 0.8 \cdot 0.8} + 
           0.1 \cdot 0.1 \cdot 0.8 + \dots \\
        &= \boxed{0.32} + 0.09 + \dots \\
        &= 0.41
\end{aligned}
$$
where $\llbracket \cdot \rrbracket$ is an indicator function, which filters assignments based on whether they satisfy the constraint. The computation above bears strong 
resemblance to that of weighted model counting, which is exploited by various 
neuro-symbolic systems.
Since these possible worlds for complex constraints can be exponential in the number 
of input variables, involved techniques have been developed to answer such queries effectively, either in an exact fashion \cite{darwiche2002knowledge, de2007problog} (our ProbLog \textsc{CC} module) or by only considering high probability proofs \cite{manhaeve2021approximate,li2023scallop} (our Scallop \textsc{cc} module). 
The boxed term above is the most possible world for the query $\texttt{p}(A, B, C)$ with a probability mass of $0.32$. 
Similar computations hold for $\texttt{q}(A, B) : \texttt{all\_different}([A, B])$ and $\texttt{r}(C): C \in \left\{\text{bird}, \text{cat}, \text{deer}, \text{dog}, \text{frog}, \text{horse}\right\}$. To summarize, the following are the probabilities for our example, as the sum of the most possible world (boxed) and the residual 
given by lower probability worlds:
\begin{align*}
    \mathrm{Pr}(\texttt{p}(A,B,C)) &= \boxed{0.32} + 0.09 = 0.41\\
    \mathrm{Pr}(\texttt{q}(A,B)) \quad &= \boxed{0.4} + 0.57 = 0.97\\
    \mathrm{Pr}(\texttt{r}(C)) \quad\quad &= \boxed{0.8} + 0.05 = 0.85.
\end{align*}
It can be noted how the approximation error of considering only the most probable possible
world (Scallop \textsc{cc} module) instead of all the possible worlds (ProbLog \textsc{cc} module) can be significant.

We now turn to the issue of independence. For our example, $\texttt{p}(A,B,C)$ and $\texttt{q}(A, B)$ are not independent since they share variables and possible worlds. Similarly, for $\texttt{p}(A, B, C)$ and $\texttt{r}(C)$. However 
$\texttt{q}(A, B)$ and $\texttt{r}(C)$ are independent because they share no variables.
Let us now consider the interaction between the probabilities of the \textsc{cc} module and the probabilities computed via the \textsc{nsp} module, on two transition examples from the automaton of Figure~\ref{fig:example-dfa}. The transition $[2\rightarrow4]: \neg \texttt{q}(A, B) \wedge \texttt{r}(C)$ corresponds to the conjunction of independent events, while constraints appearing in transition $[1\rightarrow2]: \neg \texttt{p}(A,B,C) \wedge \neg \texttt{q}(A, B)$ are dependent. Note that, from the perspective of the \textsc{nsp} module, the proof tree is lost and that \texttt{p}, \texttt{q} and \texttt{r} are now input variables.
Let the \textsc{nsp} module be implemented via an exact probabilistic implementation (sd-DNNF): both transitions are computed as products of probabilities.
We have $\mathrm{Pr}([2\rightarrow4]) = (1 - \mathrm{Pr}(\texttt{q}(A, B))) \cdot \mathrm{Pr}(\texttt{r}(C))$, corresponding to $0.48$ if the \textsc{cc} module yielded only the top proof probability, and $0.0255$ in the case of exact probabilities. As these constraints are independent, these results correspond to those obtained by weighted model counting if we had the proof tree available (i.e., if the \textsc{cc} and \textsc{nsp} module were a single module with access to input variables $A, B, C$).
For the second transition, the \textsc{nsp} module computes probabilities in the same fashion: $\mathrm{Pr}([1 \rightarrow 2]) = (1 - \mathrm{Pr}(\texttt{p}(A, B, C)) \cdot (1 - \mathrm{Pr}(\texttt{q}(A, B)))$, returning $0.408$ (top proof) or $0.0177$ (exact). However, a monolithic probabilistic implementation would not ignore the
dependence between $\texttt{p(A, B, C)}$ and $\texttt{q(A, B)}$ and would, in exact mode, compute the probability as $0.03 \neq 0.0177$.

The approximation error, caused by the violation of the independence assumption at the \textsc{nsp} level, compounds with the one caused by selecting only the top proof within the \textsc{cc} module. In this work we mitigate this error in a data-driven fashion by means of temperature calibration and intermediate level supervisions.
A natural alternative would be to implement a monolithic \textsc{cc-nsp} module, capable of computing transition probabilities from image labels in an end-to-end fashion. This approach however introduces additional scalability and optimization challenges which we reserve to address in future work.

\subsection{Results}
\begin{table}
    \centering
    \resizebox{\textwidth}{!}{
    \begin{tabular}{cccccccc}
        \toprule
        \multirow{2}{*}{Task} & \multirow{2}{*}{Category} & \multirow{2}{*}{\shortstack[c]{Best model\\(best epoch)}} & \multirow{2}{*}{\shortstack[c]{Average\\Accuracy}\raisebox{1ex}{\:$\uparrow$}} & \multirow{2}{*}{\shortstack[c]{Label\\Accuracy}\raisebox{1ex}{\:$\uparrow$}} & \multirow{2}{*}{\shortstack[c]{Constraint\\Accuracy}\raisebox{1ex}{\:$\uparrow$}} & \multirow{2}{*}{\shortstack[c]{Successor\\Accuracy}\raisebox{1ex}{\:$\uparrow$}} & \multirow{2}{*}{\shortstack[c]{Sequence\\Accuracy}\raisebox{1ex}{\:$\uparrow$}}\\
        & & & & & &\\
        \midrule
                \multirow{6}{*}{\shortstack[c]{Task 1}} & Sym. (exact)-Sym. & ProbLog sd-DNNF (49) & $0.69 \pm 0.03$ & $0.73 \pm 0.09$ & $\textbf{0.93} \pm 0.01$ & $0.61 \pm 0.02$ & $0.50 \pm 0.00$\\
                 & Sym. (exact)-Neu. & ProbLog GRU (47) & $\textbf{0.85} \pm 0.00$ & $\textbf{0.85} \pm 0.00$ & $0.92 \pm 0.00$ & $\textbf{0.72} \pm 0.02$ & $\textbf{0.93} \pm 0.00$\\
                 & Sym. (top-1)-Sym. & Scallop Fuzzy (49) & $0.70 \pm 0.01$ & $0.82 \pm 0.06$ & $0.92 \pm 0.01$ & $0.56 \pm 0.00$ & $0.50 \pm 0.00$\\
                 & Sym. (top-1)-Neu. & Scallop MLP (46) & $0.84 \pm 0.01$ & $\textbf{0.85} \pm 0.01$ & $0.90 \pm 0.00$ & $0.70 \pm 0.03$ & $0.92 \pm 0.01$\\
                 & Neu.-Sym. & MLP Fuzzy (48) & $0.70 \pm 0.01$ & $\textbf{0.85} \pm 0.01$ & $0.90 \pm 0.01$ & $0.55 \pm 0.02$ & $0.50 \pm 0.00$\\
                 & Neu.-Neu. & MLP MLP (42) & $0.82 \pm 0.04$ & $0.81 \pm 0.06$ & $0.88 \pm 0.03$ & $0.69 \pm 0.05$ & $0.91 \pm 0.01$\\
                \hdashline
                \multirow{6}{*}{\shortstack[c]{Task 2}} & Sym. (exact)-Sym. & ProbLog Fuzzy (49) & $0.79 \pm 0.03$ & $0.84 \pm 0.06$ & $0.92 \pm 0.02$ & $0.66 \pm 0.01$ & $0.72 \pm 0.04$\\
                 & Sym. (exact)-Neu. & ProbLog GRU (30) & $\textbf{0.89} \pm 0.01$ & $0.87 \pm 0.00$ & $0.91 \pm 0.00$ & $0.83 \pm 0.02$ & $\textbf{0.96} \pm 0.02$\\
                 & Sym. (top-1)-Sym. & Scallop sd-DNNF (49) & $0.79 \pm 0.01$ & $\textbf{0.90} \pm 0.00$ & $\textbf{0.93} \pm 0.00$ & $0.63 \pm 0.02$ & $0.71 \pm 0.01$\\
                 & Sym. (top-1)-Neu. & Scallop MLP (46) & $0.88 \pm 0.00$ & $0.87 \pm 0.00$ & $0.92 \pm 0.00$ & $\textbf{0.85} \pm 0.00$ & $0.88 \pm 0.00$\\
                 & Neu.-Sym. & MLP Fuzzy (46) & $0.79 \pm 0.01$ & $0.86 \pm 0.01$ & $0.88 \pm 0.01$ & $0.67 \pm 0.01$ & $0.74 \pm 0.03$\\
                 & Neu.-Neu. & MLP GRU (47) & $0.86 \pm 0.01$ & $0.86 \pm 0.01$ & $0.90 \pm 0.01$ & $0.81 \pm 0.01$ & $0.88 \pm 0.07$\\
                \hdashline
                \multirow{6}{*}{\shortstack[c]{Task 3}} & Sym. (exact)-Sym. & ProbLog Fuzzy (48) & $\textbf{0.96} \pm 0.00$ & $0.97 \pm 0.00$ & $0.97 \pm 0.00$ & $0.94 \pm 0.00$ & $\textbf{0.95} \pm 0.00$\\
                 & Sym. (exact)-Neu. & ProbLog MLP (42) & $0.74 \pm 0.01$ & $\textbf{0.99} \pm 0.00$ & $\textbf{0.99} \pm 0.00$ & $0.56 \pm 0.03$ & $0.41 \pm 0.01$\\
                 & Sym. (top-1)-Sym. & Scallop sd-DNNF (43) & $\textbf{0.96} \pm 0.00$ & $0.96 \pm 0.00$ & $0.97 \pm 0.01$ & $\textbf{0.95} \pm 0.00$ & $\textbf{0.95} \pm 0.00$\\
                 & Sym. (top-1)-Neu. & Scallop MLP (48) & $0.75 \pm 0.01$ & $0.98 \pm 0.00$ & $0.98 \pm 0.00$ & $0.59 \pm 0.02$ & $0.45 \pm 0.05$\\
                 & Neu.-Sym. & MLP Fuzzy (29) & $0.78 \pm 0.01$ & $0.93 \pm 0.02$ & $0.79 \pm 0.01$ & $0.71 \pm 0.03$ & $0.69 \pm 0.03$\\
                 & Neu.-Neu. & MLP MLP (47) & $0.72 \pm 0.01$ & $\textbf{0.99} \pm 0.00$ & $0.91 \pm 0.01$ & $0.54 \pm 0.02$ & $0.45 \pm 0.05$\\
                \hdashline
                \multirow{6}{*}{\shortstack[c]{Task 4}} & Sym. (exact)-Sym. & ProbLog sd-DNNF (37) & $0.87 \pm 0.02$ & $0.86 \pm 0.01$ & $0.89 \pm 0.01$ & $0.88 \pm 0.04$ & $0.87 \pm 0.04$\\
                 & Sym. (exact)-Neu. & ProbLog MLP (46) & $0.72 \pm 0.02$ & $\textbf{0.91} \pm 0.00$ & $\textbf{0.93} \pm 0.00$ & $0.58 \pm 0.01$ & $0.47 \pm 0.08$\\
                 & Sym. (top-1)-Sym. & Scallop sd-DNNF (50) & $\textbf{0.88} \pm 0.01$ & $0.86 \pm 0.01$ & $0.84 \pm 0.01$ & $\textbf{0.90} \pm 0.02$ & $\textbf{0.90} \pm 0.03$\\
                 & Sym. (top-1)-Neu. & Scallop MLP (41) & $0.72 \pm 0.00$ & $\textbf{0.91} \pm 0.01$ & $0.89 \pm 0.00$ & $0.55 \pm 0.03$ & $0.52 \pm 0.01$\\
                 & Neu.-Sym. & MLP Fuzzy (47) & $0.71 \pm 0.01$ & $0.83 \pm 0.02$ & $0.72 \pm 0.03$ & $0.67 \pm 0.04$ & $0.62 \pm 0.04$\\
                 & Neu.-Neu. & MLP MLP (46) & $0.70 \pm 0.02$ & $\textbf{0.91} \pm 0.01$ & $0.87 \pm 0.01$ & $0.55 \pm 0.01$ & $0.47 \pm 0.07$\\
                \hdashline
                \multirow{6}{*}{\shortstack[c]{Task 5}} & Sym. (exact)-Sym. & ProbLog sd-DNNF (19) & $0.53 \pm 0.31$ & $0.40 \pm 0.51$ & $0.71 \pm 0.23$ & $0.41 \pm 0.36$ & $0.59 \pm 0.16$\\
                 & Sym. (exact)-Neu. & ProbLog MLP (39) & $0.96 \pm 0.01$ & $0.98 \pm 0.00$ & $0.96 \pm 0.00$ & $0.94 \pm 0.00$ & $\textbf{0.98} \pm 0.03$\\
                 & Sym. (top-1)-Sym. & Scallop Fuzzy (44) & $0.51 \pm 0.34$ & $0.40 \pm 0.51$ & $0.64 \pm 0.29$ & $0.41 \pm 0.38$ & $0.60 \pm 0.17$\\
                 & Sym. (top-1)-Neu. & Scallop MLP (48) & $\textbf{0.98} \pm 0.00$ & $\textbf{0.99} \pm 0.00$ & $\textbf{0.97} \pm 0.00$ & $\textbf{0.98} \pm 0.00$ & $\textbf{0.98} \pm 0.00$\\
                 & Neu.-Sym. & MLP sd-DNNF (49) & $0.62 \pm 0.01$ & $0.94 \pm 0.02$ & $0.75 \pm 0.02$ & $0.28 \pm 0.01$ & $0.50 \pm 0.00$\\
                 & Neu.-Neu. & MLP MLP (47) & $0.80 \pm 0.03$ & $0.93 \pm 0.04$ & $0.71 \pm 0.04$ & $0.77 \pm 0.01$ & $0.80 \pm 0.03$\\
                \hdashline
                \multirow{6}{*}{\shortstack[c]{Task 6}} & Sym. (exact)-Sym. & ProbLog sd-DNNF (46) & $0.60 \pm 0.00$ & $\textbf{0.99} \pm 0.00$ & $0.53 \pm 0.00$ & $0.38 \pm 0.00$ & $0.50 \pm 0.00$\\
                 & Sym. (exact)-Neu. & ProbLog GRU (49) & $0.62 \pm 0.01$ & $\textbf{0.99} \pm 0.00$ & $0.53 \pm 0.00$ & $0.44 \pm 0.03$ & $0.50 \pm 0.00$\\
                 & Sym. (top-1)-Sym. & Scallop sd-DNNF (50) & $0.74 \pm 0.04$ & $0.97 \pm 0.01$ & $\textbf{0.96} \pm 0.01$ & $0.50 \pm 0.12$ & $0.52 \pm 0.03$\\
                 & Sym. (top-1)-Neu. & Scallop MLP (48) & $\textbf{0.91} \pm 0.03$ & $0.97 \pm 0.02$ & $\textbf{0.96} \pm 0.02$ & $\textbf{0.88} \pm 0.04$ & $\textbf{0.83} \pm 0.04$\\
                 & Neu.-Sym. & MLP Fuzzy (50) & $0.64 \pm 0.06$ & $0.97 \pm 0.01$ & $0.79 \pm 0.17$ & $0.29 \pm 0.09$ & $0.50 \pm 0.00$\\
                 & Neu.-Neu. & MLP MLP (49) & $0.86 \pm 0.03$ & $0.98 \pm 0.00$ & $0.91 \pm 0.01$ & $0.82 \pm 0.03$ & $0.73 \pm 0.06$\\
        \bottomrule
    \end{tabular}
    }
    \caption{Best models test set accuracies (mean $\pm$ std over 3 random seeds, after discarding diverging runs) for short-sequence experiments, grouped by category. Best model (named ``{\sc cc} model {\sc nsp} model'') and epoch selected by Average Accuracy on validation set.}
    \label{tab:sequential-results-short}
\end{table}
Table~\ref{tab:sequential-results-short} summarizes the best results on short sequences for every task. One result immediately observable from the table is that purely neural approaches are insufficient to adequately capture the complex nature of the tasks. Though in some cases the label accuracy matches that of neural-symbolic or symbolic-symbolic approaches (e.g., in Task 3), overall the performance is much worse, especially for successor accuracy and the final task of sequence classification. As for the other approaches, there is no clear class of winner methods for each and every task, though the approach that combines Scallop with top-1 proof for constraint classification with a Multi-Layer Perceptron for next state prediction seems to be a reasonable compromise across multiple tasks.

\begin{figure}
	\centering
    \begin{minipage}{\linewidth}
    	\includegraphics[width=1.0\textwidth]{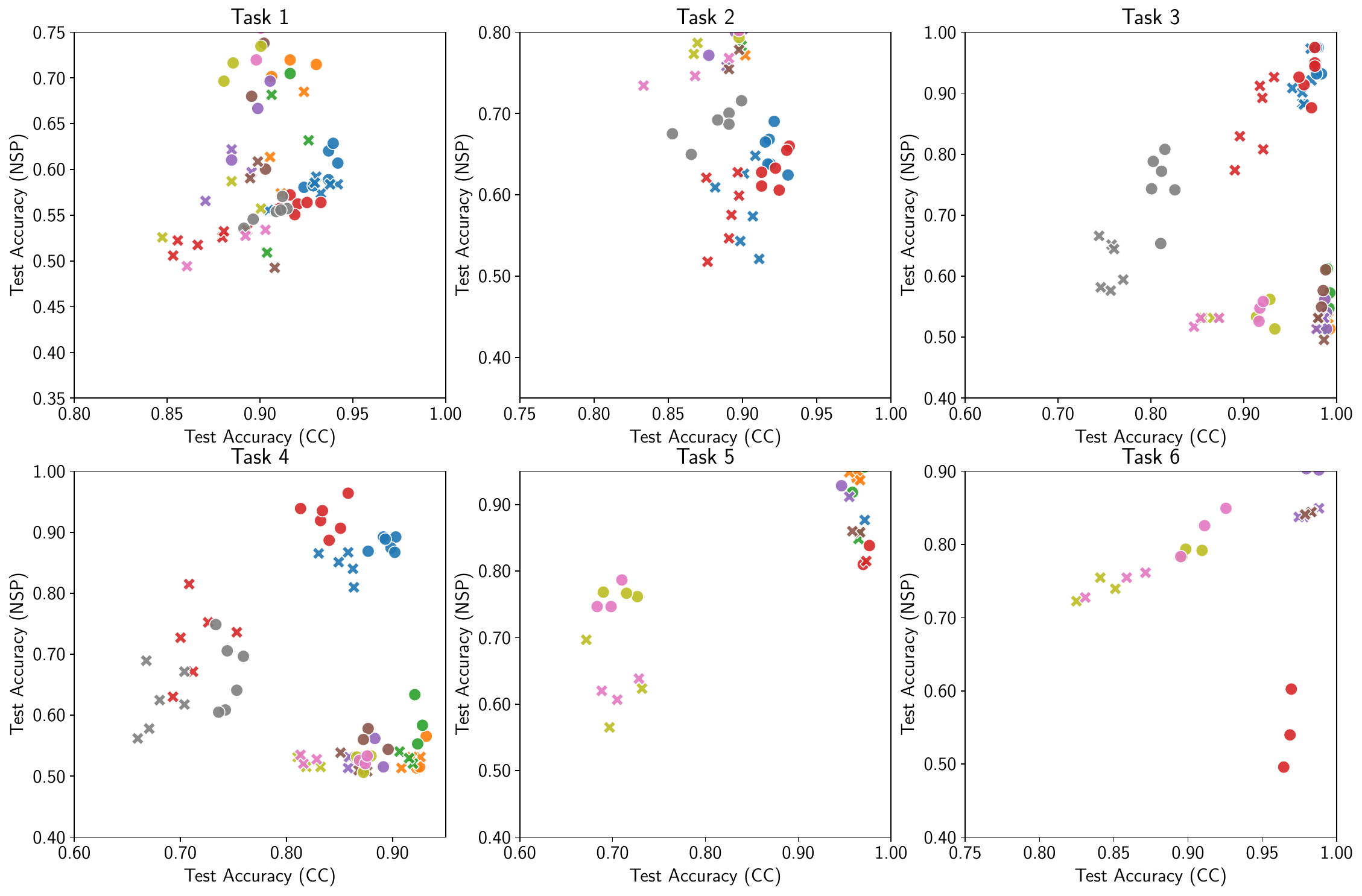}
           \end{minipage}
 \begin{minipage}{0.4\linewidth}
    \includegraphics[width=1.0\textwidth]{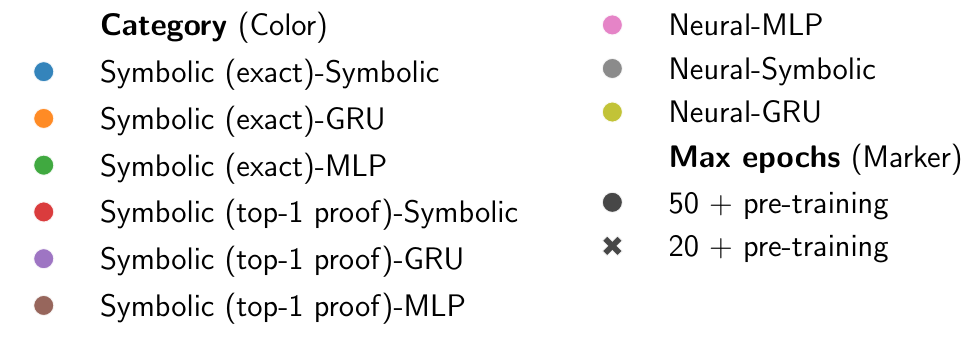}   
    \end{minipage}
	\caption{{\sc cc}-{\sc nsp} accuracy trade-off on short-sequence experiments for different categories.} 
	\label{fig:sequential-tradeoff-short}
\end{figure}

For a more detailed analysis of these results, Figure~\ref{fig:sequential-tradeoff-short} shows, for every short-sequence experimental run, the \textsc{cc}-\textsc{nsp} accuracy trade-off, highlighting different characteristics of each category. These results strongly confirm the advantages of symbolic-symbolic, or neural-symbolic, approaches with respect to purely neural architectures, even though they are harder to optimize (larger dispersion between points, compared to neural-only approaches, and fewer data points available due to diverging runs). The scatter plots also highlight the contribution of training epochs: more epochs, in fact, seem to help the system achieving an overall better performance (in the charts, this is indicated by dots closer to the top-right corner with respect to crosses, whatever the color), even though some families benefit less from additional training.
Focusing on symbolic methods for \textsc{cc} and \textsc{nsp} (red for Scallop and DFAs, blue for ProbLog and DFAs), they generally achieve similar performances, but with some trends identifiable in specific tasks. For tasks 1-4, a symbolic Scallop module is characterized by a noticeable dependency on the number of training epochs (red crosses and red dots forming two distinct clusters), while ProbLog, under the same tasks, can achieve similar performance (clusters can still be identified, with 50 epochs being slightly better, but they lie very close to each other), regardless of the number of training epochs. This behavior can in part be explained by the amount of information backpropagated through either component (only along the most probable proof for Scallop, and along the entire proof-tree for ProbLog). Tasks 5 and 6 pose optimization challenges for both methods, with only few combinations of hyper-parameters able to converge (a zero variance in Table~\ref{tab:sequential-results-short} is often due to only one experiment being able to converge).

Table~\ref{tab:sequential-results-long} corresponds to long sequence experiments. Neural-only experiments do not benefit from longer sequence horizons, as their accuracies across every objective remain roughly the same. Regardless of the \textsc{cc} module, label and constraint accuracies are also close to their short-sequence experiment counterparts. This result is somewhat expected, as neither objective is time dependent. The additional amount of data (about 5 to 10 times more abundant when moving from sequences of random length 10-20 to sequences of random length 50-100) does not seem to provide noticeable benefits in terms of label or constraint accuracy.
When focusing on \textsc{nsp} and \textsc{sc} objectives, it can be observed an improvement of performance, compared to short-sequence experiments, but only for those combinations which already managed to achieve good performance. The gap between good and bad combinations is noticeably increased, with sequence accuracy basically being either close to random guessing, or close to perfect classification.
\begin{table}
    \centering
    \resizebox{\textwidth}{!}{
    \begin{tabular}{cccccccc}
        \toprule
        \multirow{2}{*}{Task} & \multirow{2}{*}{Category} & \multirow{2}{*}{\shortstack[c]{Best model\\(best epoch)}} & \multirow{2}{*}{\shortstack[c]{Average\\Accuracy}\raisebox{1ex}{\:$\uparrow$}} & \multirow{2}{*}{\shortstack[c]{Label\\Accuracy}\raisebox{1ex}{\:$\uparrow$}} & \multirow{2}{*}{\shortstack[c]{Constraint\\Accuracy}\raisebox{1ex}{\:$\uparrow$}} & \multirow{2}{*}{\shortstack[c]{Successor\\Accuracy}\raisebox{1ex}{\:$\uparrow$}} & \multirow{2}{*}{\shortstack[c]{Sequence\\Accuracy}\raisebox{1ex}{\:$\uparrow$}}\\
        & & & & & &\\
        \midrule
                \multirow{6}{*}{\shortstack[c]{Task 1\\(long)}} & Sym. (exact)-Sym. & ProbLog Fuzzy (45) & $0.69 \pm 0.00$ & $0.81 \pm 0.00$ & $\textbf{0.94} \pm 0.00$ & $0.52 \pm 0.00$ & $0.50 \pm 0.00$\\
                 & Sym. (exact)-Neu. & ProbLog MLP (43) & $\textbf{0.90} \pm 0.01$ & $\textbf{0.87} \pm 0.02$ & $0.92 \pm 0.02$ & $\textbf{0.83} \pm 0.02$ & $\textbf{1.00} \pm 0.00$\\
                 & Sym. (top-1)-Sym. & Scallop sd-DNNF (50) & $0.69 \pm 0.00$ & $0.83 \pm 0.04$ & $0.92 \pm 0.02$ & $0.51 \pm 0.00$ & $0.50 \pm 0.00$\\
                 & Sym. (top-1)-Neu. & Scallop MLP (40) & $0.87 \pm 0.01$ & $0.85 \pm 0.03$ & $0.90 \pm 0.01$ & $0.77 \pm 0.00$ & $0.96 \pm 0.02$\\
                 & Neu.-Sym. & MLP sd-DNNF (50) & $0.68 \pm 0.03$ & $0.88 \pm 0.01$ & $0.85 \pm 0.12$ & $0.50 \pm 0.02$ & $0.50 \pm 0.00$\\
                 & Neu.-Neu. & MLP GRU (47) & $0.80 \pm 0.11$ & $0.83 \pm 0.03$ & $0.88 \pm 0.02$ & $0.62 \pm 0.26$ & $0.88 \pm 0.16$\\
                \hdashline
                \multirow{6}{*}{\shortstack[c]{Task 2\\(long)}} & Sym. (exact)-Sym. & ProbLog Fuzzy (49) & $0.64 \pm 0.03$ & $0.67 \pm 0.12$ & $0.92 \pm 0.01$ & $0.42 \pm 0.02$ & $0.54 \pm 0.01$\\
                 & Sym. (exact)-Neu. & ProbLog GRU (43) & $\textbf{0.85} \pm 0.00$ & $0.89 \pm 0.01$ & $\textbf{0.94} \pm 0.00$ & $\textbf{0.78} \pm 0.00$ & $\textbf{0.80} \pm 0.00$\\
                 & Sym. (top-1)-Sym. & Scallop Fuzzy (45) & $0.72 \pm 0.02$ & $\textbf{0.91} \pm 0.01$ & $0.93 \pm 0.00$ & $0.45 \pm 0.01$ & $0.57 \pm 0.08$\\
                 & Sym. (top-1)-Neu. & Scallop MLP (50) & $\textbf{0.85} \pm 0.01$ & $0.89 \pm 0.02$ & $0.93 \pm 0.01$ & $\textbf{0.78} \pm 0.01$ & $\textbf{0.80} \pm 0.00$\\
                 & Neu.-Sym. & MLP Fuzzy (45) & $0.70 \pm 0.06$ & $0.89 \pm 0.01$ & $0.85 \pm 0.07$ & $0.46 \pm 0.08$ & $0.58 \pm 0.08$\\
                 & Neu.-Neu. & MLP MLP (50) & $0.84 \pm 0.00$ & $0.89 \pm 0.00$ & $0.92 \pm 0.00$ & $0.76 \pm 0.00$ & $0.77 \pm 0.00$\\
                \hdashline
                \multirow{6}{*}{\shortstack[c]{Task 2\\(long)}} & Sym. (exact)-Sym. & ProbLog sd-DNNF (46) & $\textbf{0.97} \pm 0.00$ & $0.98 \pm 0.00$ & $0.98 \pm 0.00$ & $\textbf{0.96} \pm 0.00$ & $0.95 \pm 0.00$\\
                 & Sym. (exact)-Neu. & ProbLog MLP (42) & $0.75 \pm 0.01$ & $\textbf{0.99} \pm 0.00$ & $\textbf{0.99} \pm 0.00$ & $0.50 \pm 0.02$ & $0.52 \pm 0.03$\\
                 & Sym. (top-1)-Sym. & Scallop Fuzzy (45) & $0.96 \pm 0.01$ & $0.97 \pm 0.00$ & $0.97 \pm 0.00$ & $\textbf{0.96} \pm 0.01$ & $\textbf{0.96} \pm 0.01$\\
                 & Sym. (top-1)-Neu. & Scallop MLP (43) & $0.75 \pm 0.01$ & $\textbf{0.99} \pm 0.00$ & $\textbf{0.99} \pm 0.00$ & $0.50 \pm 0.02$ & $0.51 \pm 0.02$\\
                 & Neu.-Sym. & MLP sd-DNNF (40) & $0.81 \pm 0.01$ & $0.93 \pm 0.01$ & $0.80 \pm 0.02$ & $0.76 \pm 0.04$ & $0.76 \pm 0.03$\\
                 & Neu.-Neu. & MLP MLP (20) & $0.72 \pm 0.01$ & $0.98 \pm 0.00$ & $0.90 \pm 0.01$ & $0.51 \pm 0.01$ & $0.49 \pm 0.05$\\
                \hdashline
                \multirow{6}{*}{\shortstack[c]{Task 4\\(long)}} & Sym. (exact)-Sym. & ProbLog sd-DNNF (16) & $\textbf{0.86} \pm 0.01$ & $0.86 \pm 0.00$ & $0.89 \pm 0.00$ & $\textbf{0.84} \pm 0.02$ & $\textbf{0.84} \pm 0.02$\\
                 & Sym. (exact)-Neu. & ProbLog MLP (23) & $0.73 \pm 0.01$ & $0.91 \pm 0.01$ & $\textbf{0.92} \pm 0.00$ & $0.53 \pm 0.01$ & $0.54 \pm 0.04$\\
                 & Sym. (top-1)-Sym. & Scallop Fuzzy (18) & $0.79 \pm 0.02$ & $0.82 \pm 0.03$ & $0.80 \pm 0.01$ & $0.77 \pm 0.04$ & $0.78 \pm 0.03$\\
                 & Sym. (top-1)-Neu. & Scallop MLP (49) & $0.71 \pm 0.01$ & $\textbf{0.92} \pm 0.01$ & $0.91 \pm 0.00$ & $0.52 \pm 0.03$ & $0.50 \pm 0.05$\\
                 & Neu.-Sym. & MLP Fuzzy (14) & $0.67 \pm 0.03$ & $0.82 \pm 0.01$ & $0.77 \pm 0.01$ & $0.56 \pm 0.04$ & $0.54 \pm 0.05$\\
                 & Neu.-Neu. & MLP MLP (41) & $0.71 \pm 0.02$ & $0.91 \pm 0.00$ & $0.89 \pm 0.01$ & $0.52 \pm 0.01$ & $0.51 \pm 0.06$\\
                \hdashline
                \multirow{6}{*}{\shortstack[c]{Task 5\\(long)}} & Sym. (exact)-Sym. & ProbLog sd-DNNF (39) & $0.61 \pm 0.29$ & $0.50 \pm 0.70$ & $0.84 \pm 0.22$ & $0.59 \pm 0.21$ & $0.54 \pm 0.05$\\
                 & Sym. (exact)-Neu. & ProbLog GRU (43) & $\textbf{0.99} \pm 0.00$ & $\textbf{0.99} \pm 0.00$ & $\textbf{0.99} \pm 0.00$ & $\textbf{0.97} \pm 0.00$ & $\textbf{1.00} \pm 0.00$\\
                 & Sym. (top-1)-Sym. & Scallop Fuzzy (48) & $0.80 \pm 0.02$ & $\textbf{0.99} \pm 0.00$ & $\textbf{0.99} \pm 0.00$ & $0.69 \pm 0.02$ & $0.54 \pm 0.04$\\
                 & Sym. (top-1)-Neu. & Scallop MLP (50) & $0.94 \pm 0.08$ & $0.97 \pm 0.03$ & $0.96 \pm 0.04$ & $0.90 \pm 0.12$ & $0.93 \pm 0.11$\\
                 & Neu.-Sym. & MLP sd-DNNF (49) & $0.68 \pm 0.00$ & $0.97 \pm 0.00$ & $0.79 \pm 0.00$ & $0.45 \pm 0.00$ & $0.50 \pm 0.00$\\
                 & Neu.-Neu. & MLP GRU (31) & $0.68 \pm 0.28$ & $0.63 \pm 0.46$ & $0.71 \pm 0.05$ & $0.64 \pm 0.26$ & $0.75 \pm 0.35$\\
                \hdashline
                \multirow{6}{*}{\shortstack[c]{Task 6\\(long)}} & Sym. (exact)-Sym. & ProbLog sd-DNNF (35) & $0.62 \pm 0.00$ & $\textbf{0.99} \pm 0.00$ & $0.68 \pm 0.00$ & $0.31 \pm 0.00$ & $0.50 \pm 0.00$\\
                 & Sym. (exact)-Neu. & ProbLog GRU (50) & $0.64 \pm 0.03$ & $\textbf{0.99} \pm 0.00$ & $0.68 \pm 0.00$ & $0.39 \pm 0.12$ & $0.50 \pm 0.00$\\
                 & Sym. (top-1)-Sym. & Scallop sd-DNNF (49) & $0.74 \pm 0.11$ & $0.95 \pm 0.07$ & $0.89 \pm 0.18$ & $0.59 \pm 0.15$ & $0.52 \pm 0.03$\\
                 & Sym. (top-1)-Neu. & Scallop MLP (42) & $\textbf{0.97} \pm 0.00$ & $\textbf{0.99} \pm 0.00$ & $\textbf{0.99} \pm 0.00$ & $\textbf{0.94} \pm 0.00$ & $\textbf{0.98} \pm 0.00$\\
                 & Neu.-Sym. & MLP sd-DNNF (22) & $0.64 \pm 0.00$ & $0.96 \pm 0.01$ & $0.68 \pm 0.00$ & $0.43 \pm 0.00$ & $0.50 \pm 0.00$\\
                 & Neu.-Neu. & MLP MLP (39) & $0.86 \pm 0.04$ & $0.90 \pm 0.07$ & $0.83 \pm 0.02$ & $0.78 \pm 0.05$ & $0.93 \pm 0.05$\\

        \bottomrule
    \end{tabular}
    }
    \caption{Best models test set accuracies (mean $\pm$ std over 3 random seeds, after discarding diverging runs) for long-sequence experiments, grouped by category. Best model (named ``{\sc cc} model {\sc nsp} model'') and epoch selected by Average Accuracy on validation set.}
    \label{tab:sequential-results-long}
\end{table}

\begin{figure}
	\centering
    \begin{minipage}{\linewidth}
    	\includegraphics[width=1.0\textwidth]{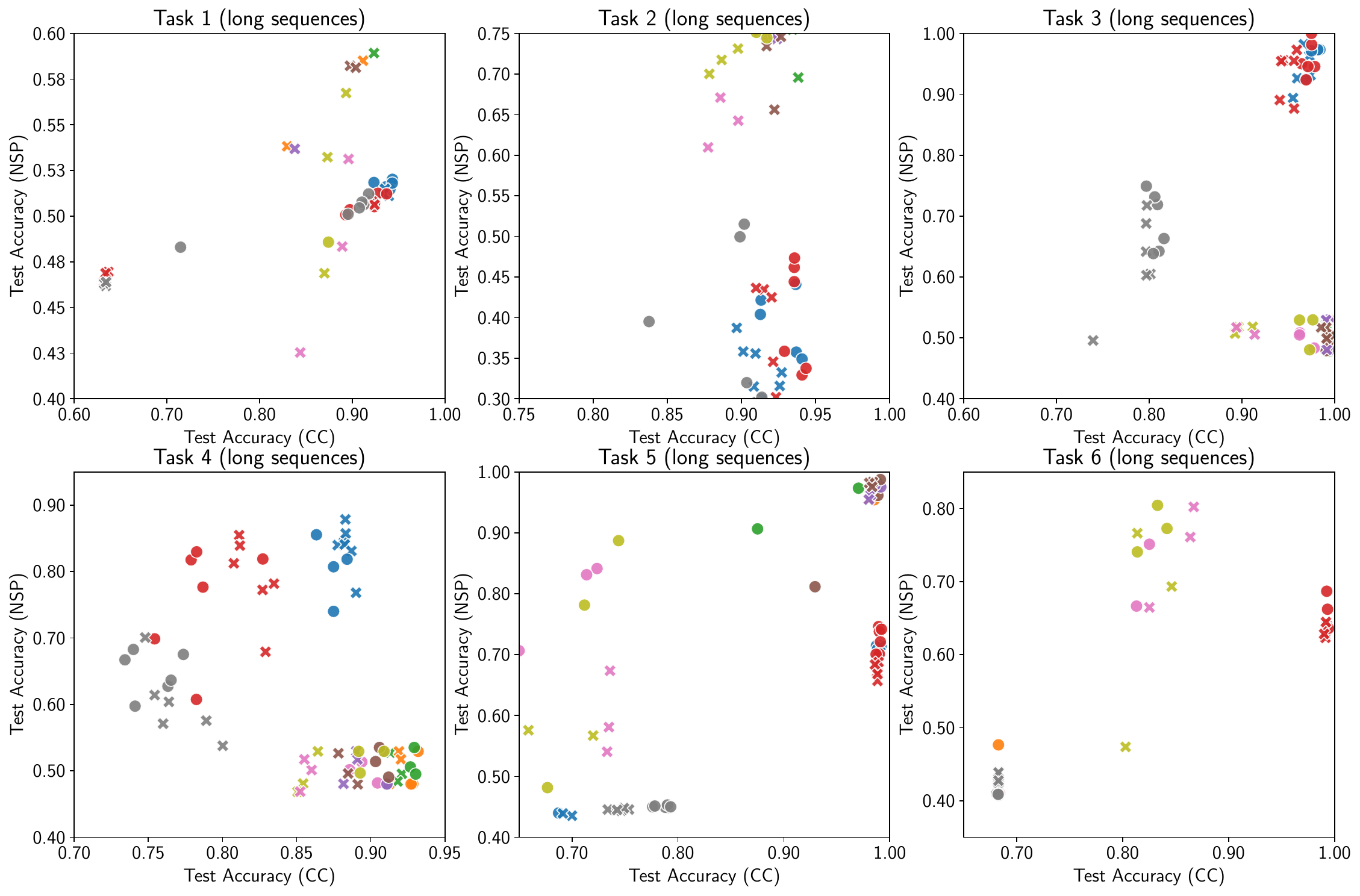}
           \end{minipage}
 \begin{minipage}{0.4\linewidth}
    \includegraphics[width=1.0\textwidth]{imgs/const-vs-succ_legend.pdf}   
    \end{minipage}
	\caption{{\sc cc}-{\sc nsp} accuracy trade-off on long-sequence experiments for different categories.} 
	\label{fig:sequential-tradeoff-long}
\end{figure}

Figure~\ref{fig:sequential-tradeoff-long} highlights the \textsc{cc}-\textsc{nsp} accuracy trade-off for long sequences. Relative positions across different categories are mostly preserved (even though distances are different), however different training epochs no longer form distinguishable clusters. Symbolic methods for \textsc{cc} and \textsc{nsp} (Scallop red, ProbLog blue) are characterized by different trade-offs, when compared to short-sequence experiments.
For instance, in task 1 Scallop with few epochs (red crosses) presents a failure case of the symbolic methods (in short-sequence experiments its performance was still competitive in terms of constraint accuracy), while ProbLog is unaffected by the number of epochs. Task 4 is characterized by completely different trade-offs: ProbLog clearly dominates both \textsc{cc} and \textsc{nsp} objectives in long-sequence experiments, while in short-sequence ones, Scallop was noticeably better in terms of \textsc{nsp} and lagging behind ProbLog on the \textsc{cc} axis.
Task 5 presents a failure case for ProbLog, as long sequences cause a dramatic drop in performance (blue markers were in the upper right corner for short sequences, while they are in the lower left in this case), however in this task Scallop is mostly unaffected (performance is only subject to a slight drop in values and a larger variation across the \textsc{nsp} axis).


\section{Class-Continual Learning with LTLZinc}\label{sec:contexp}
In this section we explore LTLZinc capabilities in evaluating novel continual learning settings, by experimenting on two class-continual tasks with complex temporal behavior, instantiated with two different perceptual complexities.

\subsection{Proposed Tasks}\label{sec:proposed-ccl}
We experiment on two class-continual learning tasks, instantiated over two different perceptual domains (MNIST digits, 10 classes, and Cifar-100, 100 classes). Every task is generated with 3 different seeds, producing different curricula which are equivalent from the perspective of the temporal specification. Either task is associated with the generated trace along the automaton and the constraint validity for each episode. Each image within an episode is annotated with class labels.
For evaluation purposes, a subset of labels is provided as targets for ``focused'' metrics.

\subsubsection{Task 1} This task models the presence of \textbf{rare classes}, appearing only once along the agent experience. Metrics will focus on evaluating knowledge retention about those rare classes.

\paragraph{MNIST}
\begin{align*}
 \mathcal{X}\colon\quad&\mathcal{X}_Y = \left\{\mathimg{mnist0}, \mathimg{mnist1}, \mathimg{mnist2}, \mathimg{mnist3}, \mathimg{mnist4}, \mathimg{mnist5}, \mathimg{mnist6}, \mathimg{mnist7}, \mathimg{mnist8}, \mathimg{mnist9}\right\}\\
 \mathcal{Y}\colon\quad&\mathcal{Y}_Y = [0,9]\\
 \mathcal{C}\colon\quad&even: Y \in \left\{2, 4, 6, 8\right\};\\
 &odd: Y \in \left\{1, 3, 5, 7, 9\right\};\\
 &zero: Y = 0\\
 \mathcal{F}\colon\quad&\neg zero \wedge (\neg zero \ltluntil (zero \wedge \ltlweaknext \ltlglobally \neg zero).
\end{align*}
Predicates $odd, even$ are orphans, guaranteed to be sampled in positive form at least for one episode along the curriculum (for $100\%$ of the samples inside that episode). Curriculum length: 10 episodes, dataset: 1000 images per episode (800 train, 100 val, 100 test). Focused accuracy and forgetting refer to predicate $zero$.
Note that since predicate $zero$ corresponds to a single class, the episode in which it appears will be characterized by a \textbf{one-class learning} setting.

\paragraph{Cifar-100}
\begin{align*}
 \mathcal{X}\colon\quad&\mathcal{X}_Y = \texttt{cifar-100 dataset}\\
 \mathcal{Y}\colon\quad&\mathcal{Y}_Y = \texttt{cifar-100 classes}\\
 \mathcal{C}\colon\quad&animals: Y \in acquatic\_mammals \cup fish \cup insects \cup large\_carnivores \cup \\
 &\quad\quad\quad\quad\quad\quad large\_omnivores\_and\_herbivores \cup medium\_sized\_mammals \cup \\
 &\quad\quad\quad\quad\quad\quad non\_insect\_invertebrates \cup people \cup reptiles \cup small\_mammals;\\
 &plants: Y \in flowers \cup fruit\_and\_vegetables \cup trees;\\
 &inanimate: Y \in food\_containers \cup household\_electrical\_devices \cup \\
 &\quad\quad\quad\quad\quad\quad household\_furniture \cup large\_manmade\_outdoors\_things \cup \\
 &\quad\quad\quad\quad\quad\quad large\_natural\_outdoors\_scenes \cup vehicles\_1 \cup vehicles\_2\\
 \mathcal{F}\colon\quad&\neg plants \wedge (\neg plants \ltluntil (plants \wedge \ltlweaknext \ltlglobally \neg plants).
\end{align*}
Predicates $animals, inanimate$ are orphans, guaranteed to be sampled in positive form at least for one episode along the curriculum (for $100\%$ of the samples inside that episode). Curriculum length: 50 episodes, dataset: 1000 images per episode (800 train, 100 val, 100 test). Focused accuracy and forgetting refer to predicate $plants$.
Learning is at the class level (100 targets), predicates are defined in terms of Cifar-100 superclasses (e.g., $acquatic\_mammals = \left\{beaver, dolphin, otter, seal, whale\right\}$).
Unlike the MNIST version of this task, predicate $plants$ corresponds to multiple classes, therefore, even though they are presented only during a single episode, such episode is still a traditional \textbf{multi-class supervised learning} setting.

\subsubsection{Task 2} This task models classes reappearing cyclically over time. Metrics will focus on evaluating knowledge retention of recurring classes.

\paragraph{MNIST}
\begin{align*}
 \mathcal{X}\colon\quad&\mathcal{X}_Y = \left\{\mathimg{mnist0}, \mathimg{mnist1}, \mathimg{mnist2}, \mathimg{mnist3}, \mathimg{mnist4}, \mathimg{mnist5}, \mathimg{mnist6}, \mathimg{mnist7}, \mathimg{mnist8}, \mathimg{mnist9}\right\}\\
 \mathcal{Y}\colon\quad&\mathcal{Y}_Y = [0,9]\\
 \mathcal{C}\colon\quad&p: Y \in \left\{0, 1, 2\right\};\\
 &q: Y \in \left\{3, 4, 5\right\};\\
 &r: Y \in \left\{6, 7, 8\right\};\\
 &s: Y = 9.\\
 \mathcal{F}\colon\quad&\ltlglobally(p \leftrightarrow (\ltlnext \neg q \wedge \ltlnext\ltlweaknext q)).
\end{align*}
Predicates $r, s$ are orphans, guaranteed to be sampled in positive form at least for one episode along the curriculum (for $50\%$ of the samples inside that episode). Curriculum length: 20 episodes, dataset: 1000 images per episode (800 train, 100 val, 100 test). Focused metrics refer to predicates $p, q$.

\paragraph{Cifar-100}
\begin{align*}
 \mathcal{X}\colon\quad&\mathcal{X}_Y = \left\{\mathimg{mnist0}, \mathimg{mnist1}, \mathimg{mnist2}, \mathimg{mnist3}, \mathimg{mnist4}, \mathimg{mnist5}, \mathimg{mnist6}, \mathimg{mnist7}, \mathimg{mnist8}, \mathimg{mnist9}\right\}\\
 \mathcal{Y}\colon\quad&\mathcal{Y}_Y = [0,9]\\
 \mathcal{C}\colon\quad&animals: Y \in acquatic\_mammals \cup fish \cup insects \cup large\_carnivores \cup \\
 &\quad\quad\quad\quad\quad\quad large\_omnivores\_and\_herbivores \cup medium\_sized\_mammals \cup \\
 &\quad\quad\quad\quad\quad\quad non\_insect\_invertebrates \cup people \cup reptiles \cup small\_mammals;\\
 &plants: Y \in flowers \cup fruit\_and\_vegetables \cup trees;\\
 &inside: Y \in food\_containers \cup household\_electrical\_devices \cup household\_furniture;\\
 &outside: Y \in large\_manmade\_outdoors\_things \cup large\_natural\_outdoors\_scenes \cup \\
 &\quad\quad\quad\quad\quad\quad vehicles\_1 \cup vehicles\_2.\\
 \mathcal{F}\colon\quad&\ltlglobally(inside \leftrightarrow (\ltlnext \neg outside \wedge \ltlnext\ltlweaknext outside)).
\end{align*}
Predicates $animals, plants$ are orphans, guaranteed to be sampled in positive form at least for one episode along the curriculum (for $50\%$ of the samples inside that episode). Curriculum length: 50 episodes, dataset: 1000 images per episode (800 train, 100 val, 100 test). Focused metrics refer to predicates $inside, outside$.
Learning is at the class level (100 targets), predicates are defined in terms of Cifar-100 superclasses.

\subsection{Methodology}
We address the proposed class-continual learning tasks by means of convolutional neural networks (CNNs), augmented (only at training-time) by a combination of strategies, representing the main categories of replay-based, regularization-based and architecture-based continual learning. Appendix~\ref{app:contexp} provides additional information to reproduce our experiments.
%
\paragraph{Architecture.} The architecture is composed of a GoogLeNet~\cite{szegedy2015going} backbone, equipped with a classification head, comprised of: a linear embedding layer of $emb\_size$ neurons, a dictionary of hidden blocks (linear layer of $hidden\_size$ neurons, followed by ReLU activations), and a linear classification neuron for each class.
If the architecture-based strategy is not active, the dictionary contains a single hidden block, acting as simple reprojection of the embedding layer, otherwise, it is composed of an independent hidden block for each of the injected \textit{knowledge units}, and the output tensors are concatenated before being passed to the classification layer.
In the MNIST version of Tasks 1 and 2, the backbone is initialized with random weights, while for the Cifar-100 version, the backbone is initialized with weights pre-trained on ImageNet~\cite{russakovsky2015imagenet}. For Cifar-100 experiments we also explore the effect of freezing the backbone (training only the downstream linear embedding, hidden blocks and linear classifier layers).

\paragraph{Continual Learning Strategies.} We address the catastrophic forgetting problem with multiple combinations of continual learning strategies, enabled for different experiments. For the experience-replay family of strategies, we experiment with class-balanced reservoir sampling~\cite{chrysakis2020online} associated with either a categorical cross-entropy loss, or a gradient episodic memory (GEM)~\cite{lopez2017gradient} loss. In both cases, a replay buffer of fixed size is capable of storing up to $\text{buffer\_size} = 500$ previously observed image samples and their class labels, which can be read in batches of size $\text{buffer\_batch} = 16$. As a representative of regularization-based strategies, we implement learning without forgetting (LwF)~\cite{li2017learning}, as a form of self-distillation of the model prediction against both new and buffered data.

\paragraph{Knowledge-Injection.} We inject knowledge at the continual learning strategy level, by means of segregation with respect to $k$ \textbf{knowledge units}. In our experiments we focus on three families of units: \textbf{None} (traditional continual learning baseline), \textbf{Predicates} and \textbf{States}.
Predicate-level units are a form of ``high-level'' knowledge, defining a partition based on which constraints in $\mathcal{C}$ are true in the current episode, while state-level units partition based on the automaton state corresponding to the current episode (and are thus at a lower level of abstraction, compared to predicate units).
Knowledge is injected in replay-based strategies by reserving $\lfloor \frac{\text{buffer\_size}}{k}\rfloor$ samples for each knowledge unit. The GEM algorithm is modified to take into account knowledge units, instead of class labels during its quadratic programming step. 
LwF is modified into a form of teacher-student distillation, by keeping in memory $k$ copies of the neural architecture, each of which is trained only on image samples corresponding to their own knowledge unit.
The modular architecture-based strategy natively supports knowledge injection, as each of the hidden blocks already corresponds to a different knowledge unit.
Please note that in this setting, as knowledge units interact only with continual learning strategies, it is not possible to inject knowledge onto the Naive combination, where no strategy is active.

\paragraph{Training.} At the beginning of each episode, an oracular controller determines the subset of knowledge units active for the episode,\footnote{For the \textbf{None} injection level, a single dummy knowledge unit is always kept active.} based on auxiliary LTLZinc annotations (i.e., current automaton state and constraint truth values).\footnote{In principle, the controller can be learned, instead of relying on ground truth annotations, e.g., \cite{zhu2024continual}.} This subset will guide continual learning strategies during training.
The training loop for a single episode is composed of a setup phase, a training phase and an evaluation phase.
In the \textbf{setup phase}, the weights of hidden blocks corresponding to inactive knowledge units are frozen, and, if LwF is active with no background knowledge available (\textbf{None} setting), weights of the entire neural network are copied onto a single teacher for self-distillation during the training phase.

The general idea of the \textbf{training phase} is to allow plasticity by updating \textbf{active knowledge units} with new samples, while regularizing the agent (which always learns from new samples) against \textbf{inactive knowledge units} to guarantee stability. To achieve this, a minibatch is sampled from the current episode, processed by the replay-based and regularization-based strategies enabled for the current experiment, and finally used to train the neural network (along with additional objectives provided by CL strategies).
The replay buffers corresponding to currently active knowledge units stochastically store each observed image in the minibatch, according to the reservoir sampling algorithm. Teachers corresponding to active knowledge units are trained against the minibatch by means of categorical cross-entropy.
The convolutional neural network is trained on the minibatch (categorical cross-entropy loss), as well as the following auxiliary objectives, when the corresponding strategy is enabled for the current experiment: experience-replay loss on a buffered minibatch sampled from inactive knowledge units (categorical cross-entropy, or GEM loss), LwF loss against inactive teachers (Kullback-Leibler divergence, both on current and buffered minibatches). Gradient is backpropagated to the backbone only through the hidden blocks corresponding to active knowledge units.

Finally, at the \textbf{evaluation phase}, well-established metrics~\cite{mai2022online} in the continual learning literature are evaluated against training, validation and test sets.
\textbf{Average accuracy} is the average of the class-related accuracies at the end of learning, \textbf{Forward transfer} measures the positive effect of learned features on yet-to-be-observed future episodes, and \textbf{Average forgetting} is the difference between the current average accuracy (averaged over every past episode) and the best observed average accuracy in the past.
For average accuracy and forgetting, we also focus their evaluation on the target classes specified for each task (i.e., rare classes for Task 1 and recurrent classes for Task 2). In the case in which an episode contains no such classes, focused episode-level accuracy is not defined, and discarded when computing global average accuracy and forgetting.
We perform model selection by measuring average accuracy at the end of training on the validation set.

\subsection{Results}
\begin{table}
    \centering
    \resizebox{\textwidth}{!}{
    \begin{tabular}{ccccccccccc}
        \toprule
        \multirow{3}{*}{Task} & \multirow{3}{*}{\shortstack[c]{Knowledge\\Availability}} & \multirow{3}{*}{Category} & \multirow{3}{*}{Buffer} & \multirow{3}{*}{Distillation} & \multirow{3}{*}{Architecture} & \multirow{3}{*}{\shortstack[c]{Average\\Accuracy}\raisebox{1.5ex}{\:$\uparrow$}} & \multirow{3}{*}{\shortstack[c]{Average\\Forgetting}\raisebox{1.5ex}{\:$\downarrow$}} & \multirow{3}{*}{\shortstack[c]{Forward\\Transfer}\raisebox{1.5ex}{\:$\uparrow$}} & \multirow{3}{*}{\shortstack[c]{Focused\\Average\\Accuracy}\raisebox{2.5ex}{\:$\uparrow$}} & \multirow{3}{*}{\shortstack[c]{Focused\\Average\\Forgetting}\raisebox{2.5ex}{\:$\downarrow$}} \\
         & & & & & & & & & & \\
         & & & & & & & & & & \\
        \midrule
                \multirow{5}{*}{\shortstack[c]{Task 1\\(MNIST)}} & \multirow{5}{*}{Predicates} & Replay & Reservoir (CCE) & No distillation & Flat & $0.95 \pm 0.02$ & $\textbf{0.03} \pm 0.02$ & $\textbf{0.70} \pm 0.03$ & $0.71 \pm 0.22$ & $0.26 \pm 0.22$\\
                 &  & Modular & No buffer & No distillation & Modular & $0.88 \pm 0.01$ & $0.11 \pm 0.00$ & $0.51 \pm 0.02$ & $0.00 \pm 0.00$ & $1.00 \pm 0.00$\\
                 &  & Replay + Distillation & Reservoir (CCE) & Teacher-distillation & Flat & $0.95 \pm 0.02$ & $\textbf{0.03} \pm 0.02$ & $0.68 \pm 0.04$ & $\textbf{0.78} \pm 0.23$ & $\textbf{0.20} \pm 0.22$\\
                 &  & Replay + Modular & Reservoir (CCE) & No distillation & Modular & $\textbf{0.96} \pm 0.02$ & $\textbf{0.03} \pm 0.03$ & $\textbf{0.70} \pm 0.03$ & $0.77 \pm 0.23$ & $0.21 \pm 0.25$\\
                 &  & All & Reservoir (CCE) & Teacher-distillation & Modular & $0.95 \pm 0.03$ & $\textbf{0.03} \pm 0.03$ & $0.67 \pm 0.03$ & $0.76 \pm 0.26$ & $0.24 \pm 0.26$\\
                \hdashline
                \multirow{5}{*}{\shortstack[c]{Task 1\\(MNIST)}} & \multirow{5}{*}{States} & Replay & Reservoir (CCE) & No distillation & Flat & $\textbf{0.97} \pm 0.00$ & $\textbf{0.01} \pm 0.01$ & $\textbf{0.66} \pm 0.02$ & $\textbf{0.89} \pm 0.05$ & $\textbf{0.09} \pm 0.06$\\
                 &  & Modular & No buffer & No distillation & Modular & $0.88 \pm 0.00$ & $0.11 \pm 0.01$ & $0.52 \pm 0.01$ & $0.00 \pm 0.00$ & $1.00 \pm 0.00$\\
                 &  & Replay + Distillation & Reservoir (GEM) & Teacher-distillation & Flat & $0.95 \pm 0.01$ & $0.03 \pm 0.01$ & $0.58 \pm 0.02$ & $0.79 \pm 0.06$ & $0.21 \pm 0.06$\\
                 &  & Replay + Modular & Reservoir (GEM) & No distillation & Modular & $0.96 \pm 0.01$ & $0.03 \pm 0.02$ & $0.58 \pm 0.03$ & $0.76 \pm 0.08$ & $0.24 \pm 0.08$\\
                 &  & All & Reservoir (CCE) & Teacher-distillation & Modular & $0.96 \pm 0.01$ & $0.02 \pm 0.01$ & $0.65 \pm 0.02$ & $0.87 \pm 0.05$ & $0.12 \pm 0.05$\\
                \hdashline
                \multirow{3}{*}{\shortstack[c]{Task 1\\(MNIST)}} & \multirow{3}{*}{None} & Naive & No buffer & No distillation & Modular & $\textbf{0.88} \pm 0.01$ & $\textbf{0.11} \pm 0.00$ & $0.47 \pm 0.03$ & $0.00 \pm 0.00$ & $1.00 \pm 0.00$\\
                 &  & Replay & Reservoir (CCE) & No distillation & Flat & $\textbf{0.88} \pm 0.01$ & $\textbf{0.11} \pm 0.01$ & $0.50 \pm 0.02$ & $0.00 \pm 0.00$ & $1.00 \pm 0.00$\\
                 &  & Replay + Distillation & Reservoir (CCE) & Self-distillation & Flat & $\textbf{0.88} \pm 0.00$ & $0.12 \pm 0.00$ & $\textbf{0.53} \pm 0.01$ & $0.00 \pm 0.00$ & $1.00 \pm 0.00$\\
                \midrule
                \multirow{5}{*}{\shortstack[c]{Task 2\\(MNIST)}} & \multirow{5}{*}{Predicates} & Replay & Reservoir (CCE) & No distillation & Flat & $\textbf{0.96} \pm 0.01$ & $\textbf{0.03} \pm 0.01$ & $\textbf{0.74} \pm 0.02$ & $\textbf{0.94} \pm 0.02$ & $\textbf{0.04} \pm 0.03$\\
                 &  & Modular & No buffer & No distillation & Modular & $0.49 \pm 0.11$ & $0.53 \pm 0.12$ & $0.31 \pm 0.02$ & $0.49 \pm 0.24$ & $0.53 \pm 0.22$\\
                 &  & Replay + Distillation & Reservoir (CCE) & Teacher-distillation & Flat & $0.92 \pm 0.05$ & $0.06 \pm 0.05$ & $0.66 \pm 0.05$ & $0.88 \pm 0.09$ & $0.10 \pm 0.09$\\
                 &  & Replay + Modular & Reservoir (CCE) & No distillation & Modular & $0.94 \pm 0.05$ & $0.05 \pm 0.05$ & $0.71 \pm 0.03$ & $0.91 \pm 0.08$ & $0.08 \pm 0.08$\\
                 &  & All & Reservoir (CCE) & Teacher-distillation & Modular & $0.92 \pm 0.10$ & $0.06 \pm 0.09$ & $0.63 \pm 0.08$ & $0.88 \pm 0.16$ & $0.09 \pm 0.15$\\
                \hdashline
                \multirow{5}{*}{\shortstack[c]{Task 2\\(MNIST)}} & \multirow{5}{*}{States} & Replay & Reservoir (CCE) & No distillation & Flat & $\textbf{0.98} \pm 0.00$ & $\textbf{0.01} \pm 0.00$ & $\textbf{0.71} \pm 0.03$ & $\textbf{0.98} \pm 0.01$ & $0.02 \pm 0.01$\\
                 &  & Modular & No buffer & No distillation & Modular & $0.47 \pm 0.12$ & $0.55 \pm 0.13$ & $0.30 \pm 0.02$ & $0.47 \pm 0.24$ & $0.55 \pm 0.22$\\
                 &  & Replay + Distillation & Reservoir (CCE) & Teacher-distillation & Flat & $0.95 \pm 0.02$ & $0.03 \pm 0.02$ & $0.57 \pm 0.11$ & $0.94 \pm 0.04$ & $0.05 \pm 0.04$\\
                 &  & Replay + Modular & Reservoir (CCE) & No distillation & Modular & $\textbf{0.98} \pm 0.01$ & $\textbf{0.01} \pm 0.00$ & $0.70 \pm 0.02$ & $\textbf{0.98} \pm 0.01$ & $\textbf{0.01} \pm 0.01$\\
                 &  & All & Reservoir (CCE) & Teacher-distillation & Modular & $0.96 \pm 0.02$ & $0.02 \pm 0.01$ & $0.61 \pm 0.07$ & $0.95 \pm 0.03$ & $0.03 \pm 0.03$\\
                \hdashline
                \multirow{3}{*}{\shortstack[c]{Task 2\\(MNIST)}} & \multirow{3}{*}{None} & Naive & No buffer & No distillation & Flat & $\textbf{0.48} \pm 0.11$ & $\textbf{0.54} \pm 0.12$ & $\textbf{0.30} \pm 0.02$ & $\textbf{0.47} \pm 0.25$ & $\textbf{0.55} \pm 0.23$\\
                 & & Replay & Reservoir (GEM) & No distillation & Modular & $0.46 \pm 0.11$ & $0.56 \pm 0.11$ & $\textbf{0.30} \pm 0.02$ & $0.46 \pm 0.25$ & $0.56 \pm 0.23$\\
                 & & Replay + Distillation & Reservoir (CCE) & Self-distillation & Modular & $0.42 \pm 0.09$ & $0.60 \pm 0.10$ & $0.29 \pm 0.01$ & $0.42 \pm 0.31$ & $0.60 \pm 0.29$\\
        \bottomrule
    \end{tabular}
    }
    \caption{Best Average Accuracy, Average Forgetting and Forward Transfer for class-continual experiments on MNIST, grouped by knowledge available to each strategy. Best models selected by Average Accuracy on validation set. Results are $mean \pm std$ over 9 runs (3 random seeds, times 3 different curricula).}
    \label{tab:incremental-results-mnist}
\end{table}

Table~\ref{tab:incremental-results-mnist} highlights the best results on MNIST experiments. For both tasks, every method performs significantly above the random guessing threshold ($0.1$ accuracy), however methods employing knowledge injection (regardless of the level, predicates or states) have a clear advantage over baseline methods exploiting uninformed continual learning strategies. When used alone, the architecture-based strategy appears to interact negatively with knowledge injection: while in its uninformed version it performs on par (task 1) or slightly better (task 2) than other strategies, when combined with background knowledge, its performance drops significantly with respect to other strategies (by about $0.1$ for task 1 and $0.5$ for task 2). This phenomenon can be explained by the fact that such strategy is characterized by additional training parameters, which receive gradients only when the corresponding hidden block is active, with the result that, compared to the uninformed baseline, for the same amount of data, each parameter receives only a fraction of the updates it would be subject to if trained naively.
With the exception of the modular architecture, every other strategy (or combination of strategies) performs equally well in terms of average accuracy and average forgetting, however, regularization-based methods (distillation alone, distillation and replay, or distillation and modular architecture), tend to be characterized by lower forward transfer.
Metrics focusing on target classes, highlight how not employing background knowledge (or relying only on architecture-based strategies) for task 1 has catastrophic outcomes: knowledge about the rare class is entirely erased by the end of training. For task 2, on the other hand, where the classes of interest are those re-appearing periodically, the trend follows the same as global metrics. This behavior suggests the traditional evaluation protocols for fully-incremental settings, where all the classes are balanced across episodes and appear only once, can be potentially inadequate when adapted to more complex temporal behaviors.

\begin{figure}
    \centering
    \begin{minipage}{\linewidth}
    \includegraphics[width=\textwidth]{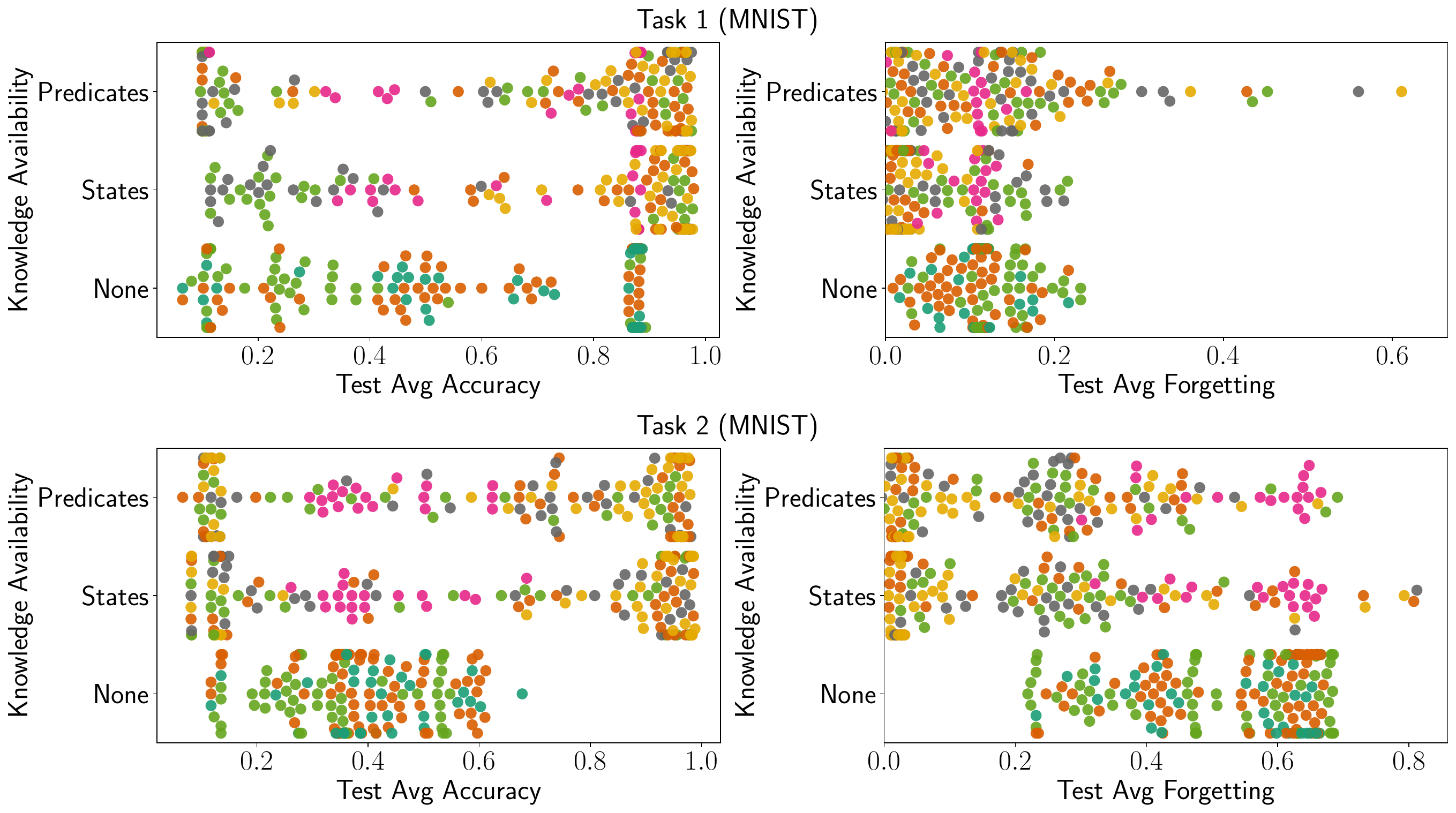}\\
    \includegraphics[width=\textwidth]{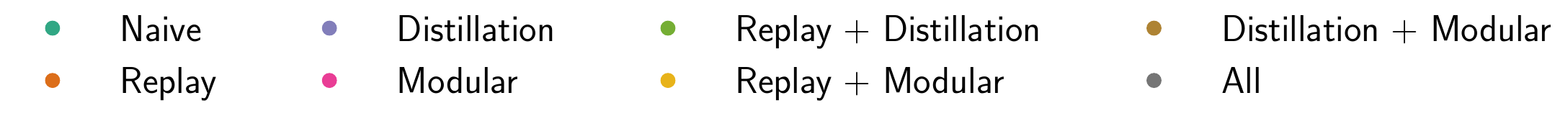}
    \end{minipage}
    \caption{Average accuracy and Average forgetting for class-continual experiments on MNIST, grouped by knowledge available to each strategy.}
    \label{fig:swarm-mnist}
\end{figure}

Figure~\ref{fig:swarm-mnist} shows average accuracy and average forgetting of every experiment, grouped by knowledge availability. It can be observed that, although knowledge availability has a potentially positive effect (best methods being able to achieve higher average accuracy and lower average forgetting compared to uninformed strategies), it is not enough on it own, to guarantee good performance (worst methods achieving as bad, for task 1, or slightly worse, in task 2, than uninformed baselines), and hyper-parameter tuning plays an equally important role.
When focusing on specific strategies, replay-based methods (and their combinations with other strategies) tend to form tight clusters of high performance (high accuracy, low forgetting) which benefit from available knowledge more than their uninformed counterparts.

\begin{table}
    \centering
    \resizebox{\textwidth}{!}{
    \begin{tabular}{ccccccccccc}
        \toprule
        \multirow{3}{*}{Task} & \multirow{3}{*}{\shortstack[c]{Knowledge\\Availability}} & \multirow{3}{*}{Category} & \multirow{3}{*}{Buffer} & \multirow{3}{*}{Distillation} & \multirow{3}{*}{Architecture} & \multirow{3}{*}{\shortstack[c]{Average\\Accuracy}\raisebox{1.5ex}{\:$\uparrow$}} & \multirow{3}{*}{\shortstack[c]{Average\\Forgetting}\raisebox{1.5ex}{\:$\downarrow$}} & \multirow{3}{*}{\shortstack[c]{Forward\\Transfer}\raisebox{1.5ex}{\:$\uparrow$}} & \multirow{3}{*}{\shortstack[c]{Focused\\Average\\Accuracy}\raisebox{2.5ex}{\:$\uparrow$}} & \multirow{3}{*}{\shortstack[c]{Focused\\Average\\Forgetting}\raisebox{2.5ex}{\:$\downarrow$}} \\
         & & & & & & & & & & \\
         & & & & & & & & & & \\
        \midrule
                
                \multirow{4}{*}{\shortstack[c]{Task 1\\(Cifar-100)}} & \multirow{4}{*}{Predicates} & Modular & No buffer & No distillation & \multirow{4}{*}{Modular} & $\textbf{0.62} \pm 0.01$ & $0.06 \pm 0.00$ & $\textbf{0.50} \pm 0.01$ & $0.00 \pm 0.00$ & $0.61 \pm 0.05$\\
                 &  & Replay + Modular & Reservoir (CCE) & No distillation &  & $0.60 \pm 0.01$ & $\textbf{0.05} \pm 0.01$ & $0.49 \pm 0.01$ & $\textbf{0.27} \pm 0.05$ & $0.25 \pm 0.06$\\
                 &  & Distillation + Modular & No buffer & Teacher-distillation &  & $0.61 \pm 0.01$ & $0.06 \pm 0.01$ & $0.48 \pm 0.02$ & $0.00 \pm 0.00$ & $0.50 \pm 0.08$\\
                 &  & All & Reservoir (CCE) & Teacher-distillation &  & $0.59 \pm 0.01$ & $\textbf{0.05} \pm 0.01$ & $0.47 \pm 0.01$ & $0.26 \pm 0.06$ & $\textbf{0.17} \pm 0.07$\\
                \hdashline
                \multirow{4}{*}{\shortstack[c]{Task 1\\(Cifar-100)}} & \multirow{4}{*}{States} & Modular & No buffer & No distillation & \multirow{4}{*}{Modular} & $\textbf{0.62} \pm 0.01$ & $0.06 \pm 0.01$ & $\textbf{0.51} \pm 0.01$ & $0.00 \pm 0.00$ & $0.62 \pm 0.07$\\
                 &  & Replay + Modular & Reservoir (CCE) & No distillation &  & $0.58 \pm 0.02$ & $\textbf{0.05} \pm 0.01$ & $0.47 \pm 0.02$ & $0.18 \pm 0.10$ & $0.32 \pm 0.07$\\
                 &  & Distillation + Modular & No buffer & Teacher-distillation &  & $0.62 \pm 0.01$ & $0.06 \pm 0.01$ & $0.49 \pm 0.01$ & $0.00 \pm 0.00$ & $0.53 \pm 0.04$\\
                 &  & All & Reservoir (CCE) & Teacher-distillation &  & $0.57 \pm 0.02$ & $0.06 \pm 0.01$ & $0.46 \pm 0.02$ & $\textbf{0.19} \pm 0.10$ & $\textbf{0.25} \pm 0.09$\\
                \hdashline
                \multirow{4}{*}{\shortstack[c]{Task 1\\(Cifar-100)}} & \multirow{4}{*}{None} & Naive & No buffer & No distillation & \multirow{4}{*}{Modular} & $\textbf{0.62} \pm 0.01$ & $\textbf{0.06} \pm 0.01$ & $\textbf{0.50} \pm 0.01$ & $0.00 \pm 0.00$ & $0.62 \pm 0.02$\\
                 &  & Replay & Reservoir (CCE) & No distillation &  & $\textbf{0.62} \pm 0.01$ & $\textbf{0.06} \pm 0.01$ & $\textbf{0.50} \pm 0.01$ & $0.00 \pm 0.00$ & $0.61 \pm 0.04$\\
                 &  & Distillation & No buffer & Self-distillation &  & $\textbf{0.62} \pm 0.01$ & $\textbf{0.06} \pm 0.01$ & $0.49 \pm 0.01$ & $0.00 \pm 0.00$ & $0.57 \pm 0.05$\\
                 &  & Replay + Distillation & Reservoir (CCE) & Self-distillation &  & $\textbf{0.62} \pm 0.01$ & $\textbf{0.06} \pm 0.01$ & $0.49 \pm 0.01$ & $0.00 \pm 0.00$ & $\textbf{0.54} \pm 0.05$\\
                \midrule
                \multirow{4}{*}{\shortstack[c]{Task 2\\(Cifar-100)}} & \multirow{4}{*}{Predicates} & Modular & No buffer & No distillation & \multirow{4}{*}{Modular} & $0.27 \pm 0.09$ & $0.36 \pm 0.05$ & $0.17 \pm 0.03$ & $0.36 \pm 0.25$ & $0.41 \pm 0.21$\\
                 &  & Replay + Modular & Reservoir (CCE) & No distillation &  & $0.45 \pm 0.07$ & $0.15 \pm 0.04$ & $\textbf{0.38} \pm 0.06$ & $0.51 \pm 0.13$ & $0.19 \pm 0.09$\\
                 &  & Distillation + Modular & No buffer & Teacher-distillation &  & $0.26 \pm 0.08$ & $0.36 \pm 0.06$ & $0.16 \pm 0.03$ & $0.34 \pm 0.24$ & $0.42 \pm 0.22$\\
                 &  & All & Reservoir (CCE) & Teacher-distillation &  & $\textbf{0.49} \pm 0.04$ & $\textbf{0.13} \pm 0.02$ & $0.35 \pm 0.06$ & $\textbf{0.58} \pm 0.08$ & $\textbf{0.15} \pm 0.06$\\
                \hdashline
                \multirow{4}{*}{\shortstack[c]{Task 2\\(Cifar-100)}} & \multirow{4}{*}{States} & Modular & No buffer & No distillation & \multirow{4}{*}{Modular} & $0.27 \pm 0.09$ & $0.36 \pm 0.05$ & $0.17 \pm 0.03$ & $0.36 \pm 0.25$ & $0.41 \pm 0.22$\\
                 &  & Replay + Modular & Reservoir (CCE) & No distillation &  & $\textbf{0.48} \pm 0.05$ & $\textbf{0.14} \pm 0.02$ & $\textbf{0.37} \pm 0.05$ & $\textbf{0.57} \pm 0.10$ & $\textbf{0.16} \pm 0.07$\\
                 &  & Distillation + Modular & No buffer & Teacher-distillation &  & $0.25 \pm 0.08$ & $0.37 \pm 0.05$ & $0.16 \pm 0.03$ & $0.32 \pm 0.23$ & $0.43 \pm 0.21$\\
                 &  & All & Reservoir (CCE) & Teacher-distillation &  & $\textbf{0.48} \pm 0.03$ & $\textbf{0.14} \pm 0.02$ & $0.35 \pm 0.06$ & $\textbf{0.58} \pm 0.07$ & $\textbf{0.16} \pm 0.05$\\
                \hdashline
                \multirow{4}{*}{\shortstack[c]{Task 2\\(Cifar-100)}} & \multirow{4}{*}{None} & Naive & No buffer & No distillation & \multirow{4}{*}{Modular} & $\textbf{0.31} \pm 0.07$ & $\textbf{0.34} \pm 0.04$ & $\textbf{0.18} \pm 0.03$ & $\textbf{0.40} \pm 0.21$ & $\textbf{0.39} \pm 0.19$\\
                 &  & Replay & Reservoir (CCE) & No distillation &  & $\textbf{0.31} \pm 0.07$ & $\textbf{0.34} \pm 0.04$ & $\textbf{0.18} \pm 0.03$ & $\textbf{0.40} \pm 0.21$ & $\textbf{0.39} \pm 0.19$\\
                 &  & Distillation & No buffer & Self-distillation &  & $0.30 \pm 0.06$ & $0.37 \pm 0.05$ & $0.17 \pm 0.02$ & $0.38 \pm 0.24$ & $0.43 \pm 0.22$\\
                 &  & Replay + Distillation & Reservoir (CCE) & Self-distillation &  & $0.30 \pm 0.07$ & $0.38 \pm 0.05$ & $0.17 \pm 0.03$ & $0.37 \pm 0.25$ & $0.43 \pm 0.23$\\
        \bottomrule
    \end{tabular}
    }
    \caption{Best Average Accuracy, Average Forgetting and Forward Transfer for class-continual experiments on Cifar-100 with trainable backbone, grouped by knowledge available to each strategy. Best models selected by Average Accuracy on validation set. Results are $mean \pm std$ over 9 runs (3 random seeds, times 3 different curricula).}
    \label{tab:incremental-results-cifar}
\end{table}

\begin{figure}
    \centering
    \begin{minipage}{\linewidth}
    \includegraphics[width=\textwidth]{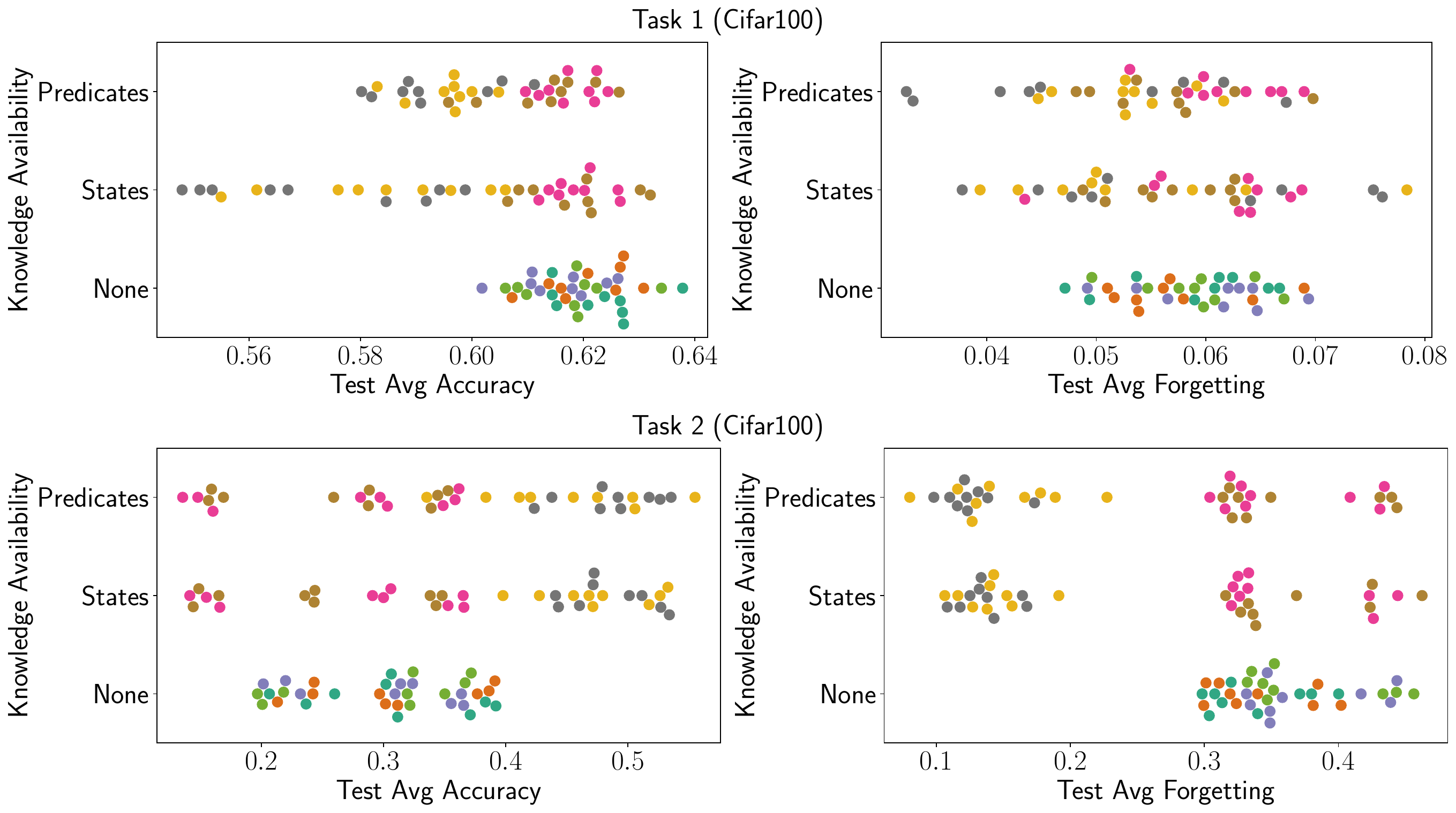}\\
    \includegraphics[width=\textwidth]{imgs/continual/swarms/swarm_legend.pdf}
    \end{minipage}
    \caption{Average accuracy and Average forgetting for class-continual experiments on Cifar-100 with trainable backbone, grouped by knowledge available to each strategy.}
    \label{fig:swarm-cifar}
\end{figure}

\begin{figure}
    \centering
    \begin{minipage}{\linewidth}
    \includegraphics[width=\textwidth]{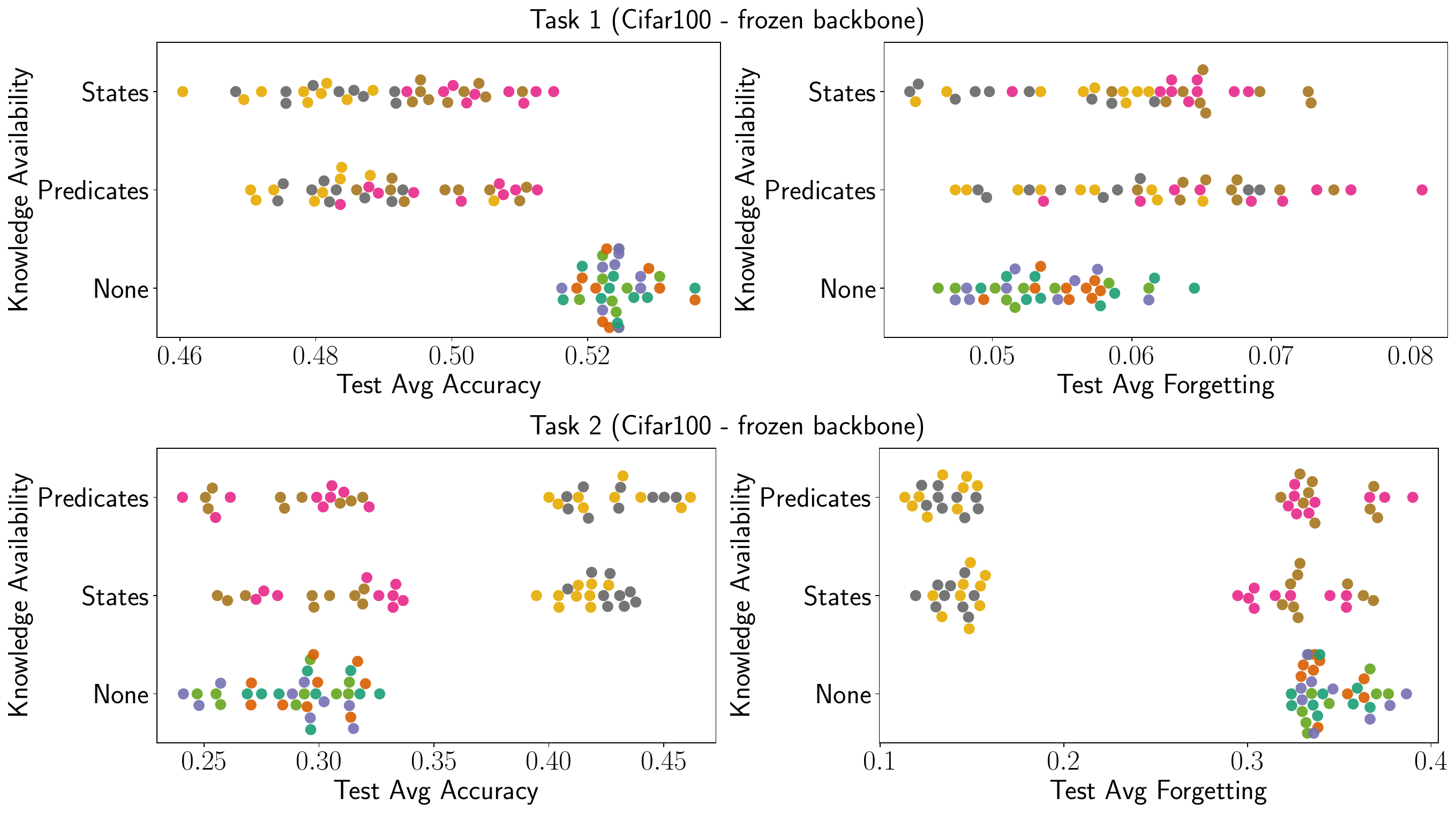}\\
    \includegraphics[width=\textwidth]{imgs/continual/swarms/swarm_legend.pdf}
    \end{minipage}
    \caption{Average accuracy and Average forgetting for class-continual experiments on Cifar-100 with frozen backbone, grouped by knowledge available to each strategy.}
    \label{fig:swarm-cifar-frozen}
\end{figure}

Table~\ref{tab:incremental-results-cifar} and Figure~\ref{fig:swarm-cifar} correspond to Cifar-100 experiments with a trainable backbone.
Also in this case, characterized by more complex perceptual features and 100 classes, we observe a performance significantly above random guessing ($0.01$ accuracy) for every method, but differences at the level of global metrics (average accuracy, average forgetting, forward transfer) are virtually non-existent for task 1, while for task 2, uninformed strategies position themselves between good and bad knowledge-driven strategies, making it impossible to draw strong conclusions on the effect of knowledge availability. This phenomenon can be explained only in part by the more challenging setting, and it is more likely due to weights pre-trained on ImageNet (which presents perceptual features easily adaptable to Cifar-100). However, a different scenario is captured by focused metrics. For task 1 it can be observed how exploiting background knowledge is the only way to avoid catastrophic forgetting on rare labels (even though there are some bad combinations which still fail to classify them by the end of training).
Like in the case of MNIST experiments, metrics on re-occurring labels (task 2) do not provide additional information, compared to global metrics.
We can observe from Figure~\ref{fig:swarm-cifar} that different strategies interact with available knowledge in completely different ways for the two tasks: while replay-based methods are penalized by knowledge in task 1, they seem to perform overall better in task 2 if they are allowed to exploit knowledge.
Table~\ref{tab:incremental-results-cifar-frozen} and Figure~\ref{fig:swarm-cifar-frozen} refer to Cifar-100 experiments with frozen backbone. Although with overall lower scores, the trends identified in experiments with a trainable backbone are confirmed, hinting for an effect of continual adaptation over time that is mostly localized at the final layers.
\begin{table}
    \centering
    \resizebox{\textwidth}{!}{
    \begin{tabular}{ccccccccccc}
        \toprule
        \multirow{3}{*}{Task} & \multirow{3}{*}{\shortstack[c]{Knowledge\\Availability}} & \multirow{3}{*}{Category} & \multirow{3}{*}{Buffer} & \multirow{3}{*}{Distillation} & \multirow{3}{*}{Architecture} & \multirow{3}{*}{\shortstack[c]{Average\\Accuracy}\raisebox{1.5ex}{\:$\uparrow$}} & \multirow{3}{*}{\shortstack[c]{Average\\Forgetting}\raisebox{1.5ex}{\:$\downarrow$}} & \multirow{3}{*}{\shortstack[c]{Forward\\Transfer}\raisebox{1.5ex}{\:$\uparrow$}} & \multirow{3}{*}{\shortstack[c]{Focused\\Average\\Accuracy}\raisebox{2.5ex}{\:$\uparrow$}} & \multirow{3}{*}{\shortstack[c]{Focused\\Average\\Forgetting}\raisebox{2.5ex}{\:$\downarrow$}} \\
         & & & & & & & & & & \\
         & & & & & & & & & & \\
        \midrule
                \multirow{4}{*}{\shortstack[c]{Task 1\\(Cifar-100)\\{[frozen]}}} & \multirow{4}{*}{Predicates} & Modular & No buffer & No distillation & \multirow{4}{*}{Modular} & $\textbf{0.50} \pm 0.01$ & $0.07 \pm 0.01$ & $\textbf{0.44} \pm 0.01$ & $0.00 \pm 0.00$ & $0.52 \pm 0.03$\\
                 &  & Replay + Modular & Reservoir (CCE) & No distillation &  & $0.48 \pm 0.01$ & $\textbf{0.06} \pm 0.01$ & $0.42 \pm 0.01$ & $\textbf{0.24} \pm 0.05$ & $\textbf{0.20} \pm 0.05$\\
                 &  & Distillation + Modular & No buffer & Teacher-distillation &  & $\textbf{0.50} \pm 0.01$ & $0.07 \pm 0.00$ & $\textbf{0.44} \pm 0.01$ & $0.00 \pm 0.00$ & $0.46 \pm 0.03$\\
                 &  & All & Reservoir (CCE) & Teacher-distillation &  & $0.48 \pm 0.01$ & $\textbf{0.06} \pm 0.01$ & $0.42 \pm 0.01$ & $\textbf{0.24} \pm 0.05$ & $\textbf{0.20} \pm 0.04$\\
                \hdashline
                \multirow{4}{*}{\shortstack[c]{Task 1\\(Cifar-100)\\{[frozen]}}} & \multirow{4}{*}{States} & Modular & No buffer & No distillation & \multirow{4}{*}{Modular} & $\textbf{0.50} \pm 0.01$ & $0.06 \pm 0.00$ & $\textbf{0.44} \pm 0.01$ & $0.00 \pm 0.00$ & $0.52 \pm 0.03$\\
                 &  & Replay + Modular & Reservoir (CCE) & No distillation &  & $0.48 \pm 0.01$ & $0.06 \pm 0.01$ & $0.41 \pm 0.01$ & $\textbf{0.18} \pm 0.08$ & $0.24 \pm 0.09$\\
                 &  & Distillation + Modular & No buffer & Teacher-distillation &  & $\textbf{0.50} \pm 0.01$ & $0.07 \pm 0.00$ & $\textbf{0.44} \pm 0.01$ & $0.00 \pm 0.00$ & $0.47 \pm 0.03$\\
                 &  & All & Reservoir (CCE) & Teacher-distillation &  & $0.48 \pm 0.01$ & $\textbf{0.05} \pm 0.01$ & $0.41 \pm 0.01$ & $0.17 \pm 0.08$ & $\textbf{0.22} \pm 0.07$\\
                \hdashline
                \multirow{4}{*}{\shortstack[c]{Task 1\\(Cifar-100)\\{[frozen]}}} & \multirow{4}{*}{None} & Naive & No buffer & No distillation & \multirow{4}{*}{Modular} & $\textbf{0.52} \pm 0.01$ & $0.06 \pm 0.01$ & $\textbf{0.44} \pm 0.01$ & $0.00 \pm 0.00$ & $0.49 \pm 0.04$\\
                 &  & Replay & Reservoir (CCE) & No distillation &  & $\textbf{0.52} \pm 0.01$ & $0.06 \pm 0.00$ & $\textbf{0.44} \pm 0.01$ & $0.00 \pm 0.00$ & $0.49 \pm 0.03$\\
                 &  & Distillation & No buffer & Self-distillation &  & $\textbf{0.52} \pm 0.00$ & $\textbf{0.05} \pm 0.00$ & $0.43 \pm 0.01$ & $0.00 \pm 0.00$ & $0.45 \pm 0.05$\\
                 &  & Replay + Distillation & Reservoir (CCE) & Self-distillation &  & $\textbf{0.52} \pm 0.00$ & $\textbf{0.05} \pm 0.00$ & $0.43 \pm 0.01$ & $0.00 \pm 0.00$ & $\textbf{0.44} \pm 0.05$\\
                \midrule
                \multirow{4}{*}{\shortstack[c]{Task 2\\(Cifar-100)\\{[frozen]}}} & \multirow{4}{*}{Predicates} & Modular & No buffer & No distillation & \multirow{4}{*}{Modular} & $0.29 \pm 0.03$ & $0.34 \pm 0.03$ & $0.20 \pm 0.01$ & $0.34 \pm 0.19$ & $0.37 \pm 0.18$\\
                 &  & Replay + Modular & Reservoir (CCE) & No distillation &  & $\textbf{0.43} \pm 0.02$ & $\textbf{0.13} \pm 0.01$ & $\textbf{0.37} \pm 0.02$ & $\textbf{0.49} \pm 0.06$ & $\textbf{0.15} \pm 0.06$\\
                 &  & Distillation + Modular & No buffer & Teacher-distillation &  & $0.28 \pm 0.03$ & $0.34 \pm 0.02$ & $0.17 \pm 0.02$ & $0.33 \pm 0.19$ & $0.37 \pm 0.18$\\
                 &  & All & Reservoir (CCE) & Teacher-distillation &  & $\textbf{0.43} \pm 0.02$ & $0.14 \pm 0.01$ & $0.35 \pm 0.03$ & $\textbf{0.49} \pm 0.06$ & $0.16 \pm 0.06$\\
                \hdashline
                \multirow{4}{*}{\shortstack[c]{Task 2\\(Cifar-100)\\{[frozen]}}} & \multirow{4}{*}{States} & Modular & No buffer & No distillation & \multirow{4}{*}{Modular} & $0.31 \pm 0.03$ & $0.32 \pm 0.02$ & $0.20 \pm 0.01$ & $0.36 \pm 0.18$ & $0.35 \pm 0.17$\\
                 &  & Replay + Modular & Reservoir (CCE) & No distillation &  & $0.41 \pm 0.01$ & $0.15 \pm 0.01$ & $\textbf{0.34} \pm 0.02$ & $0.48 \pm 0.04$ & $0.16 \pm 0.04$\\
                 &  & Distillation + Modular & No buffer & Teacher-distillation &  & $0.29 \pm 0.03$ & $0.34 \pm 0.02$ & $0.18 \pm 0.02$ & $0.34 \pm 0.18$ & $0.37 \pm 0.17$\\
                 &  & All & Reservoir (CCE) & Teacher-distillation &  & $\textbf{0.43} \pm 0.01$ & $\textbf{0.14} \pm 0.01$ & $0.33 \pm 0.03$ & $\textbf{0.50} \pm 0.05$ & $\textbf{0.15} \pm 0.05$\\
                \hdashline
                \multirow{4}{*}{\shortstack[c]{Task 2\\(Cifar-100)\\{[frozen]}}} & \multirow{4}{*}{None} & Naive & No buffer & No distillation & \multirow{4}{*}{Modular} & $\textbf{0.30} \pm 0.02$ & $\textbf{0.34} \pm 0.02$ & $\textbf{0.19} \pm 0.01$ & $\textbf{0.35} \pm 0.18$ & $\textbf{0.37} \pm 0.17$\\
                 &  & Replay & Reservoir (CCE) & No distillation &  & $\textbf{0.30} \pm 0.02$ & $\textbf{0.34} \pm 0.01$ & $\textbf{0.19} \pm 0.01$ & $\textbf{0.35} \pm 0.18$ & $\textbf{0.37} \pm 0.17$\\
                 &  & Distillation & No buffer & Self-distillation &  & $0.28 \pm 0.03$ & $0.35 \pm 0.02$ & $0.17 \pm 0.01$ & $0.34 \pm 0.20$ & $0.38 \pm 0.18$\\
                 &  & Replay + Distillation & Reservoir (CCE) & Self-distillation &  & $0.29 \pm 0.03$ & $0.35 \pm 0.02$ & $0.17 \pm 0.01$ & $0.34 \pm 0.19$ & $0.38 \pm 0.18$\\

        \bottomrule
    \end{tabular}
    }
    \caption{Best Average Accuracy, Average Forgetting and Forward Transfer for class-continual experiments on Cifar-100 with frozen backbone, grouped by knowledge available to each strategy. Best models selected by Average Accuracy on validation set. Results are $mean \pm std$ over 9 runs (3 random seeds, times 3 different curricula).}
    \label{tab:incremental-results-cifar-frozen}
\end{table}

\section{Conclusions}\label{sec:conc}
In this work we have introduced LTLZinc, a benchmarking framework for relational (constraint-based) and temporal learning and reasoning. Our generation tool allows to build benchmarks for a wide range of learning and reasoning tasks over time, that can benefit the neuro-symbolic, temporal reasoning and continual learning communities, by further pushing research boundaries towards more expressive frameworks, tighter neuro-symbolic integration, and more robust time-driven approaches. 
Experimental evaluation conducted on ten ready-to-use LTLZinc datasets highlight a challenging setting which state-of-the-art neuro-symbolic and continual learning methods struggle to solve.

In the future we plan to investigate the families of LTLZinc tasks that were not covered experimentally in this work. In particular, temporal distant supervision by means of end-to-end neuro-symbolic systems characterized by correct probabilistic semantics, and task-continual learning with background knowledge, employing strong knowledge injection techniques to prevent catastrophic forgetting.
We further plan to extend the ready-to-use LTLZinc datasets to suites of tasks for specific use cases (e.g., temporal specifications for safety-critical systems, life-long learning with periodic domain shifts, temporal reasoning for vision language models, etc.), and to propose baseline methods to address them.


\bibliographystyle{acm}
\bibliography{biblio}

\appendix

\section{Experimental details for Sequence Classification}\label{app:seqexp}
For simplicity, we describe each module of or pipeline (\textsc{ic}, \textsc{cc}, \textsc{nsp}) with a code-oriented syntax that is widely diffused in the AI community.

\subsection{\textsc{IC} Module}\label{app:backbone}
Each variable (input image) is processed by a small convolutional backbone for image classification.

\begin{align*}
 &\mathrm{Conv2d(3, 32, kernel\_size=5)}\\
 &\mathrm{MaxPool2d(kernel\_size=2)}\\
 &\mathrm{Conv2d(32, 64, kernel\_size=5)}\\
 &\mathrm{MaxPool2d(kernel\_size=2)}\\
 &\mathrm{Flatten()}\\
 &\mathrm{ReLU()}\\
 &\mathrm{Linear(64 \cdot 5 \cdot 5, 1024)}\\
 &\mathrm{Dropout(0.5)}\\
 &\mathrm{Linear(1024, num\_classes)}
\end{align*}

When variables share the same dataset, their corresponding backbones also share trainable parameters, while different datasets are processed by a different instance of the same architecture. In cases like Tasks 1 and 2, where some variables use a subset of others, we still exploit parameter sharing, suppressing the unused output neurons of the last linear layer.

\subsection{\textsc{CC} Modules}\label{app:constraints}
\subsubsection{MLP}
Each constraint is processed by the following architecture. Different constraints do not share trainable parameters.
\begin{align*}
 &\mathrm{Linear(num\_classes, num\_neurons)}\\
 &\mathrm{ReLU()}\\
 &\mathrm{Linear(num\_neurons, 1)}\\
 &\mathrm{Sigmoid()}\\
\end{align*}

If temperature calibration is enabled, the last linear layer output is divided by a separate trainable parameter, before the sigmoid layer.

\subsubsection{Symbolic Programs}

\begin{align*}
 &\mathrm{Softmax()}\\
 &\mathrm{Symbolic\_Program()}.
\end{align*}

The Scallop and ProbLog programs are manually defined for each task.\footnote{ProbLog (based on Prolog) is more expressive than Scallop (based on Datalog), however the six proposed tasks can all be encoded in Datalog.} As an example, the following is the Scallop program corresponding to Task 3:
\begin{verbatim}
type var_x(u8), 
 var_y(u8), 
 var_z(u8)
type p_0(), 
 p_1()

\\ All different constraint:
rel p_0() = var_x(x), 
 var_y(y), 
 var_z(z), 
 x != y, y != z, x != z

rel p_1() = var_x(x), 
 var_y(y), 
 var_z(z), 
 x < y + z
\end{verbatim}

And this is the same task encoded in ProbLog:
\begin{verbatim}
% All different constraint:
p_0(X, Y, Z) :- value(X, V1),
  value(Y, V2), 
  value(Z, V3), 
  V1 =\= V2, V2 =\= V3, V3 =\= V1.
  
p_1(X, Y, Z) :- value(X, V1), 
  value(Y, V2), 
  value(Z, V3), 
  V1 < V2 + V3.

query(p_0(var_x, var_y, var_z)).
query(p_1(var_x, var_y, var_z)).
\end{verbatim}

 Both input logits and output probabilities are rescaled by two independent trainable temperature calibration parameters. Since symbolic modules return probabilities, scaling is performed by first transforming the output in logit space, and then back to probabilities: $\mathrm{new\_prob} = \sigma(\frac{\sigma^{-1}(\mathrm{old\_prob})}{\mathrm{temp}})$.

\subsection{\textsc{NSP} Modules}\label{app:nextstate}
Next state prediction modules are recurrently fed with both the \textsc{CC} module output and their own past prediction.

\subsubsection{MLP}
\begin{align*}
 &\mathrm{Linear(num\_states + num\_constraints, num\_neurons)}\\
 &\mathrm{ReLU()}\\
 &\mathrm{Linear(num\_neurons, num\_states)}\\
 &\mathrm{next\_state = Softmax()}\\
 &\mathrm{sequence\_label = Sum(successor[:, accepting\_states])}\\
\end{align*}
The last linear layer output is divided by the corresponding temperature calibration parameter, before the softmax layer.

\subsubsection{GRU}
\begin{align*}
 &\mathrm{enc = Linear(num\_states, num\_neurons)}\\
 &\mathrm{dec = Linear(num\_states, num\_neurons, num\_states)}\\
 &\mathrm{gru = GruCell(num\_constraints, num\_neurons)}\\
 \\
 &\mathrm{s0 = enc(prev\_state)}\\
 &\mathrm{s1 = gru(input, s0)}\\
 &\mathrm{next\_state = Softmax(dec(s1))}\\
 &\mathrm{sequence\_label = Sum(successor[:, accepting\_states])}\\
\end{align*}
The decoder output is divided by the corresponding temperature calibration parameter, before the softmax layer.

\subsubsection{Fuzzy Automaton}
This module follows the implementation of~\cite{umili2023grounding}. We propositionalize constraints by replacing them with unique labels, and then transpose the transition table, in order to build a separate propositional formula for each next state. Each of these formulas is the disjunction of clauses in the form: $previous\_state \wedge transition\_guard$. Finally, we use Simpy~\cite{meurer2017sympy} to simplify each formula.

During inference, we evaluate each formula by performing algebraic model counting on the probability semiring (conjunctions as multiplications, disjunctions as sums).

For calibration, outputs are first converted in logit space, rescaled and finally softmaxed, akin to the Scallop constraint module.

\subsubsection{sd-DNNF Automaton}
This module follows the implementation of~\cite{manginas2024nesya}, with one major difference. In practice we start in the same way as the Fuzzy module, but instead of simplifying each formula, we compile it to a sd-DNNF (in the work of Manginas et al., only the transition guard is compiled, requiring a number of compilation steps which grows as the square of the number of states in the automaton, in our case only a linear number of compilations is needed).
Like the fuzzy modules, evaluation is performed via algebraic model counting on the probability semiring.

Temperature calibration is identic to the fuzzy module.

\subsection{Hyper-Parameters}\label{app:seqhyper}
Every experiment is repeated using 3 different random seeds. If a run diverges due to NaN or infinite gradients, it is discarded.

\noindent\textbf{Optimizer.} Adam (lr: $10^{-3}$).

\noindent\textbf{Pre-Training Epochs.} 1.

\noindent\textbf{Epochs.} 20 vs. 50.

\noindent\textbf{Pre-Training Epochs.} 1.

\noindent\textbf{Pre-Training Lambdas.} $\lambda_{\textsc{ic}} = 1.0$, $\lambda_{\textsc{cc}} = \lambda_{\textsc{nsp}} = \lambda_{\textsc{sc}} = 0.0$.

\noindent\textbf{Lambdas.} $\lambda_{\textsc{ic}} = 0.1$, $\lambda_{\textsc{cc}} = \lambda_{\textsc{nsp}} = \lambda_{\textsc{sc}} = 1.0$.

\section{Experimental Details for Continual Learning}\label{app:contexp}
Every CL experiment is repeated 9 times, using 3 different random seeds and 3 different curricula which satisfy the LTLZinc specification of each task.

\subsection{Hyper-Parameters for MNIST Experiments}

\noindent\textbf{Optimizer.} SGD (lr: $10^{-3}$).

\noindent\textbf{Epochs.} 25.

\noindent\textbf{Batch Size.} 32.

\noindent\textbf{Embedding Size.} 128 neurons.

\noindent\textbf{Hidden Block Size.} 32 neurons for each block.

\noindent\textbf{Lambdas.} $\lambda_{replay} = \lambda_{distil} = \lambda_{supervision} = 1.0$.

\noindent\textbf{Replay Buffer.} Batch size: 16, Buffer size: 500 samples. None vs. Reservoir Sampling.

\noindent\textbf{Buffer Loss.} Categorical Cross-Entropy vs. Gradient Episodic Memory.

\noindent\textbf{Learning without Forgetting.} None vs. Self-Distillation vs Teacher-Student distillation. Minibatches for distillation are taken both from the current episode and replay buffer.

\noindent\textbf{Modular Architecture.} Flat vs. One Hidden Block for each knowledge unit. Teachers are always flat. 

\noindent\textbf{Backbone.} Trained from Scratch.

\subsection{Hyper-Parameters for Cifar-100 Experiments}
Unless specified, hyper-parameters are the same as those for MNIST experiments.

\noindent\textbf{Optimizer.} Adam (lr: $10^{-4}$).

\noindent\textbf{Epochs.} 30.

\noindent\textbf{Buffer Loss.} Categorical Cross-Entropy.

\noindent\textbf{Learning without Forgetting.} None vs. Self-Distillation vs Teacher-Student distillation. Minibatches for distillation are taken only from the current episode.

\noindent\textbf{Modular Architecture.} One Hidden Block for each knowledge unit. Teachers are always flat. 

\noindent\textbf{Backbone.} Pre-Trained Weights vs. Pre-Trained Weights and Frozen. 

\section{Experiment Reproducibility Statement}
Our benchmark generator, along with every experiment discussed in this paper is available at: \url{https://github.com/continual-nesy/LTLZinc}.
All software requirements are specified in the repository. The code is written in Python and tested on various AMD64 Linux distributions.

\paragraph{Generator.} The LTLZinc library is wrapped by an entrypoint script which will take advantage of all the available CPU cores to build multiple LTLZinc datasets at once. The library itself can be invoked sequentially on a single CPU core. Memory requirements largely depend on the complexity of the LTLZinc specification, the size of the generated automaton, and the number of Minizinc instances to solve. GPU is not required for dataset generation. We tested our generator on four low-to-mid-end gaming laptops:
\begin{itemize}
    \item Asus ROG GL552VW (2016): Intel Core i7 6700HQ / 2.6 GHz quad-core, 32 Gb RAM / 64 Gb swap partition, Nvidia GTX960M / 2 Gb VRAM;
    \item Lenovo Legion 5 15ARH05H (2018): AMD Ryzen 7 4800H / 2.9 GHz octa-core, 64 Gb RAM / 128 Gb swap partition, Nvidia RTX2060 / 6 Gb VRAM;
    \item Asus TUF Dash FX516PR (2021): Intel Core i7 11370H / 3.3 GHz quad-core, 40 Gb RAM / 128 Gb swap partition, Nvidia RTX3070 / 8 Gb VRAM;
    \item Asus TUF Gaming A15 FA507NV (2023): AMD Ryzen 7 7735HS Mobile Processor / 3.2 GHz octa-core, 64 Gb RAM / 128 Gb swap partition, Nvidia RTX4060 Laptop / 8 Gb VRAM.
\end{itemize}

\paragraph{Experiments.} Neural networks for our experiments employ the Pytorch library. Experiments run locally on GPU with moderate amount of VRAM, but they do not need large quantities of system RAM.
Software requirements for experiments are different than those of the generator (please refer to \texttt{README} files in subfolders of the repository).
The results of this paper are based on about 4000 experiments, corresponding to roughly 160 days/GPU distributed across three mid-end gaming laptops:
\begin{itemize}
    \item Lenovo Legion 5 15ARH05H (2018): AMD Ryzen 7 4800H / 2.9 GHz octa-core, 64 Gb RAM / 128 Gb swap partition, Nvidia RTX2060 / 6 Gb VRAM;
    \item Asus TUF Dash FX516PR (2021): Intel Core i7 11370H / 3.3 GHz quad-core, 40 Gb RAM / 128 Gb swap partition, Nvidia RTX3070 / 8 Gb VRAM;
    \item Asus TUF Gaming A15 FA507NV (2023): AMD Ryzen 7 7735HS Mobile Processor / 3.2 GHz octa-core, 64 Gb RAM / 128 Gb swap partition, Nvidia RTX4060 Laptop / 8 Gb VRAM.
\end{itemize}

\paragraph{Privacy and Data Leakage Statement.}
Our code does not retrieve, store locally or transmit any user or hardware information, and it can run on fully isolated sandboxes. To the best of our knowledge, the libraries our software relies on do not pose privacy or security threats.

For the sake of ease of management, experiment distribution and hyper-parameter selection can exploit the Weights and Biases library\footnote{\url{https://wandb.ai}.} (this feature is disabled by default). Sweep files for each batch of experiments are included in the repository for reproducibility. Data samples and checkpoints are never transmitted or stored online (by default, our code does not store checkpoints locally either), only hyper-parameter configurations, final metrics, and hardware information (for the purpose of computing metrics such as GPU usage over time) are transmitted to Weights and Biases servers. We currently have no reason to believe large language models to have unfair advantage due to dataset leakage. 


\section{Reproducibility Checklist for JAIR}

Select the answers that apply to your research -- one per item. 

\subsection*{All articles:}

\begin{enumerate}
    \item All claims investigated in this work are clearly stated. 
    [yes]
    \item Clear explanations are given how the work reported substantiates the claims. 
    [yes]
    \item Limitations or technical assumptions are stated clearly and explicitly. 
    [yes]
    \item Conceptual outlines and/or pseudo-code descriptions of the AI methods introduced in this work are provided, and important implementation details are discussed. 
    [yes]
    \item 
    Motivation is provided for all design choices, including algorithms, implementation choices, parameters, data sets and experimental protocols beyond metrics.
    [yes]
\end{enumerate}

\subsection*{Articles containing theoretical contributions:}
Does this paper make theoretical contributions? 
[no] 

If yes, please complete the list below.

\begin{enumerate}
    \item All assumptions and restrictions are stated clearly and formally. 
    [yes/partially/no]
    \item All novel claims are stated formally (e.g., in theorem statements). 
    [yes/partially/no]
    \item Proofs of all non-trivial claims are provided in sufficient detail to permit verification by readers with a reasonable degree of expertise (e.g., that expected from a PhD candidate in the same area of AI). [yes/partially/no]
    \item
    Complex formalism, such as definitions or proofs, is motivated and explained clearly.
    [yes/partially/no]
    \item 
    The use of mathematical notation and formalism serves the purpose of enhancing clarity and precision; gratuitous use of mathematical formalism (i.e., use that does not enhance clarity or precision) is avoided.
    [yes/partially/no]
    \item 
    Appropriate citations are given for all non-trivial theoretical tools and techniques. 
    [yes/partially/no]
\end{enumerate}

\subsection*{Articles reporting on computational experiments:}
Does this paper include computational experiments? [yes]

If yes, please complete the list below.
\begin{enumerate}
    \item 
    All source code required for conducting experiments is included in an online appendix 
    or will be made publicly available upon publication of the paper.
    The online appendix follows best practices for source code readability and documentation as well as for long-term accessibility.
    [yes]
    \item The source code comes with a license that
    allows free usage for reproducibility purposes.
    [yes]
    \item The source code comes with a license that
    allows free usage for research purposes in general.
    [yes]
    \item 
    Raw, unaggregated data from all experiments is included in an online appendix 
    or will be made publicly available upon publication of the paper.
    The online appendix follows best practices for long-term accessibility.
    [yes]
    \item The unaggregated data comes with a license that
    allows free usage for reproducibility purposes.
    [yes]
    \item The unaggregated data comes with a license that
    allows free usage for research purposes in general.
    [yes]
    \item If an algorithm depends on randomness, then the method used for generating random numbers and for setting seeds is described in a way sufficient to allow replication of results. 
    [yes]
    \item The execution environment for experiments, the computing infrastructure (hardware and software) used for running them, is described, including GPU/CPU makes and models; amount of memory (cache and RAM); make and version of operating system; names and versions of relevant software libraries and frameworks. 
    [yes]
    \item 
    The evaluation metrics used in experiments are clearly explained and their choice is explicitly motivated. 
    [yes]
    \item 
    The number of algorithm runs used to compute each result is reported. 
    [yes]
    \item 
    Reported results have not been ``cherry-picked'' by silently ignoring unsuccessful or unsatisfactory experiments. 
    [yes]
    \item 
    Analysis of results goes beyond single-dimensional summaries of performance (e.g., average, median) to include measures of variation, confidence, or other distributional information. 
    [yes]
    \item 
    All (hyper-) parameter settings for 
    the algorithms/methods used in experiments have been reported, along with the rationale or method for determining them. 
    [yes]
    \item 
    The number and range of (hyper-) parameter settings explored prior to conducting final experiments have been indicated, along with the effort spent on (hyper-) parameter optimisation. 
    [NA]
    \item 
    Appropriately chosen statistical hypothesis tests are used to establish statistical significance
    in the presence of noise effects.
    [NA]
\end{enumerate}

\subsection*{Articles using data sets:}
Does this work rely on one or more data sets (possibly obtained from a benchmark generator or similar software artifact)? 
[yes]

If yes, please complete the list below.
\begin{enumerate}
    \item 
    All newly introduced data sets 
    are included in an online appendix 
    or will be made publicly available upon publication of the paper.
    The online appendix follows best practices for long-term accessibility with a license
    that allows free usage for research purposes.
    [yes]
    \item The newly introduced data set comes with a license that
    allows free usage for reproducibility purposes.
    [yes]
    \item The newly introduced data set comes with a license that
    allows free usage for research purposes in general.
    [yes]
    \item All data sets drawn from the literature or other public sources (potentially including authors' own previously published work) are accompanied by appropriate citations.
    [yes]
    \item All data sets drawn from the existing literature (potentially including authors’ own previously published work) are publicly available. [yes]
    \item All new data sets and data sets that are not publicly available are described in detail, including relevant statistics, the data collection process and annotation process if relevant.
    [NA]
    \item 
    All methods used for preprocessing, augmenting, batching or splitting data sets (e.g., in the context of hold-out or cross-validation)
    are described in detail. [NA]
\end{enumerate}

\subsection*{Explanations on any of the answers above (optional):}

[Text here; please keep this brief.]

\end{document}